\documentclass[10pt,journal,compsoc]{IEEEtran}
\usepackage{colortbl}
\usepackage{tabulary}
\usepackage{etoolbox}

\usepackage{hyperref}
\usepackage{url}
\usepackage{graphicx}
\usepackage{amsmath}
\usepackage{amsthm}
\usepackage{amssymb}
\usepackage{booktabs}
\usepackage{bbm}
\usepackage{multicol}
\usepackage{multirow}
\usepackage{graphicx}
\usepackage{xcolor}
\usepackage{enumitem}
\usepackage{makecell}

\newtheorem{proposition}{Proposition}
\newtheorem{definition}{Definition}
\newtheorem{problem}{Problem}

\newcommand{\yingda}[1]{{\color{black}#1}}

\newcommand{\jiangran}[1]{{\color{black}#1}}

\newcommand{\revision}[1]{{\color{black}#1}}
\newcommand{\retwo}[1]{{\color{black}#1}}

\newcommand{\ree}[1]{{\color{black}{#1}}}
\newcommand{\remove}[1]{\ignorespaces}
\newcommand{\SO}{\mathrm{SO}(3)}

\DeclareMathOperator*{\argmin}{arg\,min}

\ifCLASSOPTIONcompsoc
  \usepackage[nocompress]{cite}
\else
  \usepackage{cite}
\fi
\ifCLASSINFOpdf
\else
\fi
\begin{document}
%
\title{Towards Robust Probabilistic Modeling on SO(3)\\ via Rotation Laplace Distribution}
%
%
%
%

\author{Yingda Yin*,
        Jiangran Lyu*,
        Yang Wang,
        Haoran Liu, \\
        He Wang$^\dagger$,\IEEEmembership{Member,~IEEE,}
        Baoquan Chen$^\dagger$, \IEEEmembership{Fellow,~IEEE}
\IEEEcompsocitemizethanks{
\IEEEcompsocthanksitem *: Yingda Yin and Jiangran Lyu are joint first authors.  
\IEEEcompsocthanksitem $^\dagger$: He Wang and Baoquan Chen are corresponding authors. 
Email: \{hewang, baoquan\}@pku.edu.cn \protect \\
\vspace{-\baselineskip}
\IEEEcompsocthanksitem Yingda Yin, Jiangran Lyu and He Wang are with School of Computer Science, Peking University.
\IEEEcompsocthanksitem Yang Wang and Haoran Liu are with School of EECS, Peking University.
\IEEEcompsocthanksitem Baoquan Chen is with School of Intelligence Technology, Peking University.

}
\thanks{Manuscript received May 17, 2023.}}

%
%

\markboth{IEEE TRANSACTIONS ON PATTERN ANALYSIS AND MACHINE INTELLIGENCE }
{ \MakeLowercase{\textit{et al.}}: Rotation Laplace}
%



\IEEEtitleabstractindextext{%
\begin{abstract}
\yingda{
Estimating the 3DoF rotation from a single RGB image is an important yet challenging problem. 
As a popular approach, probabilistic rotation modeling additionally carries prediction uncertainty information, compared to single-prediction rotation regression.
For modeling probabilistic distribution over $\SO$, it is natural to use Gaussian-like Bingham distribution and matrix Fisher, however they are shown to be sensitive to outlier predictions, e.g. $180^\circ$ error and thus are unlikely to converge with optimal performance.
In this paper, we draw inspiration from multivariate Laplace distribution and propose a novel rotation Laplace distribution on $\SO$. Our rotation Laplace distribution is robust to the disturbance of outliers and enforces much gradient to the low-error region that it can improve.
\jiangran{In addition, we show that our method also exhibits robustness to small noises and thus tolerates imperfect annotations. With this benefit, we demonstrate its advantages in semi-supervised rotation regression, where the pseudo labels are noisy.}
\jiangran{To further capture the multi-modal rotation solution space for symmetric objects, we extend our distribution to rotation Laplace mixture model and demonstrate its effectiveness.}
Our extensive experiments show that our proposed distribution \jiangran{and the mixture model} achieve state-of-the-art performance in all the rotation regression experiments over both probabilistic and non-probabilistic baselines.
}
\end{abstract}

}

\maketitle

\IEEEdisplaynontitleabstractindextext

%
\IEEEpeerreviewmaketitle

\IEEEraisesectionheading{\section{Introduction}\label{sec:introduction}}


\yingda{
\IEEEPARstart{I}{ncorporating} neural networks \cite{Morris1990neural} to perform rotation regression is of great importance in the field of computer vision, computer graphics and robotics \cite{wang2019normalized,yin2022fishermatch,dong2021robust,breyer2021volumetric,ci2022locally}. To close the gap between the $\SO$ manifold and the Euclidean space where neural network outputs exist, one popular line of research discovers learning-friendly rotation representations including 6D continuous representation \cite{zhou2019continuity}, 9D matrix representation with SVD orthogonalization \cite{levinson2020analysis}, etc. Recently, Chen \textit{et al.} \cite{chen2022projective} focuses on the gradient backpropagating process and replaces the vanilla auto differentiation with a $\SO$ manifold-aware gradient layer, which sets the new state-of-the-art in rotation regression tasks. 

Reasoning about the uncertainty information along with the predicted rotation is also attracting more and more attention, which enables many applications in aerospace \cite{crassidis2003unscented}, autonomous driving \cite{mcallister2017concrete,gurkirt2023road} and localization \cite{fang2020towards, wei2017model}.
}
On this front, recent efforts have been developed to model the uncertainty of rotation regression via probabilistic modeling of rotation space. The most commonly used distributions are Bingham distribution \cite{bingham1974antipodally} on $\mathcal{S}^3$ for unit quaternions and matrix Fisher distribution \cite{khatri1977mises} on $\SO$ for rotation matrices. These two distributions are equivalent to each other \cite{prentice1986orientation} and resemble the Gaussian distribution in Euclidean Space \cite{bingham1974antipodally,khatri1977mises}.
While modeling noise using Gaussian-like distributions is well-motivated by the Central Limit Theorem, Gaussian distribution is well-known to be sensitive to outliers in the probabilistic regression models \cite{murphy2012machine}. This is because Gaussian distribution penalizes deviations quadratically, so predictions with larger errors weigh much more heavily with the learning than low-error ones and thus potentially result in suboptimal convergence when a certain amount of outliers exhibit \cite{yang2023cer}.

\begin{figure}[t]
    \centering
    \hspace{-3mm}
    \includegraphics[width=\linewidth]{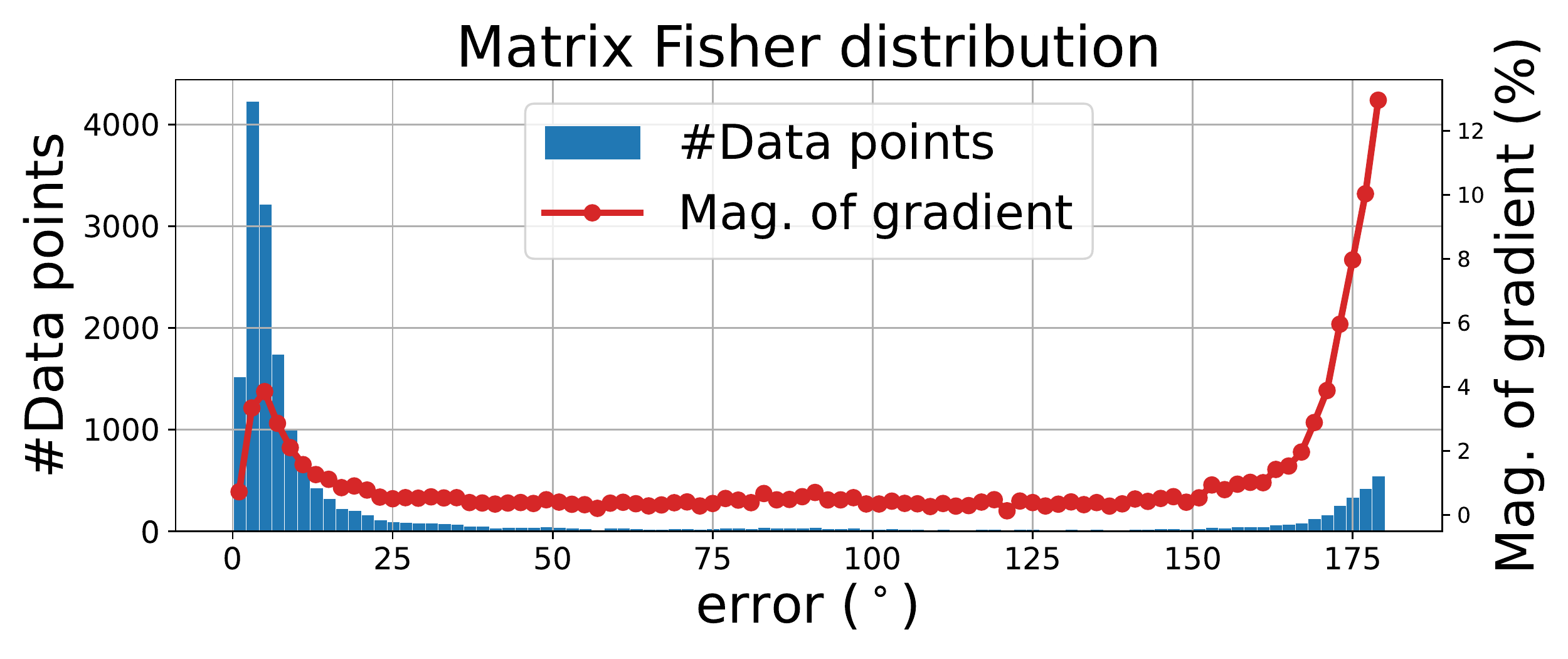}
    \hspace{2mm}\\
    \includegraphics[width=\linewidth]{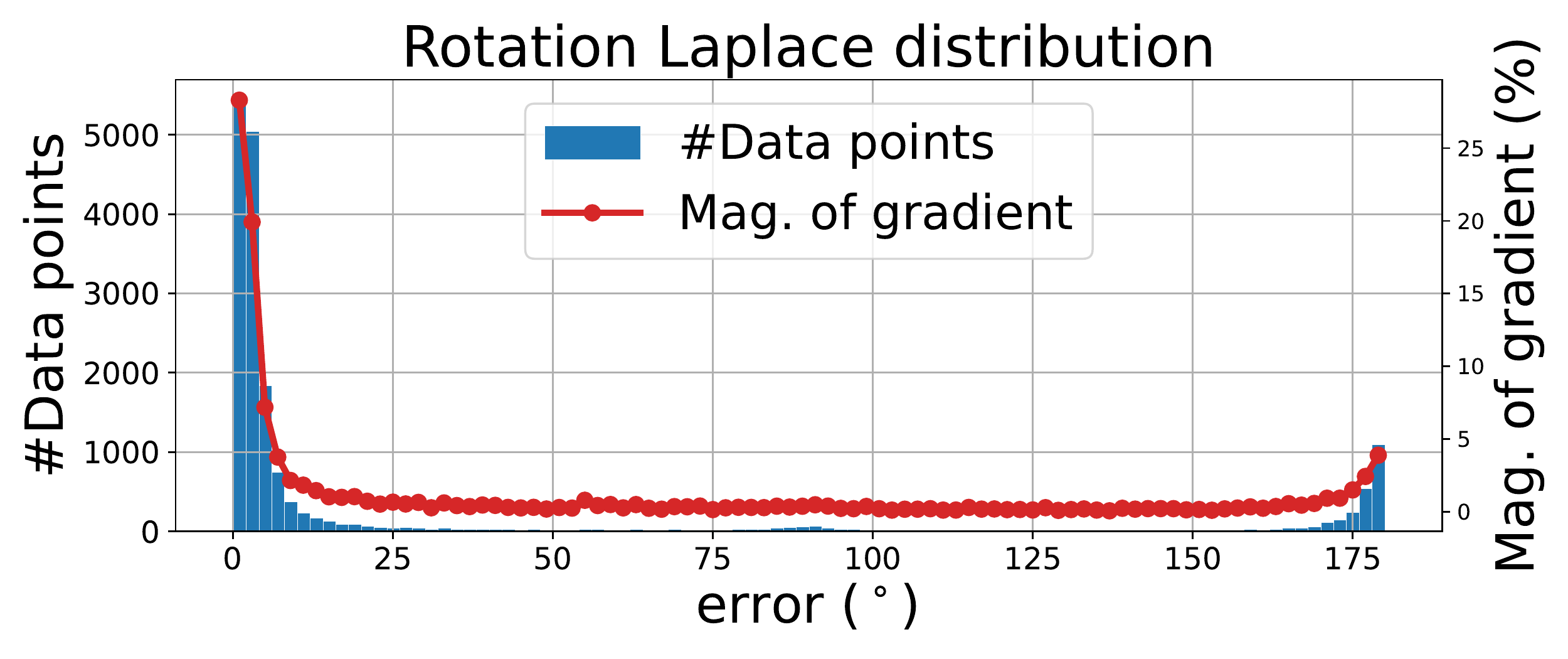}
    \hspace{-3mm}
    \vspace{-3mm}
    \caption{\small Visualization of the results of matrix Fisher distribution and rotation Laplace distribution after convergence. The horizontal axis is the geodesic distance between the prediction and the ground truth. The blue bins count the number of data points within corresponding errors (2$^\circ$ each bin). \yingda{The red dots illustrate the percentage of the sum of the gradient magnitude  $\|\partial \mathcal{L} / \partial (\text{dist. param.})\|$
    within each bin. }
    The experiment is done on all categories of ModelNet10-SO3 dataset.}
    \vspace{-2mm}
	\label{fig:teaser}
\end{figure}

Unfortunately, in certain rotation regression tasks, we fairly often come across large prediction errors, \textit{e.g.} $180^\circ$ error,  due to either the (near) symmetry nature of the objects or severe occlusions \cite{murphy2021implicit}. 
In Fig. \ref{fig:teaser}(left), using training on single image rotation regression as an example, we show the statistics of predictions 
after achieving convergence, assuming matrix Fisher distribution (as done in \cite{mohlin2020probabilistic}). The blue histogram shows the population with different prediction errors and the red dots are the impacts of these predictions on learning, evaluated by computing the sum of their gradient magnitudes  $\|\partial \mathcal{L} / \partial (\text{distribution param.})\|$    within each bin and then normalizing them across bins.  
It is clear that the 180$^\circ$ outliers dominate the gradient as well as the network training though their population is tiny, while the vast majority of points with low error predictions are deprioritized. Arguably, at convergence, the gradient should focus more on refining the low errors rather than fixing the inevitable large errors (\textit{e.g.} arose from symmetry). This motivates us to find a better probabilistic model for rotation.

As pointed out by \cite{murphy2012machine}, Laplace distribution, with heavy tails, is a better option for robust probabilistic modeling. Laplace distribution drops sharply around its mode and thus allocates most of its probability density to a small region around the mode; meanwhile, it also tolerates and assigns higher likelihoods to the outliers, compared to Gaussian distribution.
Consequently, it encourages predictions near its mode to be even closer, thus fitting \textit{sparse} data well, most of whose data points are close to their mean with the exception of several outliers\cite{mitianoudis2012generalized,munoz2016laplace}, which makes Laplace distribution to be favored in the context of deep learning\cite{goodfellow2016deep}.

In this work, we propose a novel Laplace-inspired distribution on $\SO$ for rotation matrices, namely rotation Laplace distribution, for probabilistic rotation regression. 
We devise rotation Laplace distribution to be an approximation of multivariate Laplace distribution in the tangent space of its mode.
As shown in the visualization in Fig. \ref{fig:teaser}(right), our rotation Laplace distribution is robust to the disturbance of outliers, with most of its gradient contributed by the low-error region, and thus leads to a better convergence along with significantly higher accuracy.
Moreover, our rotation Laplace distribution is simply parameterized by an unconstrained $3\times3$ matrix and thus accommodates the Euclidean output of neural networks with ease. This network-friendly distribution requires neither complex functions to fulfill the constraints of parameterization nor any normalization process from Euclidean to rotation manifold which has been shown harmful for learning \cite{chen2022projective}.For completeness of the derivations, we also propose the Laplace-inspired distribution on $\mathcal{S}^3$ for quaternions. We show that rotation Laplace distribution is equivalent to Quaternion Laplace distribution, similar to the equivalence of matrix Fisher distribution and Bingham distribution.

We extensively compare our rotation Laplace distributions to methods that parameterize distributions on $\SO$ for pose estimation, and also non-probabilistic approaches including multiple rotation representations and recent $\SO$-aware gradient layer \cite{chen2022projective}.
On common benchmark datasets of rotation estimation from RGB images, we achieve a significant and consistent performance improvement over all baselines. 
\jiangran{For example, on ModelNet10-SO3 dataset, rotation Laplace distribution achieves relative improvement of around 35\% on median error, and over 50\% on 3 degree accuracy against the best competitor.
}
\revision{Additionally, we apply our method to the monocular 6D object pose estimation task, which results in performance improvements on both YCB-video and LINEMOD datasets.
}

\jiangran{


The superiority of rotation Laplace distribution are mainly benefited from the robustness to outliers. To gain a deeper understanding of this property, we provide more analysis from the aspect of training gradients, and further conduct experiments by manually injecting outliers into the perfectly labeled synthetic dataset. 
The results demonstrate that not only our model outperforms the baselines, but also better tolerate the outlier injections with significantly less performance degradation.


Additionally, due to the heavy-tail nature of our distribution, there is more probability density at the mode. One may concern that it will be sensitive to small noise perturbations of ground truths, since the  wrong gradient around the mode can be large. 
To address this concern, we experiment with the perturbed data containing small noise injections and find that our method outperforms the baseline at all levels of perturbations and lead to comparable performance drop with the noise injections, illustrating its robustness to noise. 
Building upon these insights, we apply rotation Laplace distribution to the task of semi-supervised rotation regression where the pseudo labels are imperfect. Our method achieves new state-of-the-art in semi-supervised rotation regression tasks. 

To better capture the multimodal rotation space, particularly for symmetric objects, we propose an extension of the rotation Laplace distribution in the form of a mixture model. Our rotation Laplace mixture model allows for the generation of multiple candidate predictions for one object, thereby improving its ability to capture the complete pose space for symmetric objects. 
We compare rotation Laplace mixture model with other methods that are capable capturing multimodal solutions and demonstrate the superior performance of our model.

}

\section{Related Work}



\subsection{Probabilistic regression} 
Nex and Weigend \cite{nix1994estimating} first proposes to model the output of the neural network as a Gaussian distribution and learn the Gaussian parameters by the negative log-likelihood loss function, through which one obtains not only the target but also a measure of prediction uncertainty.
More recently, Kendall and Gal \cite{kendall2017uncertainties} offers more understanding and analysis of the underlying uncertainties. Lakshminarayanan \textit{et al.} \cite{lakshminarayanan2017simple} further improves the performance of uncertainty estimation by network ensembling and adversarial training. Makansi \textit{et al.} \cite{makansi2019overcoming} stabilizes the training with the winner-takes-all and iterative grouping strategies.
Probabilistic regression for uncertainty prediction has been widely used in various applications, including optical flow estimation\cite{ilg2018uncertainty,de2022how}, depth estimation \cite{poggi2020uncertainty,qi2022geonet++}, weather forecasting \cite{wang2019deep}, \textit{etc.}

Among the literature of decades, the majority of probabilistic regression works model the network output by a Gaussian-like distribution, while Laplace distribution is less discovered. 
Li \textit{et al.} \cite{li2021human} empirically finds that assuming a Laplace distribution in the process of maximum likelihood estimation yields better performance than a Gaussian distribution, in the field of 3D human pose estimation. Recent work \cite{nair2022maximum} makes use of Laplace distribution to improve the robustness of maximum likelihood-based uncertainty estimation. Due to the heavy-tailed property of Laplace distribution, the outlier data produces comparatively less loss and have an insubstantial impact on training.
Other than in Euclidean space, Mitianoudis \textit{et al.} \cite{mitianoudis2012generalized} develops Generalized
Directional Laplacian distribution in $\mathcal{S}^d$ for the application of audio separation.

\subsection{Probabilistic rotation regression}
Several works focus on utilizing probability distributions on the rotation manifold for rotation uncertainty estimation. 
Prokudin \textit{et al.} \cite{prokudin2018deep} uses the mixture of von Mises distributions \cite{mardia2000directional} over Euler angles using Biternion networks. In \cite{gilitschenski2019deep} and \cite{deng2022deep}, Bingham distribution over unit quaternion is used to jointly estimate a probability distribution over all axes. 
Mohlin \textit{et al.} \cite{mohlin2020probabilistic} leverages matrix Fisher distribution \cite{khatri1977mises} on $\SO$ over rotation matrices for deep rotation regression. 
Though both bear similar properties with Gaussian distribution in Euclidean space, matrix Fisher distribution benefits from the continuous rotation representation and unconstrained distribution parameters, which yields better performance \cite{murphy2021implicit}.
Recently, Murphy \textit{et al.} \cite{murphy2021implicit} introduces a non-parametric implicit pdf over $\SO$, with the distribution properties modeled by the neural network parameters. Implicit-pdf especially does good for modeling rotations of symmetric objects.

\subsection{Non-probabilistic rotation regression}
The choice of rotation representation is one of the core issues concerning rotation regression. The commonly used representations include Euler angles \cite{kundu20183d,tulsiani2015viewpoints}, unit quaternion \cite{kendall2017geometric,kendall2015posenet,xiang2017posecnn,qin2023fast} and axis-angle \cite{do2018deep,gao2018occlusion,ummenhofer2017demon}, \textit{etc}. However, Euler angles may suffer from gimbal lock, and unit quaternions doubly cover the group of $\SO$, which leads to two disconnected local minima. Moreover, Zhou \textit{et al.} \cite{zhou2019continuity} points out that all representations in the real Euclidean spaces of four or fewer dimensions are discontinuous and are not friendly for deep learning. To this end, the continuous 6D representation with Gram-Schmidt orthogonalization \cite{zhou2019continuity} and 9D representation with SVD orthogonalization \cite{levinson2020analysis} have been proposed, respectively. More recently, Chen \textit{et al.} \cite{chen2022projective} investigates the gradient backpropagation in the backward pass and proposes a $\SO$ manifold-aware gradient layer.


\jiangran{\section{Notations and Definitions}
\label{sec:notation}

\subsection{Notations for Lie Algebra and Exponential \& Logarithm Map}

This paper follows the common notations for Lie algebra and exponential \& logarithm map \cite{lee2018bayesian,teed2021tangent,sola2018micro}.

The three-dimensional special orthogonal group $\SO$ is defined as 
{\footnotesize \begin{equation*}
    \SO = \{ \mathbf{R} \in \mathbb{R}^{3\times3} | \mathbf{RR}^T = \mathbf{I}, \det{(\mathbf{R})} = 1 \}.
\end{equation*}
}The Lie algebra of $\SO$, denoted by $\mathfrak{so}(3)$, is the tangent space of $\SO$ at $\mathbf{I}$, given by
{\footnotesize\begin{equation*}
    \mathfrak{so}(3) = \{ \boldsymbol{\Phi} \in \mathbb{R}^{3\times 3}| \boldsymbol{\Phi} = -\boldsymbol{\Phi}^T\}.
\end{equation*}
}$\mathfrak{so}(3)$ is identified with $(\mathbb{R}^3, \times)$ by the \textit{hat} $\wedge$ map and the \textit{vee} $\vee$ map defined as
{\small\begin{equation*}
    \mathfrak{s o}(3) \ni\left[\begin{array}{ccc}
0 & -\phi_z & \phi_y \\
\phi_z & 0 & -\phi_x \\
-\phi_y & \phi_x & 0
\end{array}\right] 
\stackrel{\text { vee } \vee} {\underset{\text { hat }\wedge}{\rightleftarrows}}
\left[\begin{array}{l}\phi_x \\\phi_y \\\phi_z\end{array}\right] \in \mathbb{R}^3
\end{equation*}}

The exponential map, taking skew symmetric matrices to rotation matrices is given by
{\footnotesize\begin{equation*}
    \exp(\hat{\boldsymbol{\phi}})=\sum_{k=0}^{\infty}{\frac{\hat{\boldsymbol{\phi}}^k}{k!}}=\mathbf{I}+\frac{\sin{\theta}}{\theta}{\hat{\boldsymbol{\phi}}}+\frac{1-\cos{\theta}}{\theta^2}{\hat{\boldsymbol{\phi}}^2},
\end{equation*}
}where $\theta=\left\lVert {\boldsymbol{\phi}} \right\rVert$.
The exponential map can be inverted by the logarithm map, going from $\SO$ to $\mathfrak{so}(3)$ as
{\footnotesize\begin{equation*}
    \log(\mathbf{R})=\frac{\theta}{2\sin{\theta}}(\mathbf{R}-\mathbf{R}^T),
\end{equation*}
}where $\theta=\arccos{\frac{\operatorname{tr}(\mathbf{R})-1}{2}}$.

\subsection{Haar Measure}
\label{sec:haar}
To evaluate the normalization factors and therefore the probability density functions, the measure $\mathrm{d}\mathbf{R}$ on $\SO$ needs to be defined. For the Lie group $\SO$, the commonly used bi-invariant measure is referred to as Haar measure \cite{haar1933massbegriff,james1999history}. Haar measure is unique up to scalar multiples \cite{chirikjian2000engineering} and we follow the common practice \cite{mohlin2020probabilistic,lee2018bayesian} that the Haar measure $\mathrm{d}\mathbf{R}$ is scaled such that $\int_{\SO} \mathrm{d} \mathbf{R}=1$.
}
\section{Laplace-inspired Distribution on SO(3)}
\subsection{Revisit matrix Fisher distribution}

\subsubsection{Matrix Fisher Distribution}

Matrix Fisher distribution (or von Mises-Fisher matrix distribution) \cite{khatri1977mises} is one of the widely used distributions for probabilistic modeling of rotation matrices. 

\begin{definition} \textit{\emph{Matrix Fisher distribution}}.
The random variable $\mathbf{R}\in \SO$ follows matrix Fisher distribution with parameter $\mathbf{A}$, if its probability density function is defined as
\begin{equation}
    p(\mathbf{R}; \mathbf{A}) = \frac{1}{F(\mathbf{A})}
    \exp\left(
    \operatorname{tr}(\mathbf{A}^T \mathbf{R})
    \right)
\end{equation}
where $\mathbf{A}\in \mathbb{R}^{3\times 3}$ is an unconstrained matrix, and $F(\mathbf{A})\in \mathbb{R}$ is the normalization factor. Without further clarification, we denote $F$ as the normalization factor of the corresponding distribution in the remaining of this paper. We also denote matrix Fisher distribution as $\mathbf{R} \sim \mathcal{MF}(\mathbf{A})$.
\end{definition}
Suppose the singular value decomposition of matrix $\mathbf{A}$ is given by $\mathbf{A} = \mathbf{U}^\prime \mathbf{S}^\prime (\mathbf{V}^\prime)^T$, \textit{proper} SVD is defined as $\mathbf{A} = \mathbf{USV}^T$
where 
{\small
\begin{equation*}
    \begin{aligned}
        \mathbf{U} = \mathbf{U}^\prime\operatorname{diag}(1, 1, \det(\mathbf{U}^\prime)) \qquad
        \mathbf{V} = \mathbf{V}^\prime \operatorname{diag}(1, 1, \det(\mathbf{V}^\prime)) \\
        \mathbf{S} = \operatorname{diag}({s_1}, {s_2}, {s_3}) = 
        \operatorname{diag}(s_1^\prime, s_2^\prime, \det(\mathbf{U}^\prime\mathbf{V}^\prime)s_3^\prime)\\
    \end{aligned}
\end{equation*}
}The definition of $\mathbf{U}$ and $\mathbf{V}$ ensures that $\det(\mathbf{U})=\det(\mathbf{V})=1$ and $\mathbf{U}, \mathbf{V} \in \SO$. 



\subsubsection{Relationship between Matrix Fisher Distribution in $\SO$ and Gaussian Distribution in $\mathbb{R}^3$}
\label{sec:fisher_gauss}
It is shown that matrix Fisher distribution is highly relevant with zero-mean Gaussian distribution near its mode
\ree{\cite{lee2018bayesian,lee2018bayesian_approximate}}.
Denote $\mathbf{R}_0$ as the mode of matrix Fisher distribution, and define $\mathbf{\widetilde{R}}=\mathbf{R}_0^T\mathbf{R}$, the relationship is shown as follows. Please refer to supplementary for the proof.
\begin{proposition}
\label{prop:fisher_gaussian}
Let $\boldsymbol{\Phi} = \log \mathbf{\widetilde{R}} \in \mathfrak{so}(3)$ and $\boldsymbol{\phi} = {\boldsymbol{\Phi}^\vee} \in \mathbb{R}^3$. For rotation matrix $\mathbf{R} \in \SO$ following \emph{matrix Fisher distribution}, when 
\ree{$\|\mathbf{R} - \mathbf{R}_0 \| \rightarrow 0$}
, $\boldsymbol{\phi}$ follows zero-mean \emph{multivariate Gaussian distribution}.
\end{proposition}

\subsection{Rotation Laplace Distribution}
We get inspiration from multivariate Laplace distribution \ree{\cite{eltoft2006multivariate,kozubowski2013multivariate}}, defined as follows.
\begin{definition} \textit{\emph{Multivariate Laplace distribution.}}
If means $\boldsymbol{\mu}=\mathbf{0}$, the d-dimensional multivariate Laplace distribution with covariance matrix $\boldsymbol{\Sigma}$ 
is defined as
{
\begin{equation*}
    p(\mathbf{x};\boldsymbol{\Sigma}) = \frac{1}{F}
    \left(\mathbf{x}^T \boldsymbol{\Sigma}^{-1} \mathbf{x}\right)^{v / 2} K_{v}\left(\sqrt{2 \mathbf{x}^T \boldsymbol{\Sigma}^{-1} \mathbf{x}}\right)
\end{equation*}
}where $v = (2 - \ree{d}) / 2$ and $K_v$ is modified Bessel function of the second kind.
\end{definition}
We consider three dimensional Laplace distribution of $\mathbf{x}\in \mathbb{R}^3$
\ree{(i.e. $d=3$ and $v=-\frac{1}{2}$). Given the property $K_{-\frac{1}{2}}(\xi)\propto \xi^{-\frac{1}{2}} \exp (-\xi)$, three dimensional Laplace distribution is} defined as
{
\begin{equation*}
    p(\mathbf{x};\boldsymbol{\Sigma}) = \frac{1}{F}
    \frac{\exp \left( -\sqrt{2 \mathbf{x}^T \boldsymbol{\Sigma}^{-1} \mathbf{x}} \right)}{\sqrt{\mathbf{x}^T \boldsymbol{\Sigma}^{-1} \mathbf{x}}}
\end{equation*}}

In this section, we first give the definition of our proposed rotation Laplace distribution and then shows its relationship with multivariate Laplace distribution.

\begin{definition} \textit{\emph{Rotation Laplace distribution.}}
The random variable $\mathbf{R}\in \SO$ follows rotation Laplace distribution with parameter $\mathbf{A}$, if its probability density function is defined as
\begin{equation}
\label{eq:rl}
\small
    p(\mathbf{R}; \mathbf{A}) = \frac{1}{F(\mathbf{A})}
    \frac{\exp\left(-\sqrt{\operatorname{tr}\left(\mathbf{S} - \mathbf{A}^T \mathbf{R}\right)}\right)}
    {\sqrt{\operatorname{tr}\left(\mathbf{S} -\mathbf{A}^T \mathbf{R}\right)}}
\end{equation}
where $\mathbf{A}\in \mathbb{R}^{3\times 3}$ is an unconstrained matrix, and $\mathbf{S}$ is the diagonal matrix composed of the proper singular values of matrix $\mathbf{A}$, i.e., $\mathbf{A=USV}^T$. We also denote rotation Laplace distribution as $\mathbf{R} \sim \mathcal{RL}(\mathbf{A})$.
\end{definition}

Denote $\mathbf{R}_0$ as the mode of rotation Laplace distribution and define $\mathbf{\widetilde{R}}=\mathbf{R}_0^T\mathbf{R}$, the relationship between rotation Laplace distribution and multivariate Laplace distribution is shown as follows.
\begin{proposition}
Let $\boldsymbol{\Phi} = \log \mathbf{\widetilde{R}} \in \mathfrak{so}(3)$ and $\boldsymbol{\phi} = {\boldsymbol{\Phi}^\vee} \in \mathbb{R}^3$. For rotation matrix $\mathbf{R} \in \SO$ following \emph{rotation Laplace distribution}, when
\ree{$\|\mathbf{R} - \mathbf{R}_0\|\rightarrow 0$}
, $\boldsymbol{\phi}$ follows zero-mean \emph{multivariate Laplace distribution}.
\label{prop:laplace}
\end{proposition}

\revision{We provide the proof of Prop. \ref{prop:laplace} in the supplementary.}



\revision{
\subsection{Properties}

\begin{figure}[t]
    \centering
    \begin{tabular}{ccc}
    \includegraphics[clip,trim=1cm 3cm 1cm 0cm,width=0.25\linewidth]{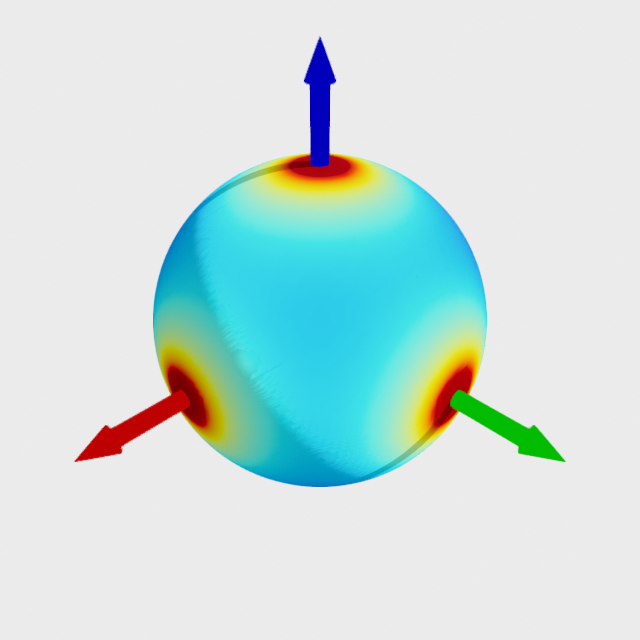}\hspace{0mm}
    &
    \includegraphics[clip,trim=1cm 3cm 1cm 0cm,width=0.25\linewidth]{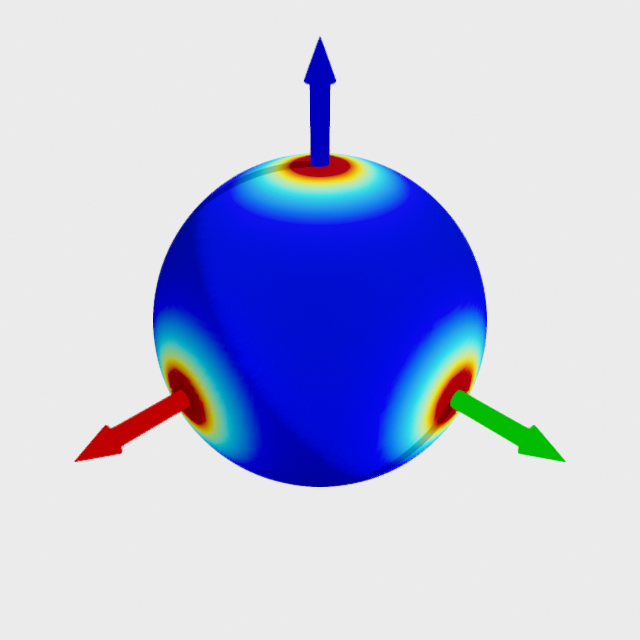}\hspace{0mm}
    &
    \includegraphics[clip,trim=1cm 3cm 1cm 0cm,width=0.25\linewidth]{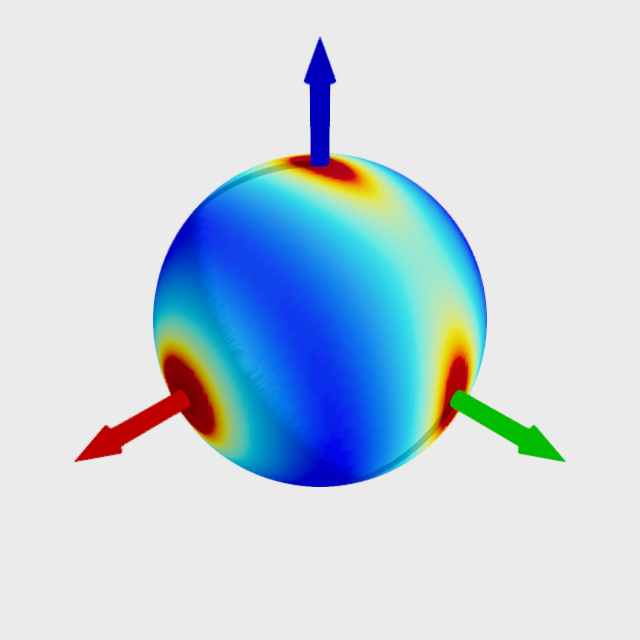}
    \\
    {\scriptsize (a) diag(5, 5, 5)} \hspace{0mm} &
    {\scriptsize (b) diag(25, 25, 25)} \hspace{0mm} &
    {\scriptsize (c) diag(25, 5, 1)} 
    \\
    \includegraphics[clip,trim=1cm 3cm 1cm 0cm,width=0.25\linewidth]{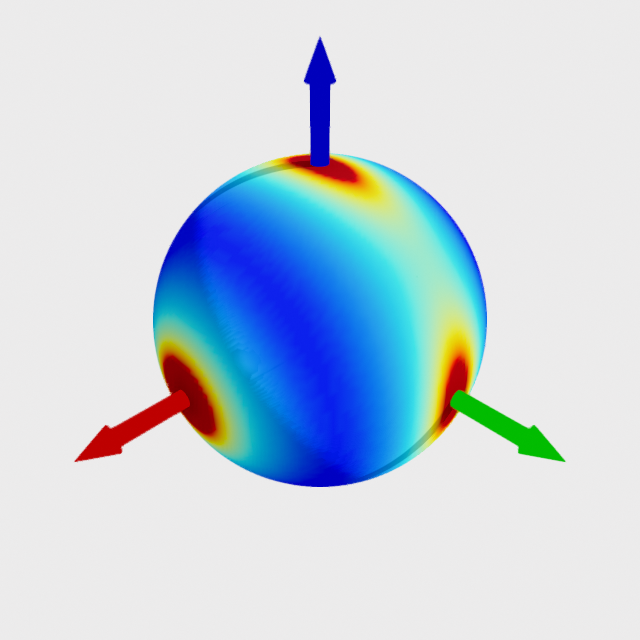}\hspace{0mm}
    &\includegraphics[clip,trim=1cm 3cm 1cm 0cm,width=0.25\linewidth]{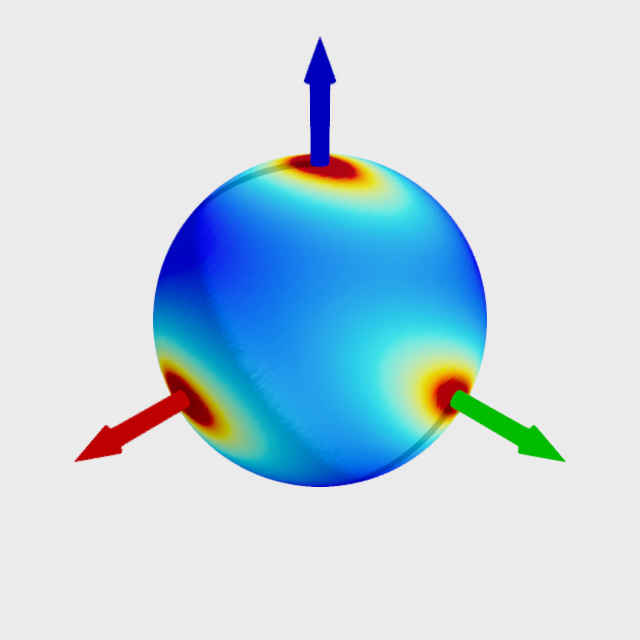}\hspace{0mm}
    &\includegraphics[clip,trim=1cm 3cm 1cm 0cm,width=0.25\linewidth]{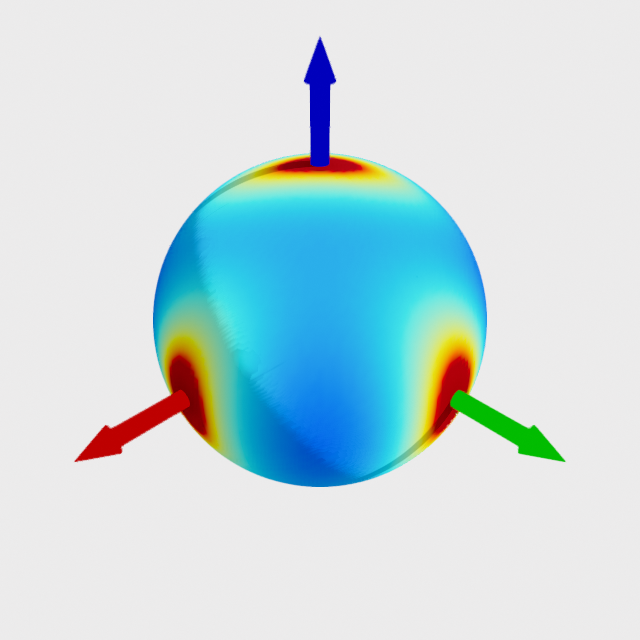}
    \\
    {\scriptsize (d) $\mathbf{T}_x$ diag(25, 5, 1) $\mathbf{T}_x^T$} \hspace{0mm}&
    {\scriptsize (e) $\mathbf{T}_y$ diag(25, 5, 1) $\mathbf{T}_y^T$} \hspace{0mm}&
    {\scriptsize (f) $\mathbf{T}_z$ diag(25, 5, 1) $\mathbf{T}_z^T$} 
    
    \end{tabular}
    \vspace{-2mm}
    \caption{\small \textbf{Effect of the parameter $\mathbf{A}$ on the shape of rotation Laplace distribution.} $\mathbf{T}_i$ refers to the rotation matrix by rotating $\pi/4$ around $e_i$.}
    \vspace{-2mm}
    \label{fig:vis_shape}
\end{figure}

In this section, we provide several statistical properties of rotation Laplace distribution. Detailed derivations and proofs are included in the supplementary materials.

\textbf{Mode.}
Suppose the proper singular value decomposition of the parameter $\mathbf{A}$ is given by $\mathbf{A}=\mathbf{USV}^T$. 
\retwo{
Denote $\mathbf{S}=\operatorname{diag}(s_1, s_2, s_3)$,
the uniqueness of the {mode} depends on the singular values. In most cases, i.e., $s_2 + s_3 > 0$, the {mode} is unique and 
}
computed as 
\begin{equation}
    \mathbf{R}_0=\mathbf{UV}^T
\end{equation}
\retwo{
We provide the discussion in other cases in the supplementary.
}
With the parameter $\mathbf{A}$, we determine the mode of the distribution and present it as the output attitude. 

\textbf{Chordal Mean.}
We define the \textit{chordal mean} \cite{hartley2013rotation,sarlette2009consensus,moakher2002means} of the distribution under the chordal metrics over $\SO$:
\begin{equation}
    \Bar{\mathbf{R}} = \argmin_{\mathbf{R}\in \SO} \int_{\mathbf{R}^* \in \SO} p(\mathbf{R}^*) d_{\text{chord}}(\mathbf{R}, \mathbf{R}^*) \mathrm{d}\mathbf{R}^*
    \label{eq:definition}
\end{equation}
where $d_{\text{chord}}(\cdot, \cdot)$ denotes the L2 chordal metrics of two rotations:
\begin{equation}
    d_{\text{chord}}(\mathbf{R}_1, \mathbf{R}_2) = \| \mathbf{R}_1 - \mathbf{R}_2 \|_F^2
\end{equation}
\retwo{In most cases, i.e., $s_2 + s_3 > 0$,}
the mean of our distribution under the chordal metrics is:
\begin{equation}
    \Bar{\mathbf{R}} = \mathbf{UV}^T
\end{equation}



\textbf{Uncertainty}
The degree of uncertainty of rotation Laplace distribution is measured via $\frac{1}{s_i+s_j}$, as the correlation of our distribution and multivariate Laplace distribution, shown in Prop. \ref{prop:laplace}


\textbf{Shape.}
Following the study of matrix Fisher distribution \cite{lee2018bayesian,mohlin2020probabilistic}, we illustrate the impact of $\mathbf{A}$ on the shape of the rotation Laplace distribution in Figure \ref{fig:vis_shape}.
(a) For a diagonal $\mathbf{A}$, the mode of the distribution is identity and the principal axes correspond to $e_1, e_2, e_3$, where the distribution for each axis is circular and identical. (b) For $\mathbf{A}$ with larger singular values, the distribution are more peaked. (c) The distributions for the $x$- axis are more peaked than the $y$- and $z$- axes, since the first singular value dominates. (d,e,f) The parameter $\mathbf{A}$ is obtained by left-multiplying $\mathbf{T}_i$ and right-multiplying $\mathbf{T}_i^T$, where $\mathbf{T}_i$ refers to the rotation matrix by rotating $\pi/4$ around $e_i$. The mode of the distribution is still identity. However, the principal axes vary according to the columns of $\mathbf{A}$. Thus the orientation of the spread has been affected by the direction of the principal axes.
}

\subsection{Negative Log-likelihood Loss}
Given a collection of observations $\mathcal{X}=\{\boldsymbol{x}_i\}$ and the associated ground truth rotations $\mathcal{R}=\{\mathbf{R}_i\}$, we aim at training the network to best estimate the parameter $\mathbf{A}$ of rotation Laplace distribution. This is achieved by maximizing a likelihood function so that, under our probabilistic model, the observed data is most probable, which is known as maximum likelihood estimation (MLE).
We use the negative log-likelihood of $\mathbf{R}_{\boldsymbol{x}}$ as the loss function:
\begin{equation}
\label{eq:nll_loss}
    \mathcal{L}(\boldsymbol{x}, \mathbf{R}_{\boldsymbol{x}}) = -\log p\left( \mathbf{R}_{\boldsymbol{x}};\mathbf{A}_{\boldsymbol{x}} \right)
\end{equation}

\revision{
\subsection{Normalization Factor}
In this section, we present the normalization factor as a one-dimensional integration form and then introduce its approximation during regression tasks.

Denote $s_1, s_2, s_3$ as the singular values of $\mathbf{A}$, 
we define $t_1=2(s_2+s_3),t_2=2(s_1+s_3),t_3=2(s_1+s_2)$.
The normalizing factor can be expressed as a one-dimensional integration:
    \begin{equation}
    \footnotesize
    \begin{aligned}
        F(\mathbf{A})=&\frac{2}{\pi}\left(\int_{t_1}^{t_2}\frac{\left(\mathbf{L}_{-1}(\sqrt{k})-\mathbf{I}_{1}(\sqrt{k})\right)\mathbf{K}\left(\frac{(k-t_1)(t_3-t_2)}{(t_1-t_2)(k-t_3)}\right)}{\sqrt{k(t_2-t_1)(t_3-k)}}\mathrm{d}k \right.\\
        &\left.+ \int_{t_2}^{t_3}\frac{\left(\mathbf{L}_{-1}(\sqrt{k})-\mathbf{I}_{1}(\sqrt{k})\right)\mathbf{K}\left(\frac{(k-t_3)(t_1-t_2)}{(t_3-t_2)(k-t_1)}\right)}{\sqrt{k(t_2-t_3)(t_1-k)}}\mathrm{d}k\right)
    \end{aligned}
    \end{equation}
where $\mathbf{L}$ is the modified Struve function, $\mathbf{I}$ is the modified Bessel function of the first kind, and $\mathbf{K}$ is the complete elliptic integral of the first kind.

Specifically, when $s_1 = s_2 = s_3 = s$, it can be simplified as 
\begin{equation}
    F(\mathbf{A})=\frac{\mathbf{L}_{-1}(2\sqrt{s})-\mathbf{I}_1(2\sqrt{s})}{\sqrt{s}}
\end{equation}

Please refer to supplementary for the proofs.

Efficiently and accurately estimating the normalization factor for our distribution is non-trivial. 
}
Inspired by \cite{murphy2021implicit}, we approximate the normalization factor of rotation Laplace distribution through equivolumetric discretization over $\SO$ manifold. We employ the discretization method introduced in \cite{yershova2010generating}, which starts with the equal area grids on the 2-sphere \cite{gorski2005healpix} and covers $\SO$ by threading a great circle through each point on the surface of a 2-sphere with Hopf fibration.
\ree{Concretely, we discretize $\SO$ space into a finite set of equivolumetric grids $\mathcal{G}=\left\{\mathbf{R}|\mathbf{R}\in \SO\right\}$},
the normalization factor of Laplace Rotation distribution is computed as
{\footnotesize
\begin{equation*}
\begin{aligned}
    F(\mathbf{A}) &= \int_{\SO} \frac{\exp\left(-\sqrt{\operatorname{tr}\left(\mathbf{S} - \mathbf{A}^T \mathbf{R}\right)}\right)}
    {\sqrt{\operatorname{tr}\left(\mathbf{S} -\mathbf{A}^T \mathbf{R}\right)}} \mathrm{d}\mathbf{R}\\ 
    &\approx 
    \ree{
    \sum_{\mathbf{R}_i \in \mathcal{G}} \frac{\exp\left(-\sqrt{\operatorname{tr}\left(\mathbf{S} - \mathbf{A}^T \mathbf{R}_i\right)}\right)}
    {\sqrt{\operatorname{tr}\left(\mathbf{S} -\mathbf{A}^T \mathbf{R}_i\right)}} \Delta \mathbf{R}_i 
    }
\end{aligned}
\end{equation*}
}\ree{where $\Delta \mathbf{R}_i=
\frac{\int_{SO(3)} \mathrm{d} \mathbf{R}}{|\mathcal{G}|}=\frac{1}{|\mathcal{G}|}$}.
\ree{We set $|\mathcal{G}|$ as about 37k in experiments.}

\ree{
To avoid online computations (which is especially useful for devices without GPUs), we also provide a lookup table of normalization factors w.r.t. the proper singular values $\mathbf{S}$ for both the forward and backward pass. One can then apply trilinear interpolation to obtain the factor and gradient for the query singular values. This technique is also used in \cite{gilitschenski2019deep,deng2022deep}. Please refer to supplementary for more details.
}



\retwo{
\subsection{Singularities}
Similar to the multivariate Laplace distribution in Euclidean space, rotation Laplace distribution in $\SO$ manifold is also not smooth near the mode. This intrinsic singularity holds for both distributions.

The distribution suffers from singularity around the mode in two aspects according to Prop. \ref{prop:laplace}.
On one hand, when $\mathbf{S}$ is close to zero, the distribution becomes ill-defined. This property leads to difficulties in fitting distributions with very large uncertainties, such as the uniform distribution. On the other hand, when the predicted mode is close to the ground truth, the probability density tends to become infinite, which may result in unstable training. Although we employ a probability density function clipping strategy to alleviate the phenomenon, the unstable training still occurs when we try to fit a Dirac distribution.
We include a detailed experimental analysis for these two singularities in Section \ref{sec:exp_singularity}. 

Specifically, probability density function clipping strategy refers to the clipping of $\operatorname{tr}(\mathbf{S} - \mathbf{A}^T\mathbf{R})$ in the denominator by $\max(\epsilon, \operatorname{tr}(\mathbf{S} - \mathbf{A}^T\mathbf{R}))$, where $\epsilon=1e-8$. The strategy is effective to avoid numerical issues in most regression tasks. For a detailed analysis of the robustness of this clipping parameter, we conduct an ablation study in Section \ref{sec:ablation_clip}.

}





\subsection{Quaternion Laplace Distribution}



In this section, we introduce our extension of Laplace-inspired distribution for quaternions, namely, quaternion Laplace distribution.

\begin{definition} \textit{\emph{Quaternion Laplace distribution.}}
The random variable $\mathbf{q}\in \mathcal{S}^3$ follows quaternion Laplace distribution with parameter $\mathbf{M}$ and $\mathbf{Z}$, if its probability density function is defined as
\begin{equation}
    p(\mathbf{q}; \mathbf{M}, \mathbf{Z}) = \frac{1}{F(\mathbf{Z})}\frac{\exp
    \left(-\sqrt{-\mathbf{q}^T \mathbf{M} \mathbf{Z} \mathbf{M}^T \mathbf{q} }\right)}
    {\sqrt{- \mathbf{q}^T \mathbf{M} \mathbf{Z} \mathbf{M}^T \mathbf{q}}}
\end{equation}
where $\mathbf{M}\in \mathbf{O}(4)$ is a $4\times4$ orthogonal matrix, and $\mathbf{Z}=\operatorname{diag}(0, z_1, z_2, z_3)$ is a $4\times4$ diagonal matrix with $0\ge z_1\ge z_2\ge z_3$. We also denote quaternion Laplace distribution as $\mathbf{q} \sim \mathcal{QL}(\mathbf{M}, \mathbf{Z}).$
\end{definition}

\begin{proposition}
\label{prop:quat}
Denote $\mathbf{q}_0$ as the mode of quaternion Laplace distribution. Let $\pi$ be the tangent space of $\mathbb{S}^3$  at $\mathbf{q}_0$, and $\pi(\mathbf{x}) \in \mathbb{R}^4$ be the projection of $\mathbf{x} \in \mathbb{R}^4$ on $\pi$.
For quaternion $\mathbf{q} \in \mathbb{S}^3$ following \emph{Bingham distribution} / \emph{quaternion Laplace distribution}, when $\mathbf{q}\rightarrow\mathbf{q}_0$, $\pi(\mathbf{q})$ follows zero-mean \emph{multivariate Gaussian distribution} / zero-mean \emph{multivariate Laplace distribution}.
\end{proposition}
Both Bingham distribution and quaternion Laplace distribution exhibit antipodal symmetry on $\mathcal{S}^3$, i.e., $p(\mathbf{q}) = p(-\mathbf{q})$, which captures the nature that the quaternions $\mathbf{q}$ and $-\mathbf{q}$ represent the same rotation on $\SO$.

\begin{proposition}
\label{prop:eq}
Denote $\gamma$ as the standard transformation from unit quaternions to corresponding rotation matrices. For rotation matrix $\mathbf{R}\in \SO$ following \emph{rotation Laplace distribution}, $\mathbf{q}=\gamma^{-1}(\mathbf{R})\in \mathbb{S}^3$ follows \emph{quaternion Laplace distribution}.
\end{proposition}

Prop. \ref{prop:eq} shows that our proposed rotation Laplace distribution is equivalent to quaternion Laplace distribution, similar to the equivalence of matrix Fisher distribution and Bingham distribution \cite{prentice1986orientation}, demonstrating the consistency of our derivations.
Please see supplementary for the proofs to the above propositions.

The normalization factor of quaternion Laplace distribution is also approximated by dense discretization, and a pre-computed lookup table is used to avoid online computations.
{\footnotesize\begin{equation*}
\begin{aligned}
    F(\mathbf{Z}) &= \oint_{\mathcal{S}^3} \frac{\exp
    \left(-\sqrt{-\mathbf{q}^T \mathbf{M} \mathbf{Z} \mathbf{M}^T \mathbf{q} }\right)}
    {\sqrt{- \mathbf{q}^T \mathbf{M} \mathbf{Z} \mathbf{M}^T \mathbf{q}}}
    \mathrm{d}\mathbf{q} \\
    &\approx 
    \ree{
    \sum_{\mathbf{q}_i \in \mathcal{G}_\mathbf{q}} \frac{\exp
    \left(-\sqrt{-\mathbf{q}_i^T \mathbf{M} \mathbf{Z} \mathbf{M}^T \mathbf{q}_i }\right)}
    {\sqrt{- \mathbf{q}_i^T \mathbf{M} \mathbf{Z} \mathbf{M}^T \mathbf{q}_i}}
    \Delta \mathbf{q}_i 
    }
\end{aligned}    
\end{equation*}
}\ree{where $\mathcal{G}_\mathbf{q}=\left\{ \mathbf{q} | \mathbf{q}\in \mathcal{S}^3 \right\}$ 
denotes the set of equivolumetric grids and $\Delta \mathbf{q}_i=\frac{\oint_{\mathcal{S}^3} \mathrm{d}\mathbf{q}}{|\mathcal{G}_\mathbf{q}|} = \frac{2\pi^2}{|\mathcal{G}_\mathbf{q}|}$.}

\section{Experiment with single prediction}
\label{sec:exp}

\ree{
Following the previous state-of-the-arts \cite{murphy2021implicit,mohlin2020probabilistic}, we evaluate our method on the task of object rotation estimation from single RGB images, where object rotation is the relative rotation between the input object and the object in the canonical pose.
}
Concerning this task,
we find two kinds of independent research tracks with slightly different evaluation settings. One line of research focuses on probabilistic rotation regression with different parametric or non-parametric distributions on $\SO$ \cite{prokudin2018deep,gilitschenski2019deep,deng2022deep,mohlin2020probabilistic,murphy2021implicit}, and the other non-probabilistic track proposes multiple rotation representations \cite{zhou2019continuity,levinson2020analysis,peretroukhin2020smooth} or improves the gradient of backpropagation \cite{chen2022projective}.
To fully demonstrate the capacity of our rotation Laplace distribution, we leave the baselines in their original optimal states and adapt our method to follow the common experimental settings in each track, respectively. 


\subsection{Datasets \& Evaluation Metrics}

\subsubsection{Datasets}
\textit{ModelNet10-SO3} \cite{liao2019spherical} is a commonly used synthetic dataset for single image rotation estimation containing 10 object classes. It is synthesized by rendering the CAD models of ModelNet-10 dataset \cite{wu20153d} that are rotated by uniformly sampled rotations in $\SO$.
\textit{Pascal3D+} \cite{xiang2014beyond} is a popular benchmark on real-world images for pose estimation. It covers 12 common daily object categories. The images in Pascal3D+ dataset are sourced from Pascal VOC and ImageNet datasets, and are split into ImageNet\_train, ImageNet\_val, PascalVOC\_train, and PascalVOC\_val sets.

\subsubsection{Evaluation metrics}
We evaluate our experiments with the geodesic distance of the network prediction and the ground truth. This metric returns the angular error and we measure it in degrees. In addition, we report the prediction accuracy within the given error threshold.


\subsection{Comparisons with Probabilistic Methods}
\label{exp:track1}

\subsubsection{Evaluation Setup}
\textbf{Settings.} In this section, we follow the experiment settings of the latest work \cite{murphy2021implicit} and quote its reported numbers for baselines. Specifically, we train one single model for all categories of each dataset. For Pascal3D+ dataset, we follow \cite{murphy2021implicit} to use (the more challenging) PascalVOC\_val as test set. 
Note that \cite{murphy2021implicit} only measure the coarse-scale accuracy (e.g., Acc@30$^\circ$) which may not adequately satisfy the downstream tasks \cite{wang2019normalized,fang2020towards}. To facilitate finer-scale comparisons (e.g., Acc@5$^\circ$), we further re-run several recent baselines and report the reproduced results in parentheses ($\cdot$).


\textbf{Baselines.}
We compare our method to recent works which utilize probabilistic distributions on $\SO$ for the purpose of pose estimation. 
In concrete, the baselines are with mixture of \textit{von Mises} distributions \cite{prokudin2018deep}, \textit{Bingham} distribution \cite{gilitschenski2019deep,deng2022deep}, \textit{matrix Fisher} distribution \cite{mohlin2020probabilistic} and Implicit-PDF \cite{murphy2021implicit}.
We also compare to the spherical regression work of \cite{liao2019spherical} 
as \cite{murphy2021implicit} does.


\subsubsection{Results}
Table \ref{tab:modelnet} shows the quantitative comparisons of our method and baselines on ModelNet10-SO3 dataset. From the multiple evaluation metrics, we can see that maximum likelihood estimation with the assumption of rotation Laplace distribution significantly outperforms the other distributions for rotation, including matrix Fisher distribution \cite{mohlin2020probabilistic}, Bingham distribution \cite{do2018deep} and von-Mises distribution \cite{prokudin2018deep}. Our method also gets superior performance than the non-parametric implicit-PDF \cite{murphy2021implicit}. Especially, our method improves the fine-scale Acc@3$^\circ$ and Acc@5$^\circ$ accuracy by a large margin, showing its capacity to precisely model the target distribution.

\begin{table*}[t]
  \centering
  \fontsize{7.5}{9}\selectfont
  \vspace{0mm}
  \caption{\small Numerical comparisons with probabilistic baselines on ModelNet10-SO3 dataset averaged on all categories. Numbers in parentheses ($\cdot$) are our reproduced results. Please refer to supplementary for comparisons with each category.}
    \begin{tabular}{lcccccc}
    \toprule
          & Acc@3$^\circ$$\uparrow$ & Acc@5$^\circ$$\uparrow$ & Acc@10$^\circ$$\uparrow$ & Acc@15$^\circ$$\uparrow$ & Acc@30$^\circ$$\uparrow$ & Med.($^\circ$)$\downarrow$  \\
    \midrule
    
    Liao \textit{et al.}\cite{liao2019spherical}        &      -&   -&      -&  0.496&  0.658&  28.7  \\
    Prokudin \textit{et al.}\cite{prokudin2018deep}         &      -&   -&      -&  0.456&  0.528&   49.3 \\
    Deng \textit{et al.}\cite{deng2022deep}             &  (0.138)&   (0.301)&  (0.502)&  0.562 (0.584)&  0.694 (0.673)&   32.6 (31.6)\\
    Mohlin \textit{et al.}\cite{mohlin2020probabilistic}  &  (0.164)& (0.389) &  (0.615) &  0.693 (0.684)&  0.757 (0.751)&   17.1 (17.9) \\
    Murphy \textit{et al.}\cite{murphy2021implicit}       &  (0.294)&  (0.534)  & (0.680) &  0.719 (0.714)&  0.735 (0.730)&   21.5 (20.3)\\
    \midrule
    quaternion Laplace              &  0.242  &  0.469 &  0.647  &  0.692  & 0.741  &  18.3   \\
    rotation Laplace                &  \textbf{0.445}  & \textbf{0.611}  &  \textbf{0.716}  &  \textbf{0.742}  &  \textbf{0.771} & \textbf{13.0}   \\
    \bottomrule
    \end{tabular}%
  \label{tab:modelnet}%
\end{table*}%

\begin{table*}[t]
  \centering
  \fontsize{7.5}{9}\selectfont
  \caption{\small Numerical comparisons with probabilistic baselines on Pascal3D+ dataset averaged on all categories. Numbers in parentheses ($\cdot$) are our reproduced results. Please refer to supplementary for comparisons with each category.}
  \vspace{2mm}
    \begin{tabular}{lcccccc}
    \toprule
          & Acc@3$^\circ$$\uparrow$ & Acc@5$^\circ$$\uparrow$ & Acc@10$^\circ$$\uparrow$ & Acc@15$^\circ$$\uparrow$ & Acc@30$^\circ$$\uparrow$ & Med.($^\circ$)$\downarrow$   \\
    \midrule
    
    Tulsiani \& Malik\cite{tulsiani2015viewpoints}   &      -&      -&       -&      -&  0.808&  13.6 \\
    Mahendran \textit{et al.}\cite{mahendran2018mixed}       &      -&      -&      -&       -&  0.859&  10.1 \\
    Liao \textit{et al.}\cite{liao2019spherical}        &      -&      -&      -&       -&  0.819&  13.0 \\
    Prokudin \textit{et al.}\cite{prokudin2018deep}         &      -&      -&      -&       -&  0.838&  12.2 \\
    Mohlin \textit{et al.}\cite{mohlin2020probabilistic}  &  (0.089)&  (0.215)&    (0.484)& (0.650)&  0.825 (0.827)&  11.5 (11.9) \\
    Murphy \textit{et al.}\cite{murphy2021implicit}       &  (0.102)&   (0.242)&  (0.524)& (0.672)&  0.837 (0.838)&  10.3 (10.2) \\
    \midrule
    quaternion Laplace              & 0.125   & 0.252  &   0.502 & 0.648   & 0.834  &   11.4  \\
    rotation Laplace                 &  \textbf{0.134} & \textbf{0.292} & \textbf{0.574}  &  \textbf{0.714}  & \textbf{0.874} &  \textbf{9.3}  \\
    \bottomrule
    \end{tabular}%
  \label{tab:pascal}%
\end{table*}%

The experiments on Pascal3D+ dataset are shown in Table \ref{tab:pascal}, where our rotation Laplace distribution outperforms all the baselines. While our method gets reasonably good performance on the median error and coarser-scale accuracy, we do not find a similar impressive improvement on fine-scale metrics as in ModelNet10-SO3 dataset. 
We suspect it is because the imperfect human annotations of real-world images may lead to comparatively noisy ground truths, increasing the difficulty for networks to get rather close predictions with GT labels.
Nevertheless, our method still manages to obtain superior performance, which illustrates the robustness of our rotation Laplace distribution.




\subsection{Comparisons with Non-probabilistic Methods}
\label{exp:track2}

\subsubsection{Evaluation Setup}
\textbf{Settings.}
For comparisons with non-probabilistic methods, we follow the latest work of \cite{chen2022projective} to learn a network for each category. For Pascal3D+ dataset, we follow \cite{chen2022projective} to use ImageNet\_val as our test set. We use the same evaluation metrics as in \cite{chen2022projective} and quote its reported numbers for baselines.

\textbf{Baselines.}
We compare to multiple baselines that leverage different rotation representations to directly regress the prediction given input images, including 
6D \cite{zhou2019continuity}, 9D / 9D-Inf \cite{levinson2020analysis} and 10D \cite{peretroukhin2020smooth}. We also include regularized projective manifold gradient (RPMG) series of methods \cite{chen2022projective}.

\subsubsection{Results}
We report the numerical results of our method and on-probabilistic baselines on ModelNet10-SO3 dataset in Table \ref{tab:rpmg_modelnet}. Our method obtains a clear superior performance to the best competitor under all the metrics among all the categories. Note that we train a model for each category (so do all the baselines), thus our performance in Table \ref{tab:rpmg_modelnet} is better than Table \ref{tab:modelnet} where one model is trained for the whole dataset.
The results on Pascal3D+ dataset are shown in Table \ref{tab:rmpg_pascal} where our method with rotation Laplace distribution achieves state-of-the-art performance.

\setlength{\tabcolsep}{4pt}
\begin{table*}[t]
    \caption{\small Numerical comparisons with non-probabilistic baselines on ModelNet10-SO3 dataset. One model is trained for each category.}
    \vspace{2mm}
    \centering
    \scriptsize
    \begin{tabular}{l|ccc|ccc|ccc|ccc}
    \toprule
    \multirow{2}{*}{Methods}&& Chair &&&Sofa&&& Toilet&&&Bed&\\
    \cmidrule{2-13}  
        & Mean$\downarrow$ & Med.$\downarrow$ & Acc@5$\uparrow$ 
        & Mean$\downarrow$ & Med.$\downarrow$ & Acc@5$\uparrow$ 
        & Mean$\downarrow$ & Med.$\downarrow$ & Acc@5$\uparrow$ 
        & Mean$\downarrow$ & Med.$\downarrow$ & Acc@5$\uparrow$ 
        \\
        \midrule   
        
        6D & 19.6 & 9.1  & 0.19 & 17.5 & 7.3 & 0.27  &    10.9&6.2&0.37&32.3&11.7&0.11     \\
        9D & 17.5 & 8.3 & 0.23 & 19.8 & 7.6 & 0.25  &    11.8&6.5&0.34&30.4&11.1&0.13     \\
        9D-Inf & 12.1 & 5.1 & 0.49& 12.5 & 3.5 & 0.70 & 7.6&3.7&0.67&22.5&4.5&0.56     \\
        10D & 18.4 & 9.0 & 0.20 & 20.9 & 8.7 & 0.20 &  11.5& 5.9&0.39& 29.9&11.5&0.11      \\
        \midrule
        RPMG-6D & 12.9 & 4.7 & 0.53 & 11.5 & 2.8 & 0.77 &      7.8&3.4&0.71&20.3&3.6&0.67     \\
        RPMG-9D & {11.9} & {4.4} & {0.58} & {10.5} & {2.4} & {0.82}  &     7.5&3.2&0.75&20.0&{2.9}&{0.76}    \\
         RPMG-10D & 12.8& 4.5& 0.55 & 11.2&{2.4}&{0.82}& {7.2}&{3.0}&{0.76}&{19.2}&{2.9}&0.75\\
        \midrule
        quat. Laplace & 12.6 & 5.2 & 0.49 & 13.1 & 3.7 & 0.67 & 5.9 & 3.4 & 0.69 & 17.7 & 3.4 & 0.69\\
        rot. Laplace & \textbf{9.7}  & \textbf{3.5}  &  \textbf{0.68} & \textbf{8.8} & \textbf{2.1} & \textbf{0.84}  & \textbf{5.3} & \textbf{2.6} & \textbf{0.83} & \textbf{15.5} & \textbf{2.3} & \textbf{0.82}  \\
        \bottomrule
    \end{tabular}
  \label{tab:rpmg_modelnet}
\end{table*}
\setlength{\tabcolsep}{7.5pt}
\begin{table*}[t]
  \centering
  \scriptsize
      \caption{\small Numerical comparisons with non-probabilistic baselines on Pascal3D+ dataset. One model is trained for each category.}
      \vspace{2mm}
    \begin{tabular}{l|cccc|cccc}
    \toprule
    \multicolumn{1}{c|}{\multirow{2}[4]{*}{Methods}} & \multicolumn{4}{c|}{Bicycle}  & \multicolumn{4}{c}{Sofa} \\
\cmidrule{2-9}          & Acc@10$\uparrow$ & Acc@15$\uparrow$ & Acc@20$\uparrow$ & Med.$\downarrow$  & Acc@10$\uparrow$ & Acc@15$\uparrow$ & Acc@20$\uparrow$ & Med.$\downarrow$ \\
    \midrule
    6D                  &      0.218& 0.390 & 0.553 & 18.1  &  0.508& 0.767 & 0.890 & 9.9                      \\
    9D                  &      0.206 & 0.376 & 0.569&18.0   &  0.524 & 0.796 & 0.903&9.2                       \\
    9D-Inf              &      0.380 & 0.533 & 0.699&13.4   &  0.709 & {0.880} & 0.935&{6.7}     \\
    10D                 &      0.239&0.423&0.567&17.9       &  0.502&0.770&0.896&9.8                           \\
    \midrule
    RPMG-6D             &      0.354 & 0.572 &0.706&13.5                       &    0.696 & 0.861 &0.922&{6.7}                           \\
    RPMG-9D             &      0.368&0.574 & {0.718}&{12.5}      &    {0.725} &{0.880} &{0.958}&{6.7} \\
    RPMG-10D            &      {0.400} & {0.577} & 0.713 & 12.9  &    0.693 & 0.871 & 0.939 & 7.0                                 \\
    \midrule
    quat. Laplace        &   0.398  &  0.559  & 0.686  & 12.2  &   0.615 &   0.769 & 0.923 & 8.4   \\
    rot. Laplace        &  \textbf{0.435}   &   \textbf{0.641}&  \textbf{0.744}&   \textbf{11.2}&   \textbf{0.735}&    \textbf{0.900}&     \textbf{0.964}&  \textbf{6.3} \\
    
    \bottomrule
    \end{tabular}%
  \label{tab:rmpg_pascal}%
\end{table*}%


\subsection{Qualitative Results}
\begin{figure*}[t]
    \centering
    \begin{tabular}{cccccc}
    \includegraphics[width=0.12\linewidth]{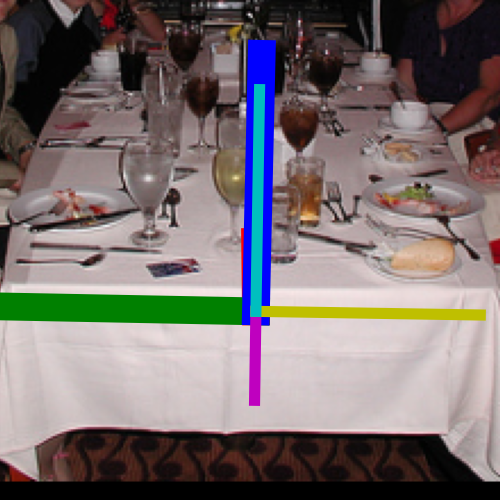}
    &\includegraphics[width=0.12\linewidth]{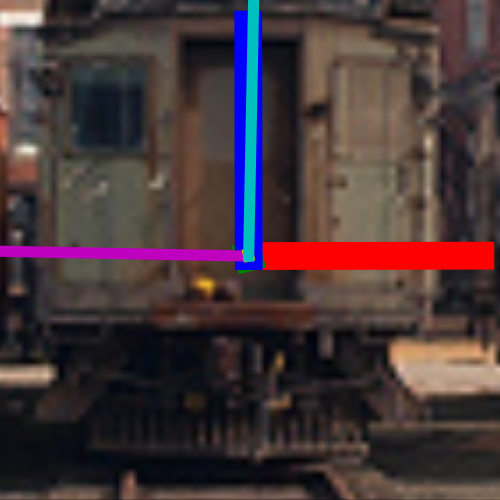}
    &\includegraphics[width=0.12\linewidth]{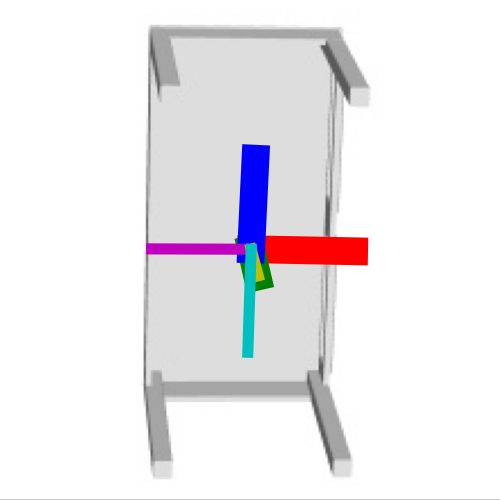}
    &\includegraphics[width=0.12\linewidth]{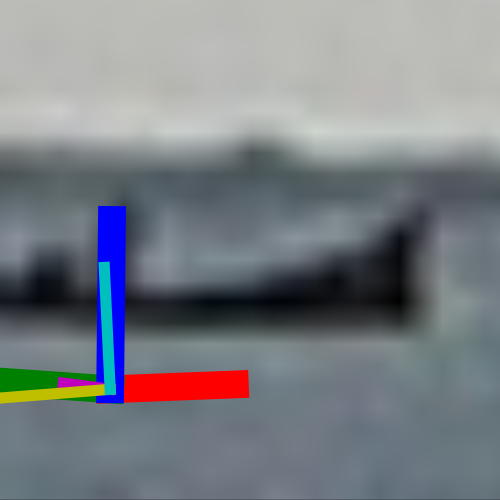}
    &\includegraphics[width=0.12\linewidth]{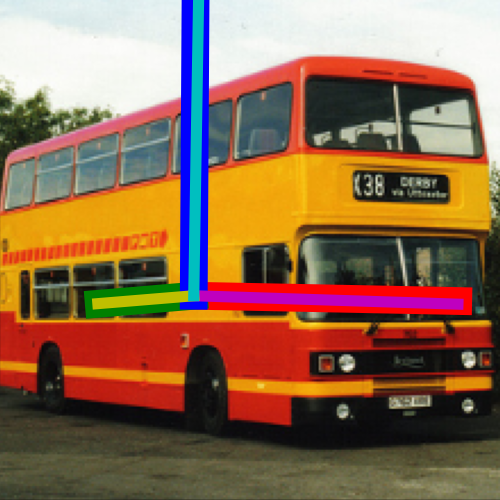}
    &\includegraphics[width=0.12\linewidth]{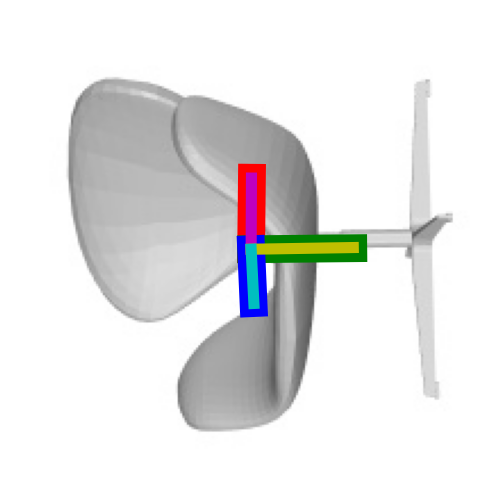}
    \\
    \includegraphics[clip,trim=4.5cm 4cm 4.5cm 1.5cm, width=0.12\linewidth]{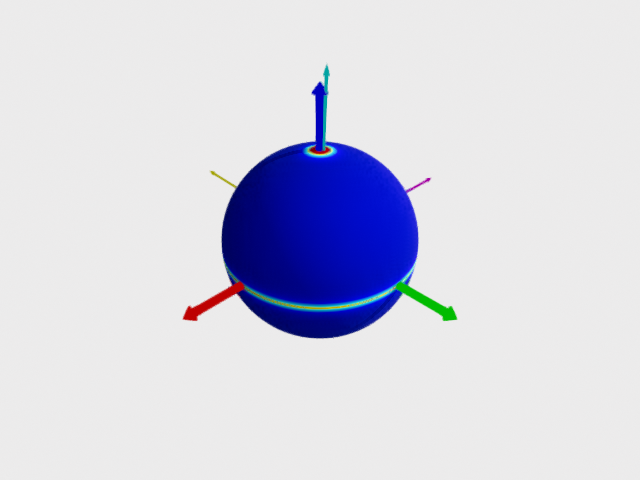}
    &\includegraphics[clip,trim=4.5cm 4cm 4.5cm 1.5cm, width=0.12\linewidth]{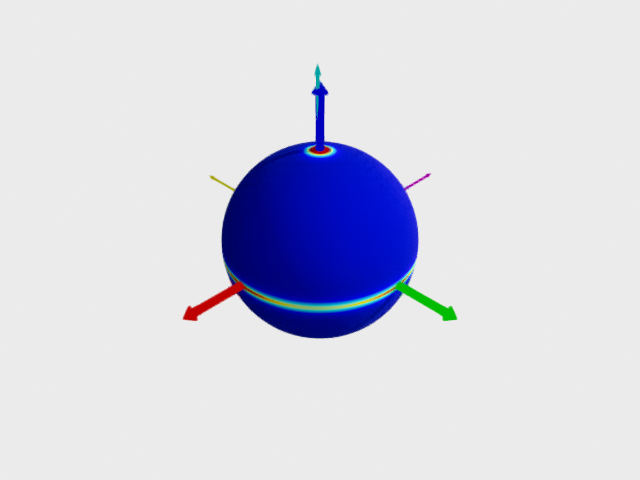}
    &\includegraphics[clip,trim=4.5cm 4cm 4.5cm 1.5cm, width=0.12\linewidth]{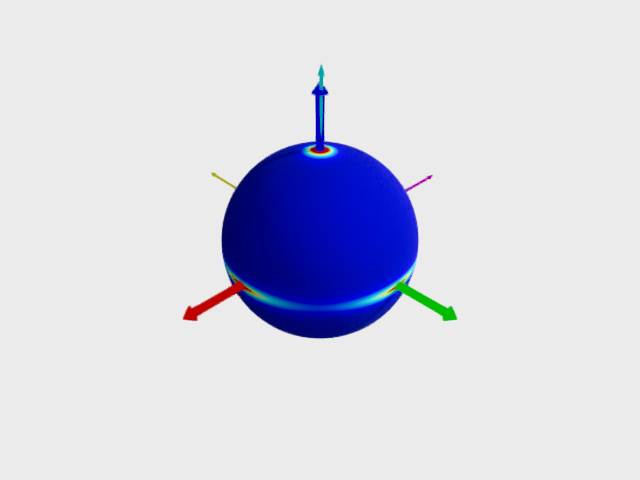}
    &\includegraphics[clip,trim=4.5cm 4cm 4.5cm 1.5cm, width=0.12\linewidth]{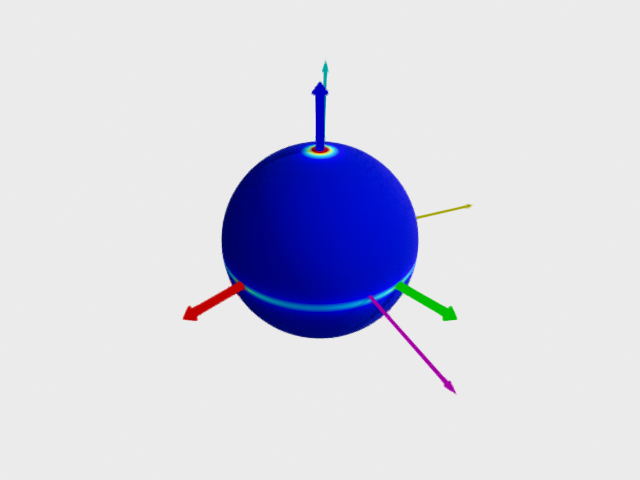}
    &\includegraphics[clip,trim=4.5cm 4cm 4.5cm 1.5cm, width=0.12\linewidth]{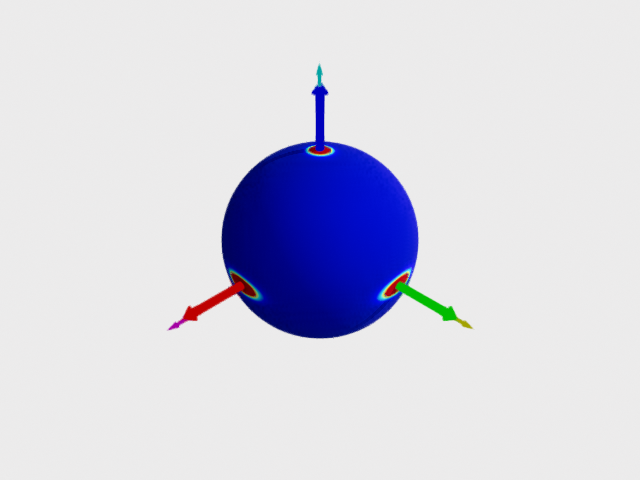}
    &\includegraphics[clip,trim=4.5cm 4cm 4.5cm 1.5cm, width=0.12\linewidth]{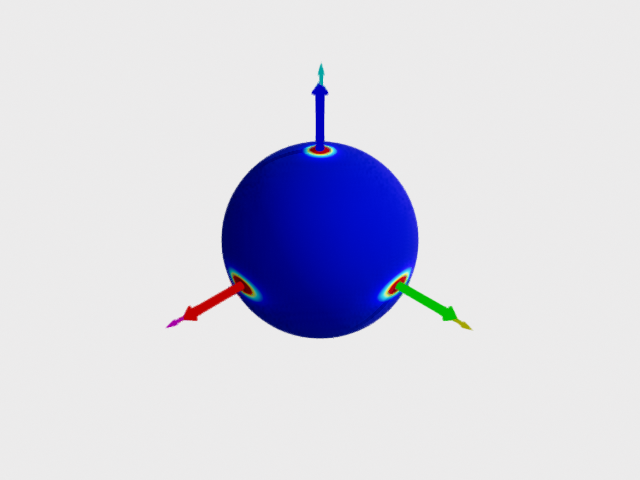}
    \\
    \small{(a)} &\small{(b)} &\small{(c)} &\small{(d)} &\small{(e)} & \small{(f)} 
    \end{tabular}
    \caption{\small \ree{Visualizations of the predicted distributions. The top row displays example images with the projected axes of predictions (thick lines) and ground truths (thin lines) of the object. The bottom row shows the visualization of the corresponding predicted distributions of the image.
    For clarity we have aligned the predicted poses with the standard axes.
    }}
    \vspace{-2mm}
	\label{fig:visual}
\end{figure*}

We visualize the predicted distributions in Figure \ref{fig:visual} with the visualization method in \cite{mohlin2020probabilistic}. The visualization in \cite{mohlin2020probabilistic} is achieved by summing the three marginal distributions over the standard basis of $\mathbb{R}^3$ and displaying them on the sphere with color coding. 
As shown in the figure, the predicted distributions can exhibit high uncertainty when the object has rotational symmetry, leading to near 180$^\circ$ errors (a-c), or the input image is with low resolution (d). Subfigure (e-f) show cases with high certainty and reasonably low errors.

Please refer to supplementary for more visualization results.

\subsection{Implementation Details}

For fair comparisons, we follow the implementation designs of \cite{mohlin2020probabilistic} and merely change the distribution from matrix Fisher distribution to our rotation Laplace distribution.
\ree{Please refer to supplementary for additional experiment details.}





\subsection{\ree{Comparisons of Rotation Laplace Distribution and Quaternion Laplace Distribution}}

\revision{
For completeness, we also experiment with the proposed quaternion Laplace distribution and report the performance in Table \ref{tab:modelnet}, \ref{tab:pascal}, \ref{tab:rpmg_modelnet} and \ref{tab:rmpg_pascal}.
As shown in the tables, quaternion Laplace distribution consistently outperforms its competitor, i.e., Bingham distribution, which validates the effectiveness of our Laplace-inspired derivations.
However, it achieves inferior performance than rotation Laplace distribution. This performance degradation can be attributed to the rotation representation employed, namely quaternion, which has been noted in the literature \cite{zhou2019continuity} for its lack of continuity.
}



\retwo{
\subsection{Experiments on Singularies}
\label{sec:exp_singularity}
\textbf{Singularity 1:} 
    When $\mathbf{S}$ is close to zero, the distribution becomes ill-defined, leading to difficulties in fitting distributions with very large uncertainties, such as the uniform distribution. 
    In our experiment, we compare the matrix Fisher distribution and our rotation Laplace distribution in fitting a uniform distribution on SO(3). Specifically, we sampled rotations uniformly over the SO(3) manifold and optimize the parameters of both distributions via negative log likelihood (NLL) loss. We randomly sample a batch of 16 rotations on the fly and optimize over 10k iterations for each distribution.
    After convergence, we uniformly sample 100k samples as the test set. The test NLL loss for the matrix Fisher distribution is 0.0001, while for the rotation Laplace distribution, it is 0.0677. We further visualize the statistical histogram for the probability density of the test set, as illustrated in Fig. \ref{fig:uniform_fit}. 
    Compared to the matrix Fisher distribution, the rotation Laplace distribution exhibit a biased distribution, indicating its inadequacy in representing a uniform distribution. This result highlights the limitation that the rotation Laplace distribution is unsuitable for data that closely follows a uniform distribution.
    
    \begin{figure}[th]
        \centering
        \begin{tabular}{cc}
        \includegraphics[width=0.45\linewidth]{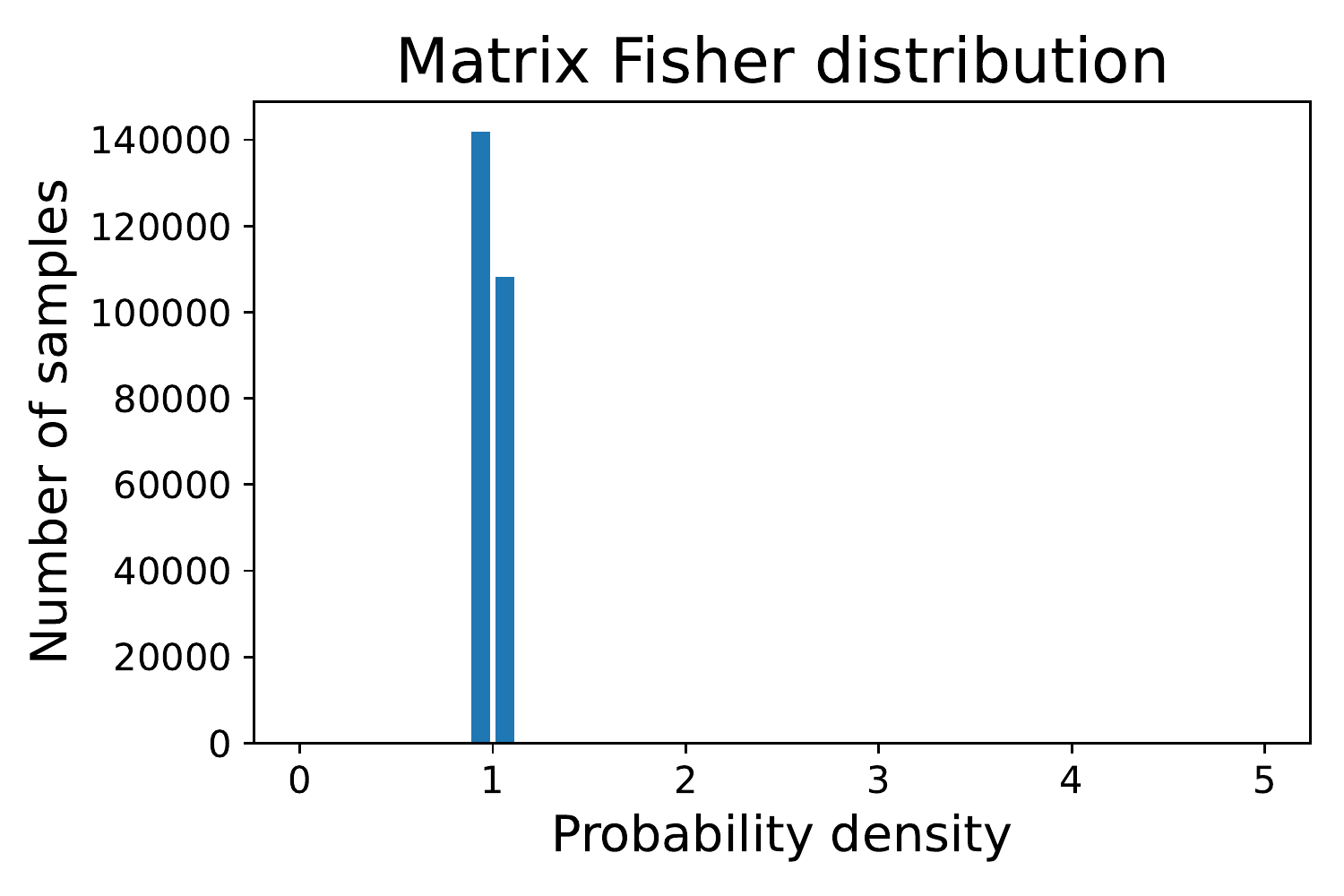} &
        \includegraphics[width=0.45\linewidth]{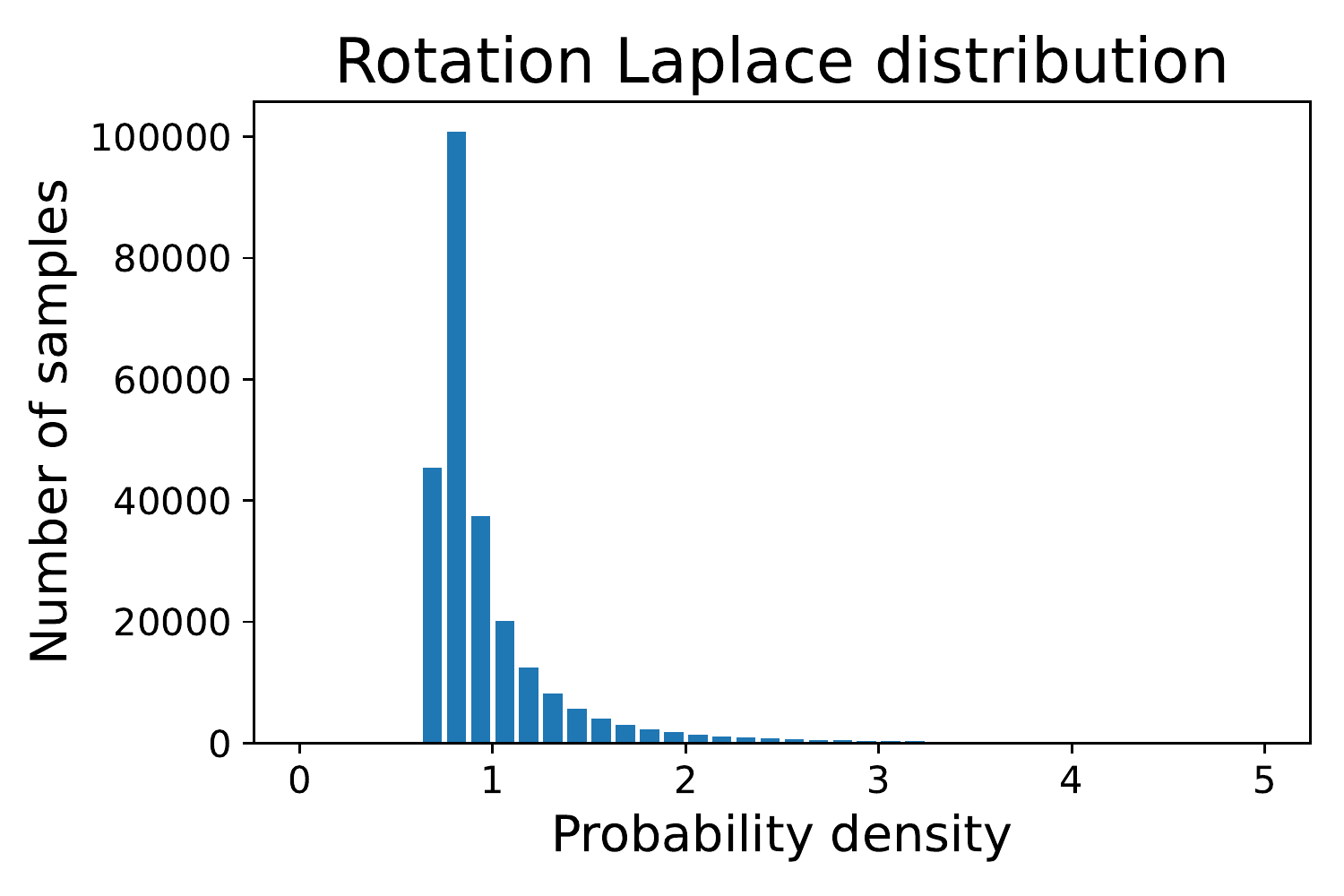}
        \end{tabular}
        \caption{\textbf{Histogram visualization of the probability density of fitting a uniform distribution.} We let the matrix Fisher distribution and rotation Laplace distribution to fit a uniform distribution, and visualize the probability density function after convergence.}
        \label{fig:uniform_fit}
    \end{figure}

    \textbf{Singularity 2:} 
    When the predicted mode is close to the ground truth, the probability density tends to become infinite, leading to numerical issues. We adopt a clipping strategy to address this problem. However, a remaining issue is the large gradient for predictions with tiny errors, which may result in unstable training.
    To highlight this problem, we conduct an experiment where the matrix Fisher distribution and the rotation Laplace distribution are used to optimize a single rotation, or in other words, to fit a Dirac distribution. Specifically, we use NLL loss and the gradient descent optimizer. The clipping parameter is set as $\epsilon=1e-8$. The optimization curves and results are shown in Fig. \ref{fig:dirac}.
    For the rotation Laplace distribution, when the predicted mode $\mathbf{R}_0$ is very close to the ground truth, the probability density function is clipped by $\max(\epsilon, \operatorname{tr}(\textbf{S} - \mathbf{A}^T\mathbf{R}_0))$, causing the loss function to become NaN, the gradient to become zero, and the optimization process to terminate. In contrast, the matrix Fisher distribution can be continuously optimized and theoretically approach the Dirac distribution infinitely.
    Additionally, the optimization process of our rotation Laplace distribution is not smooth. As the predicted mode gets closer to the ground truth, the gradient becomes larger, leading to noticeable oscillations. These results highlight the limitation of our approach, indicating a risk of learning instability.


    However, we notice that the oscillation is not observed in downstream learning tasks such as image-based rotation regression. 
    This absence of oscillation may be attributed to the difficulty of the predicted mode being accurate enough to trigger instability.

    \begin{figure}[t]
        \centering
        \begin{tabular}{@{}ccc@{}}
        \includegraphics[width=0.3\linewidth]{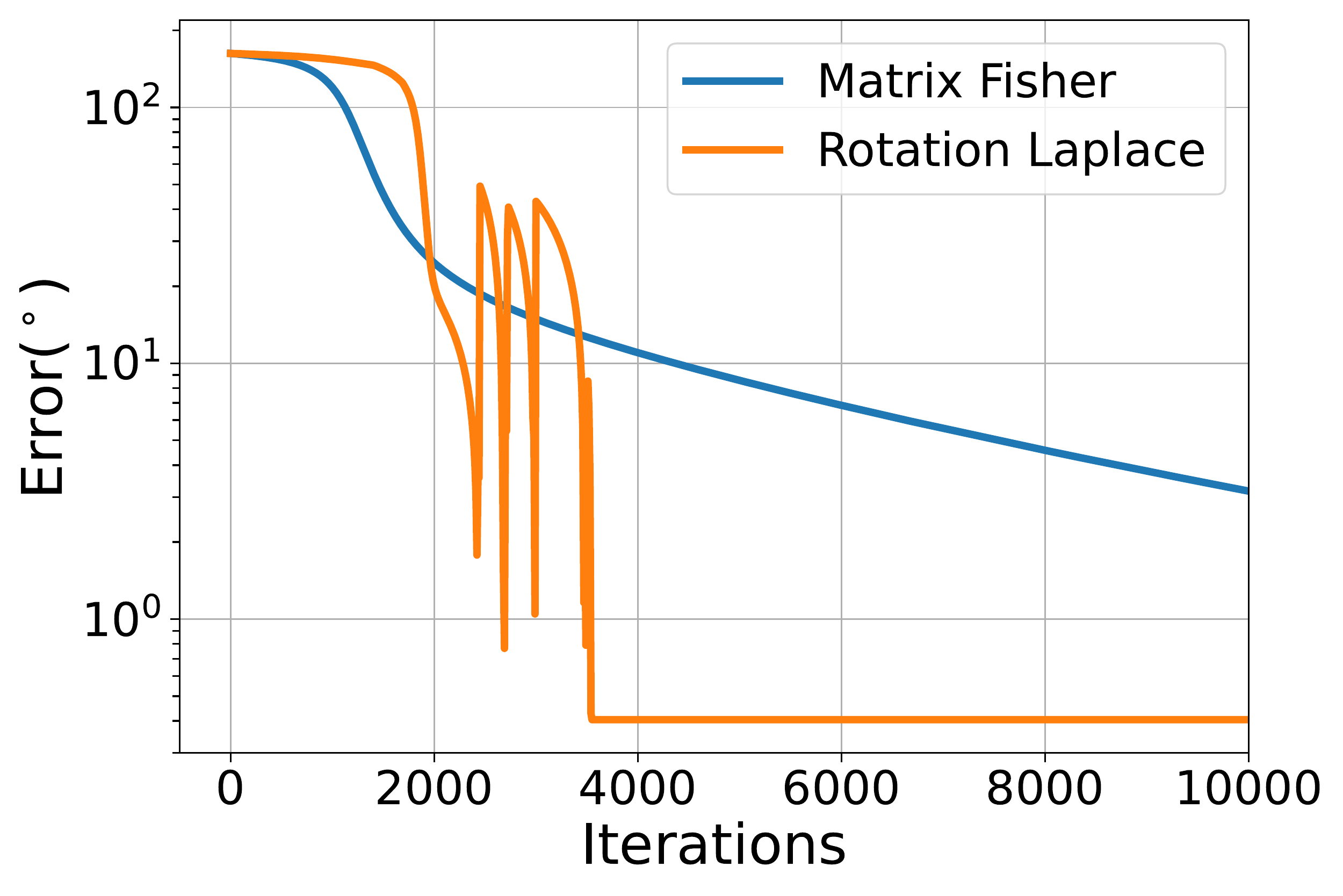} &
        \includegraphics[width=0.3\linewidth]{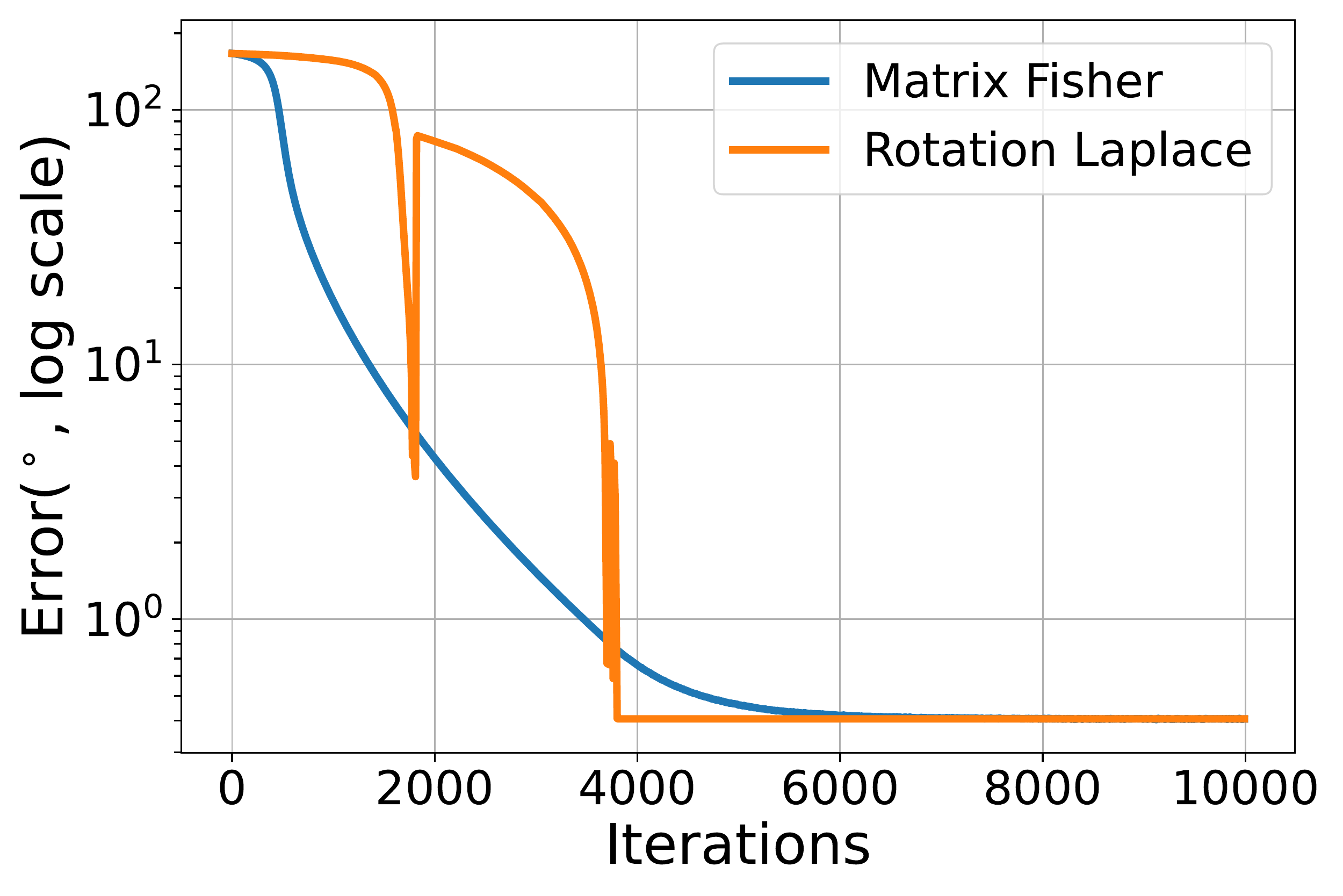} & 
        \includegraphics[width=0.3\linewidth]{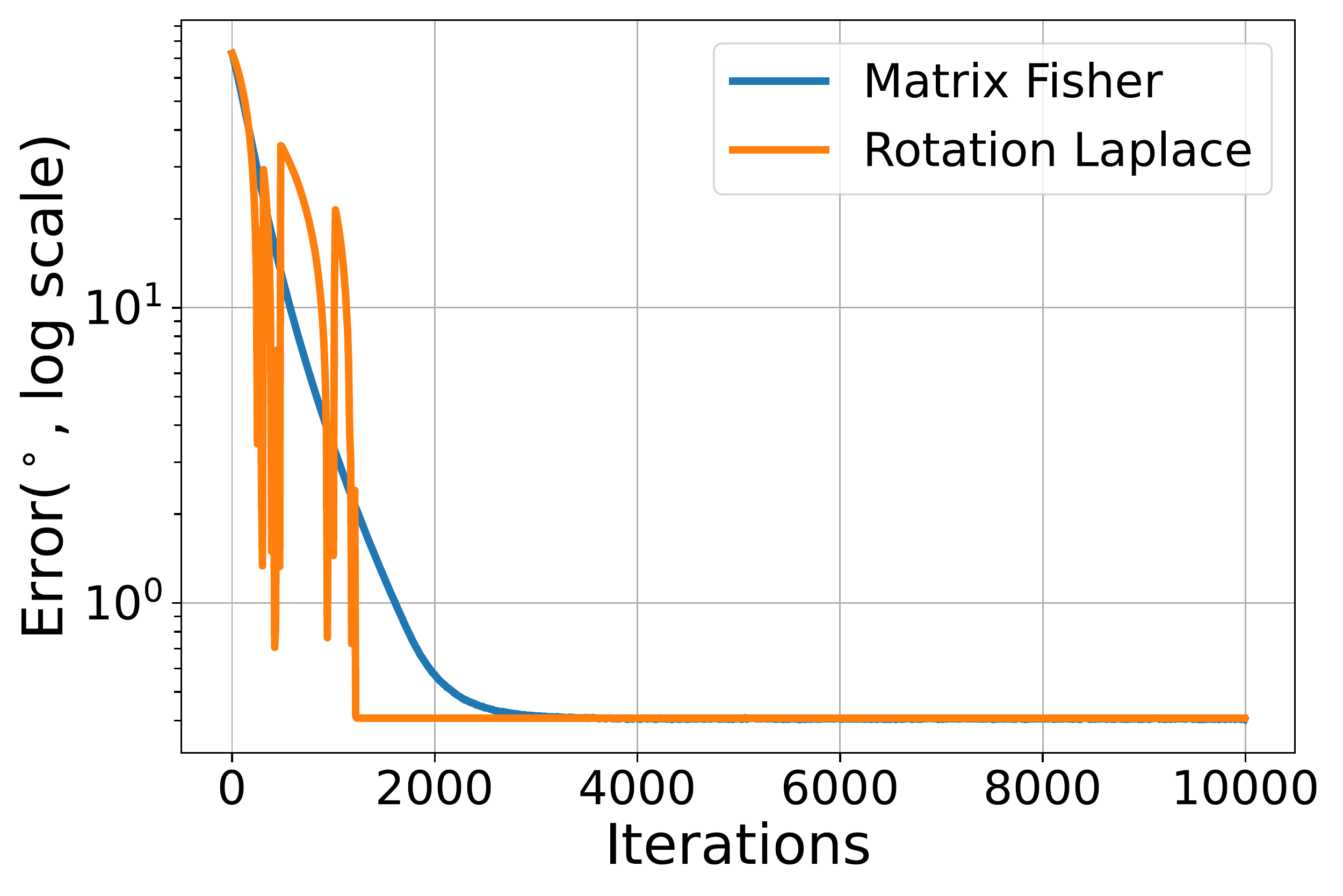} \\
        \includegraphics[width=0.3\linewidth]{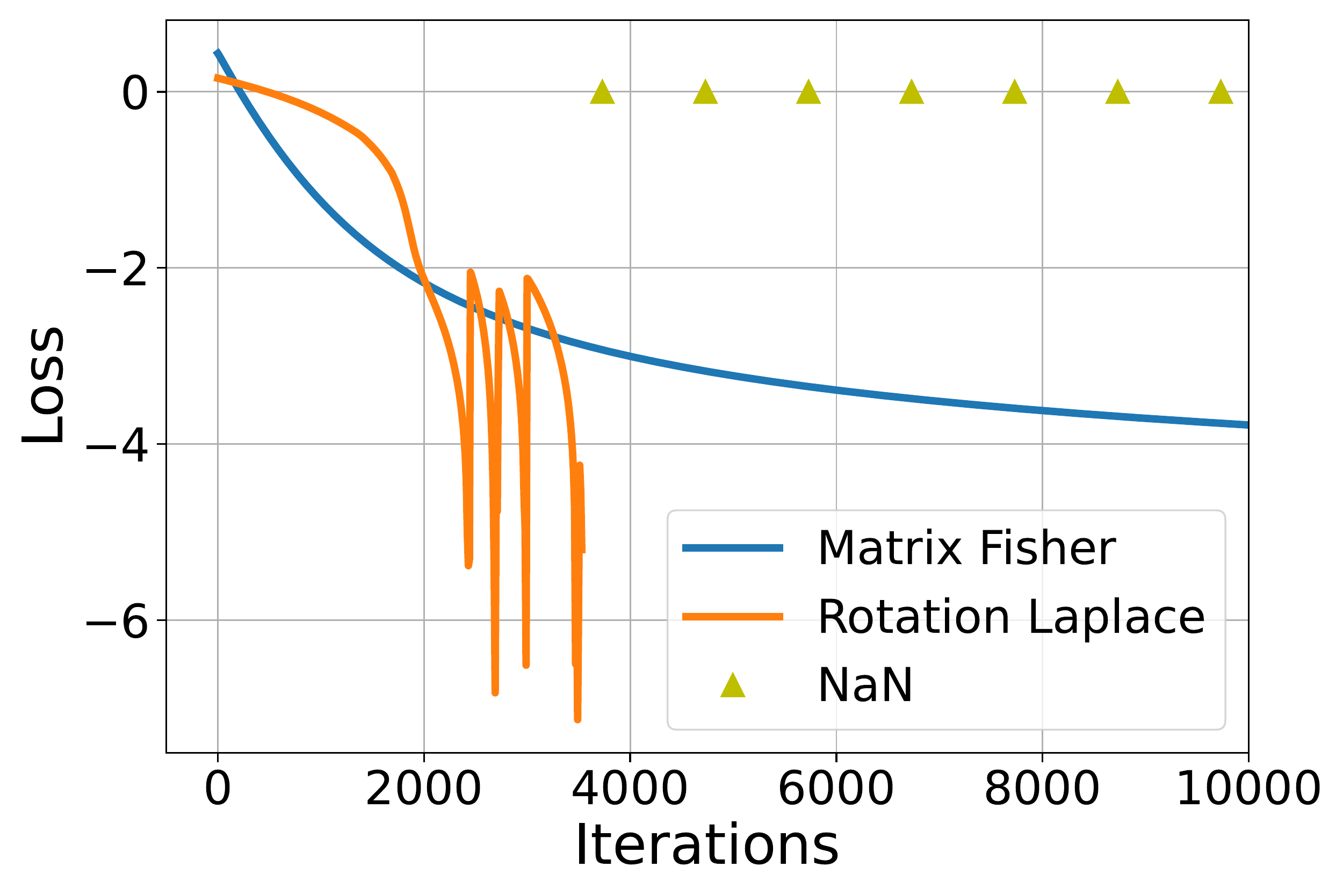} &
        \includegraphics[width=0.3\linewidth]{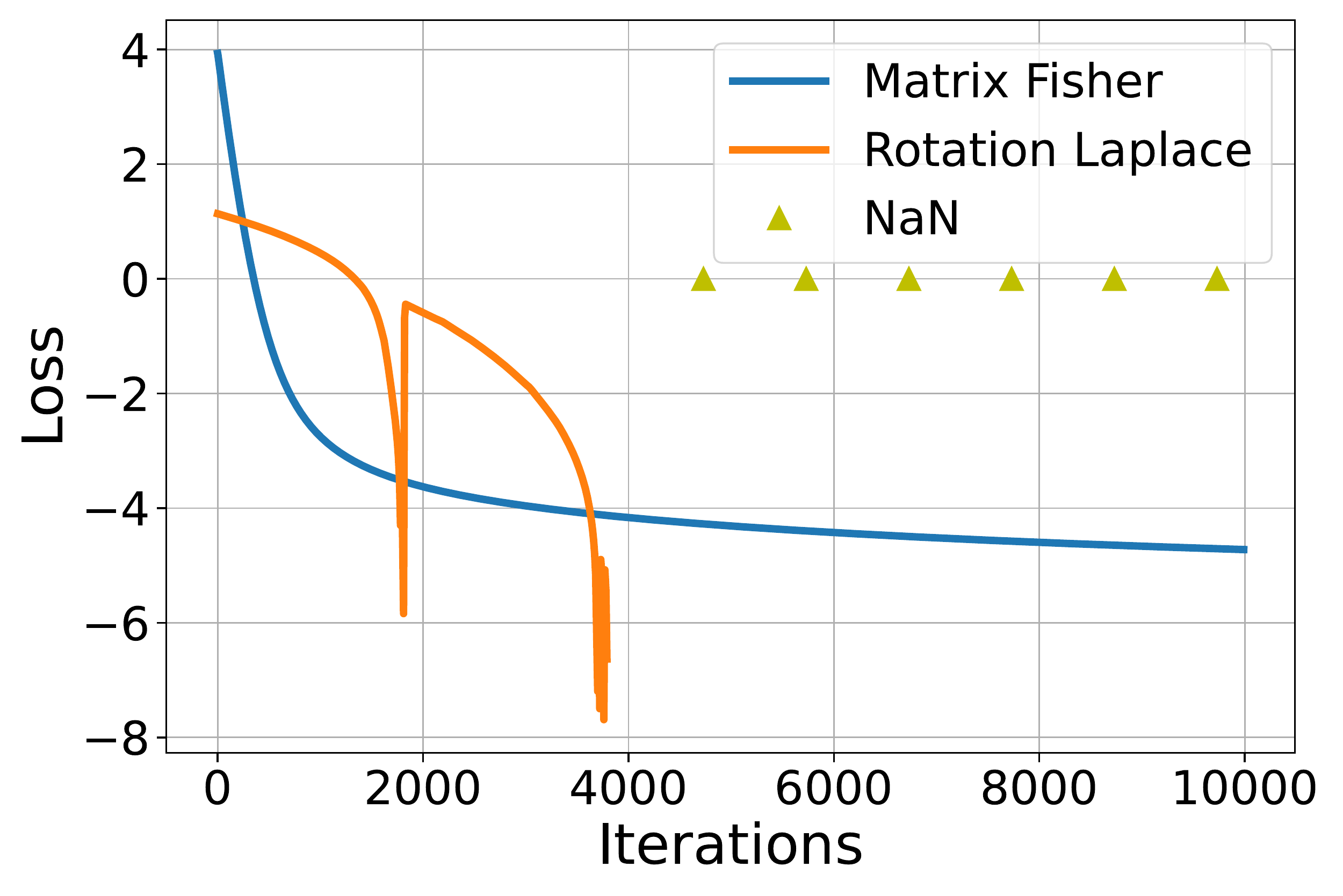} & 
        \includegraphics[width=0.3\linewidth]{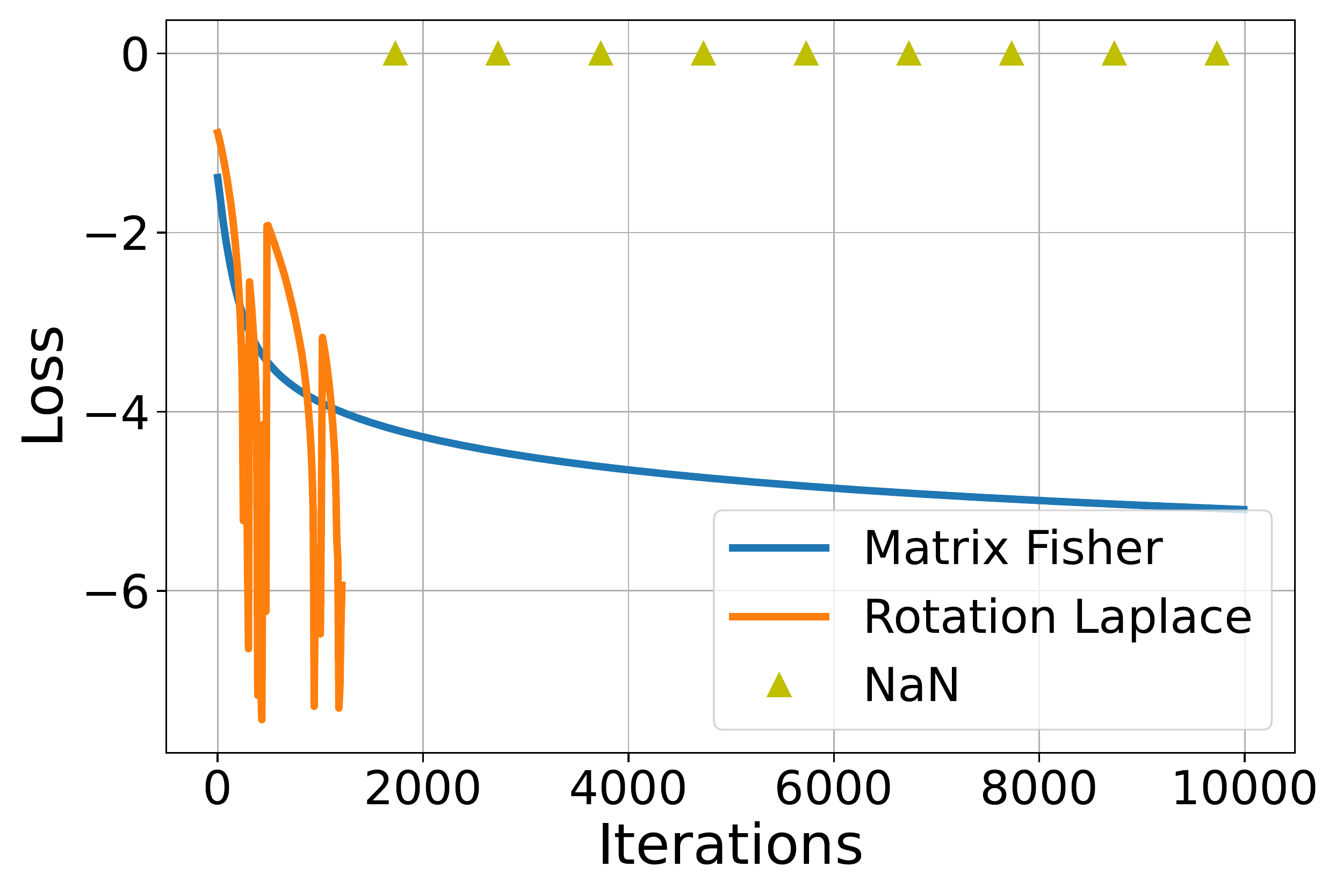} \\
        lr = 1e-4  & lr = 5e-4  & lr = 1e-3 \\        
        \end{tabular}
        \caption{\textbf{Visualization of the optimization curves of fitting a Dirac distribution.} We let the matrix Fisher distribution and rotation Laplace distribution to fit a Dirac distribution and plot the training curves. 
        }
        \label{fig:dirac}
    \end{figure}

}

\begin{figure}[t]
    \centering
    \begin{tabular}{@{}ccc@{}}
    \includegraphics[width=0.3\linewidth]{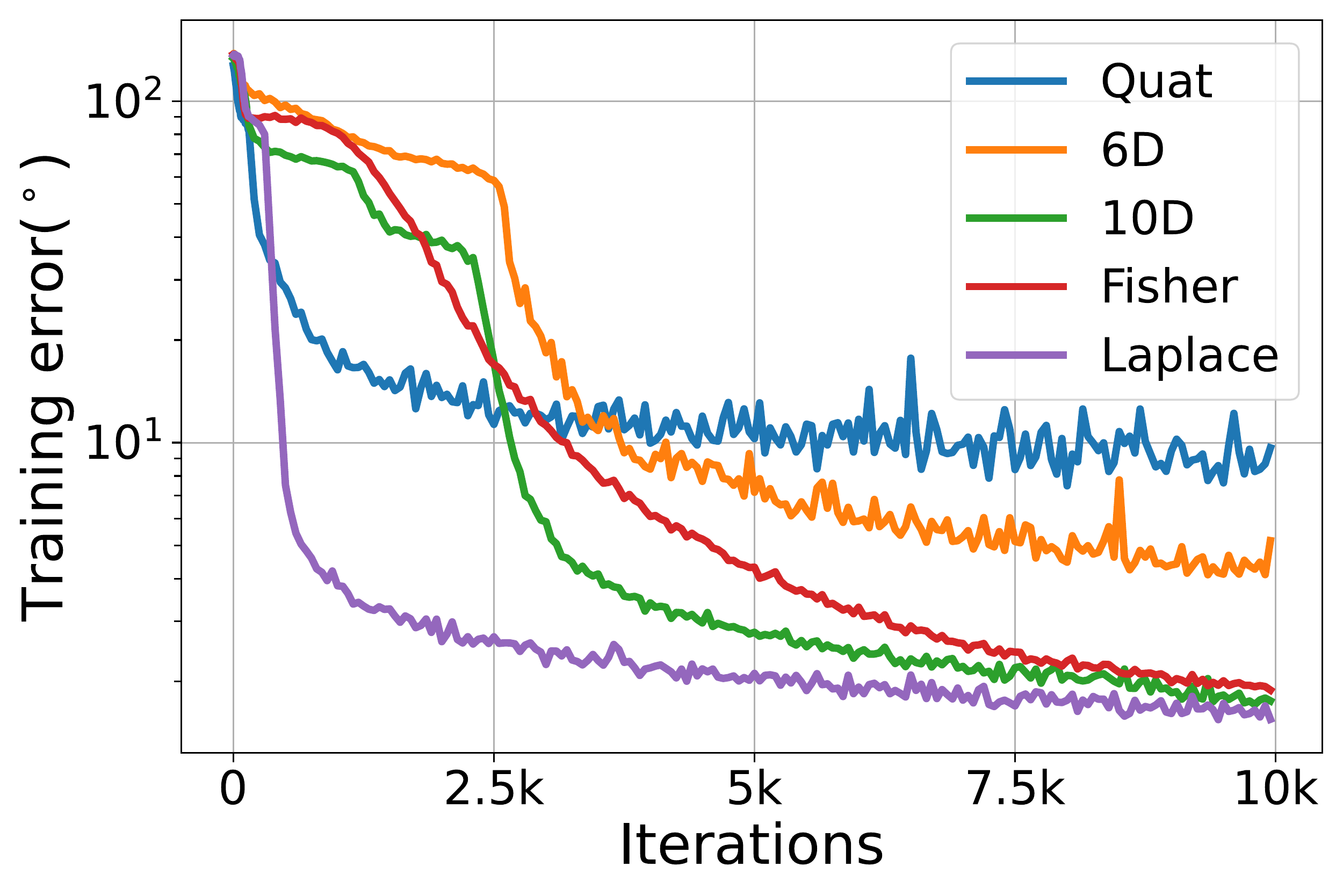} 
    & \includegraphics[width=0.3\linewidth]{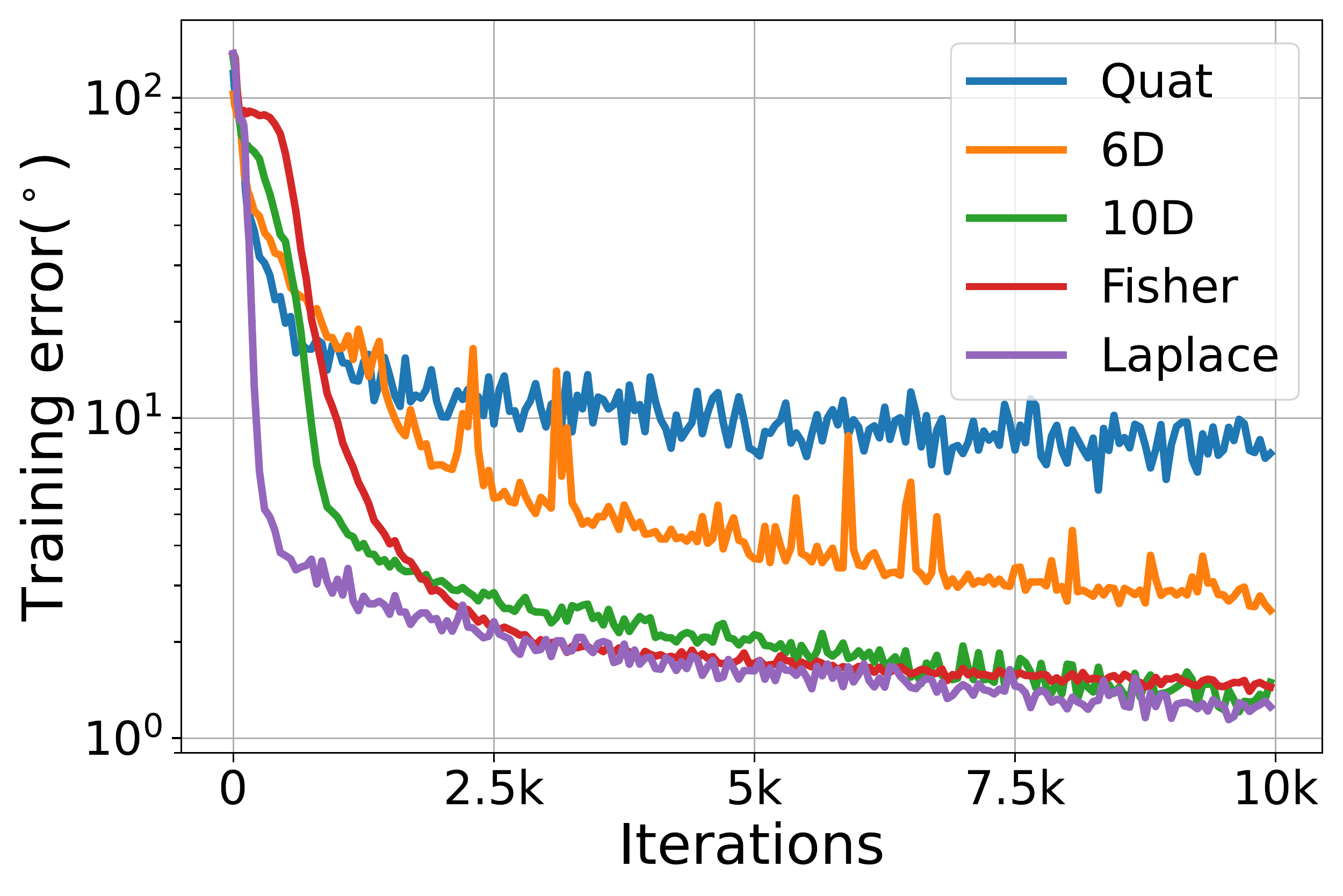} 
    & \includegraphics[width=0.3\linewidth]{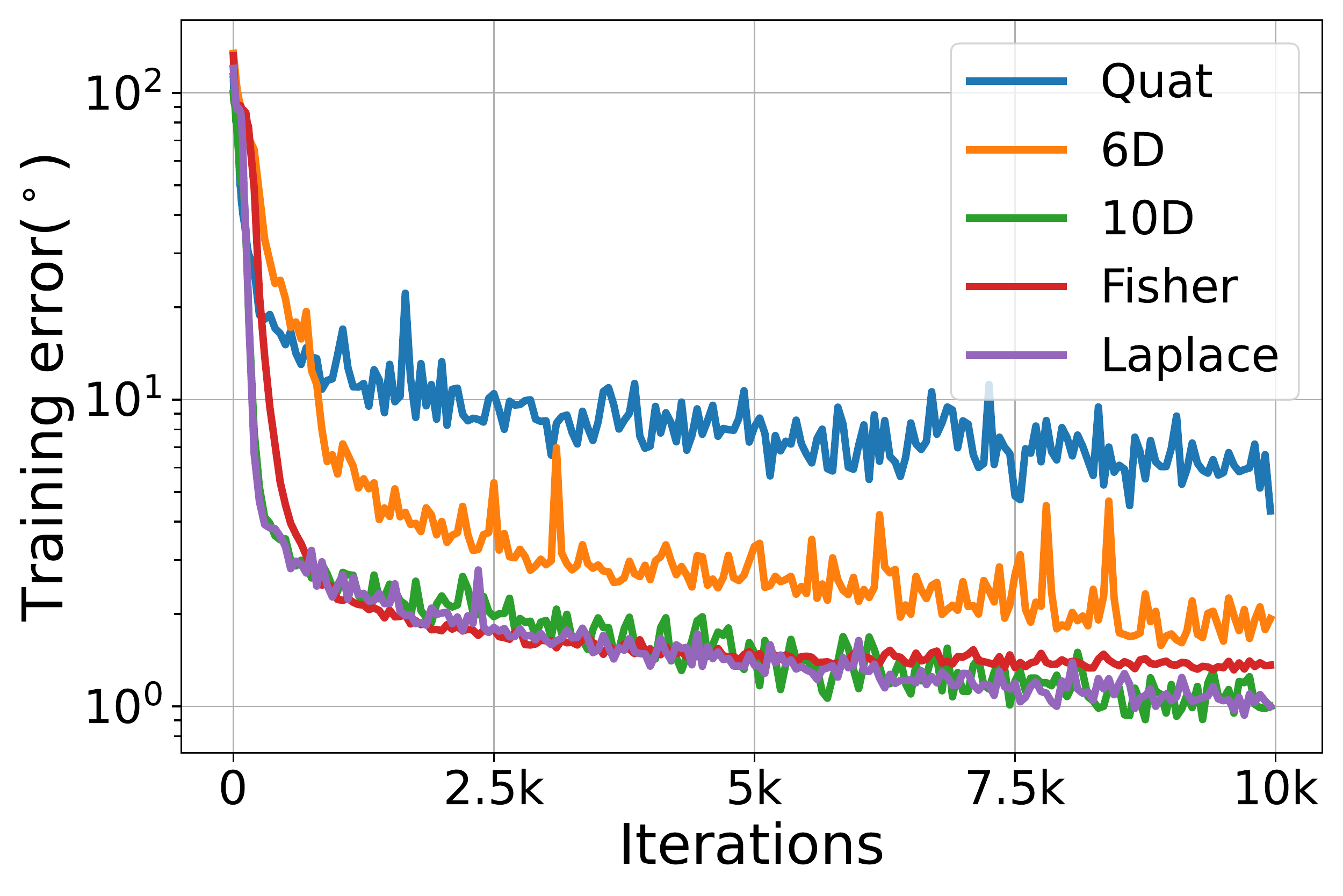} \\
    
    \includegraphics[width=0.3\linewidth]{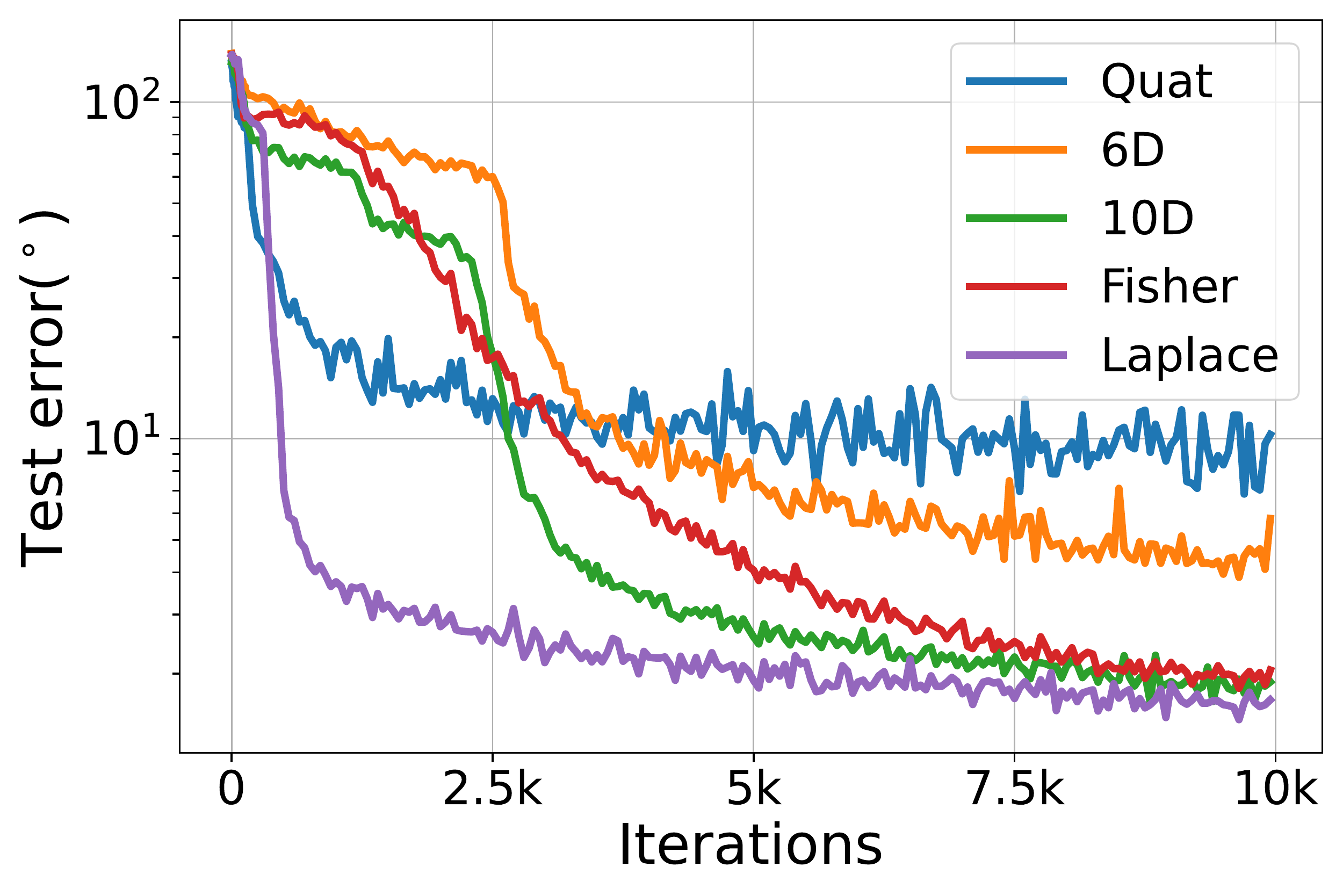}
    & \includegraphics[width=0.3\linewidth]{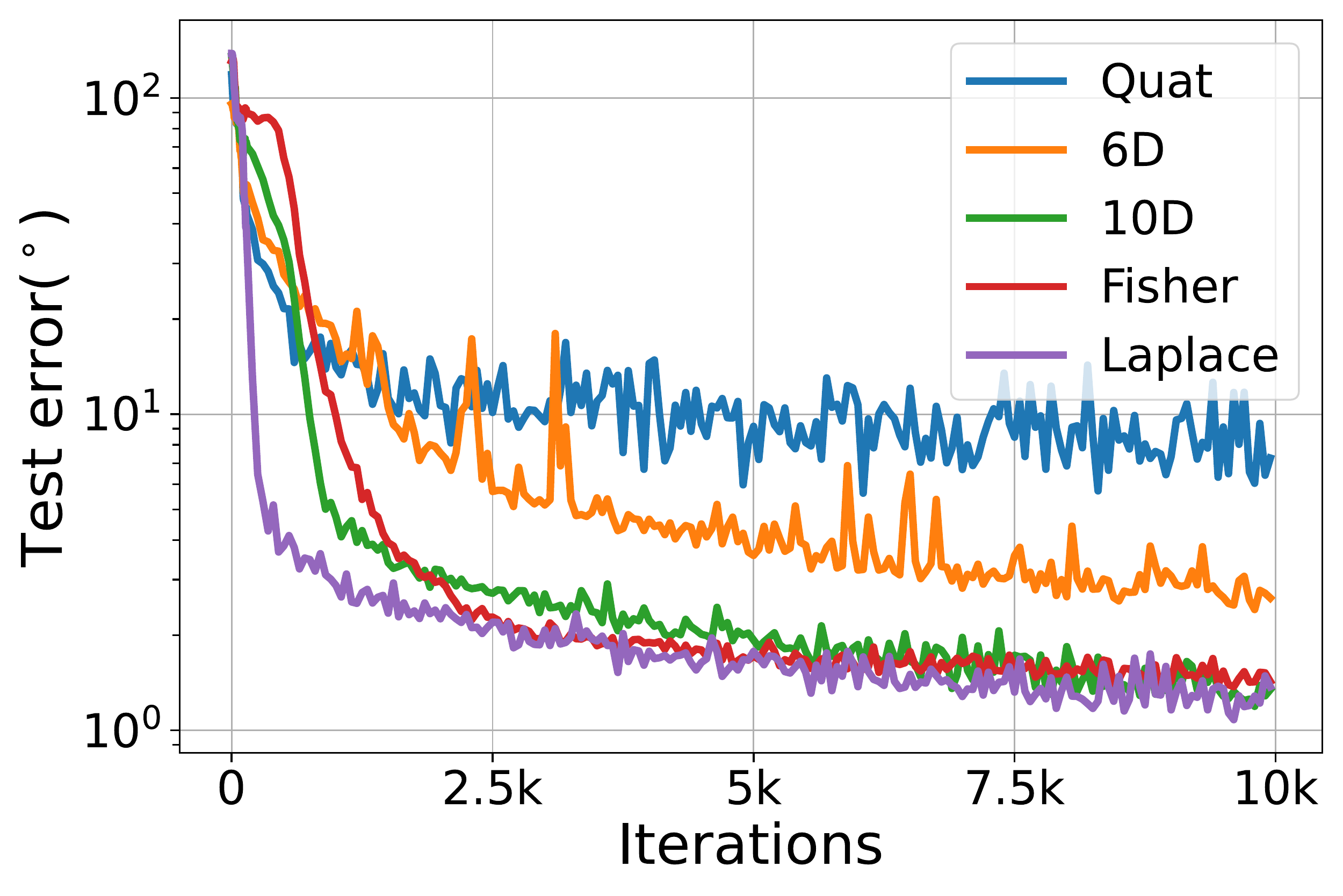} 
    & \includegraphics[width=0.3\linewidth]{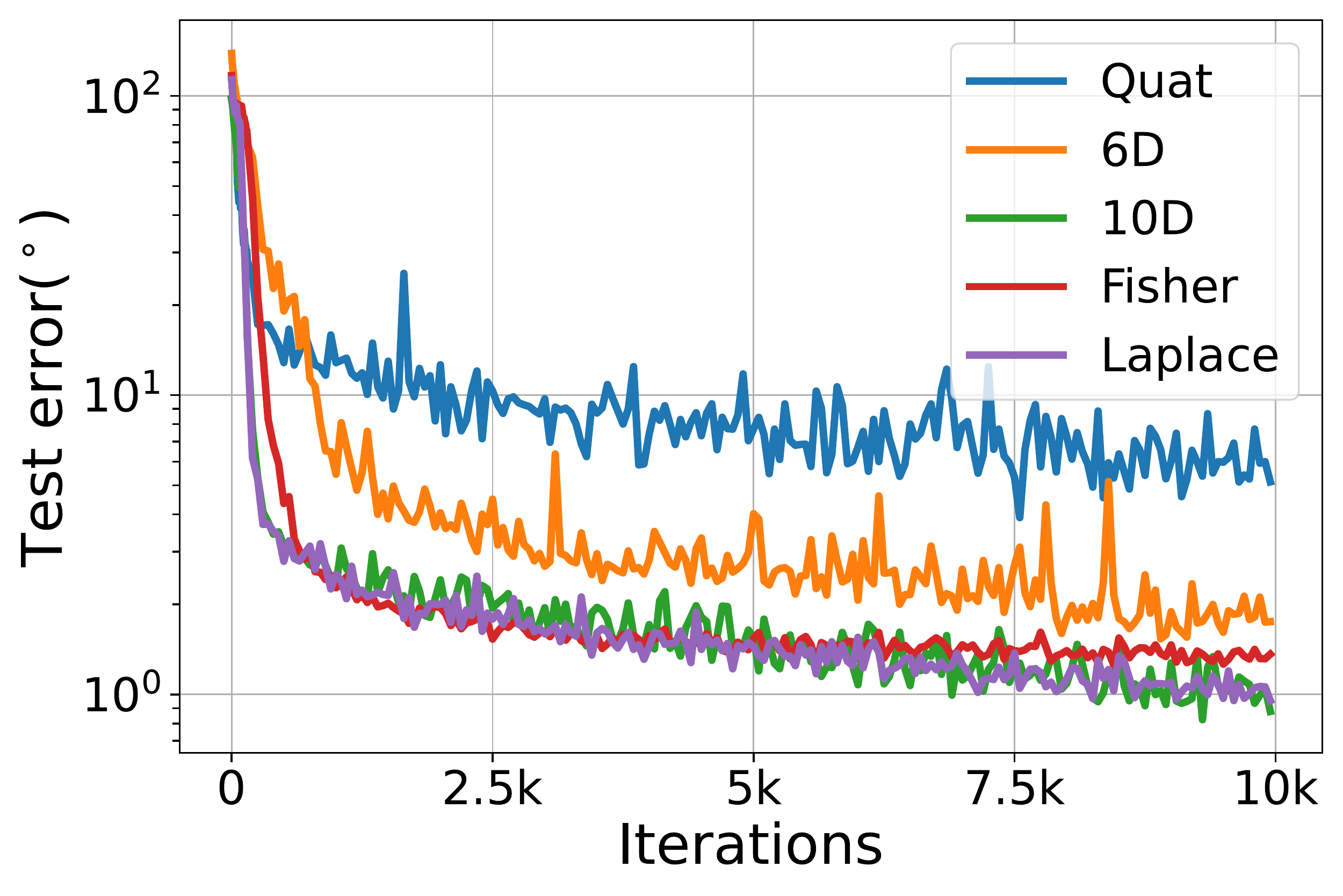}
    \\
    \small{lr=1e-5} & \small{lr=3e-5}  &  \small{lr=1e-4}
    \end{tabular}
    \vspace{-2mm}
    \caption{\small \textbf{Wahba's Problem: Visualization of the training and test error along with the training process.} We compare our method with different rotation representations and distributions. The experiments are conducted with different learning rates.}
    \vspace{-2mm}
	\label{fig:wahba}
\end{figure}

\revision{
\subsection{Ablation Study on the Number of Quantizing Grids}

To investigate the impact of the computation of the normalization factor, we conduct an ablation study on the number of quantizing grids using the ModelNet10-SO(3) toilet category dataset, shown in Table \ref{tab:num_sample}. As outlined in \cite{murphy2021implicit}, the number of grids starts at 72 and iteratively grows by a factor of 8, resulting in $72 \times 8^l$ samples. We explore values for $l$ ranging from 0 to 4. The forward and backward time is measured as the time taken for one pass of a batch of 32 instances on a single 3090 GPU.

\begin{table*}[h]
    \centering
    \footnotesize
    \caption{ \textbf{Ablation studies on the impact of the number of samples for the normalization factor.} The experiments are on ModelNet10-SO(3) toilet category dataset.}
    \begin{tabular}{lccccc}
        \toprule
        \#grids & Mean ($^\circ$)$\downarrow$ & Median ($^\circ$)$\downarrow$ & Acc@5$^\circ$$\uparrow$ & Forward time (ms) & Backward time (ms) \\
        \midrule
        72 $(l=0)$      & 6.0 & 3.2 & 0.76 & 2.3 & 0.7 \\
        576 $(l=1)$     & 5.8 & 2.8 & 0.80 & 2.3 & 0.7 \\
        4608 $(l=2)$    & 5.3 & 2.6 & 0.82 & 2.7 & 0.8 \\
        36864 $(l=3)$   & 5.2 & 2.6 & 0.82 & 5.1 & 1.7 \\
        294912 $(l=4)$  & 5.3 & 2.5 & 0.82 & 26.4 & 12.9 \\
        \bottomrule
    \end{tabular}
    \label{tab:num_sample}
\end{table*}

Table \ref{tab:num_sample} illustrates that using too few grid samples leads to inferior performance, and increasing the number of samples results in better performance at the cost of longer runtime. The performance improvement saturates when $l \ge 2$. After considering both performance and computational efficiency, we choose 37K samples as a balanced option.

\subsection{Ablation Study on the Clipping Parameter}
\label{sec:ablation_clip}
We apply a probability density function clipping strategy to address the numerical issues of our distribution in regression tasks.
To assess the robustness of our clipping strategy, we examine the impact of different values for the clipping parameter $\epsilon$. As illustrated in Table \ref{tab:ablation_clip} and Figure \ref{fig:ablation_clip}, our method exhibits similar performance and training processes with different $\epsilon$ values, without encountering numerical issues. These experiments underscore the effectiveness and robustness of our strategy when applying the rotation Laplace distribution to regression tasks.

\begin{table*}[h]
    \centering
    \footnotesize
    \caption{\textbf{Ablation study on the clipping parameter $\epsilon$.} The clipping parameter does not affect the results. The experiments are on ModelNet10-SO3 dataset averaged on all categories.}
    \begin{tabular}{ccccccc}
        \toprule
        $\epsilon$ & Acc@3$^\circ$$\uparrow$ & Acc@5$^\circ$$\uparrow$ & Acc@10$^\circ$$\uparrow$ & Acc@15$^\circ$$\uparrow$ & Acc@30$^\circ$$\uparrow$ & Med.($^\circ$)$\downarrow$  \\
        \midrule
        $1\mathrm{e}{-4}$      & 0.450 & 0.614 & 0.718 & 0.743 & 0.769 & 12.8 \\
        $1\mathrm{e}{-8}$      & 0.445 & 0.611 & 0.716 & 0.742 & 0.771 & 13.0 \\
        $1\mathrm{e}{-12}$      & 0.441 & 0.608 & 0.714 & 0.742 & 0.769 & 12.5 \\
        \bottomrule
    \end{tabular}
    \label{tab:ablation_clip}
\end{table*}

\begin{figure}[t]
    \centering
    \begin{tabular}{@{}ccc@{}}
    \includegraphics[width=0.3\linewidth]{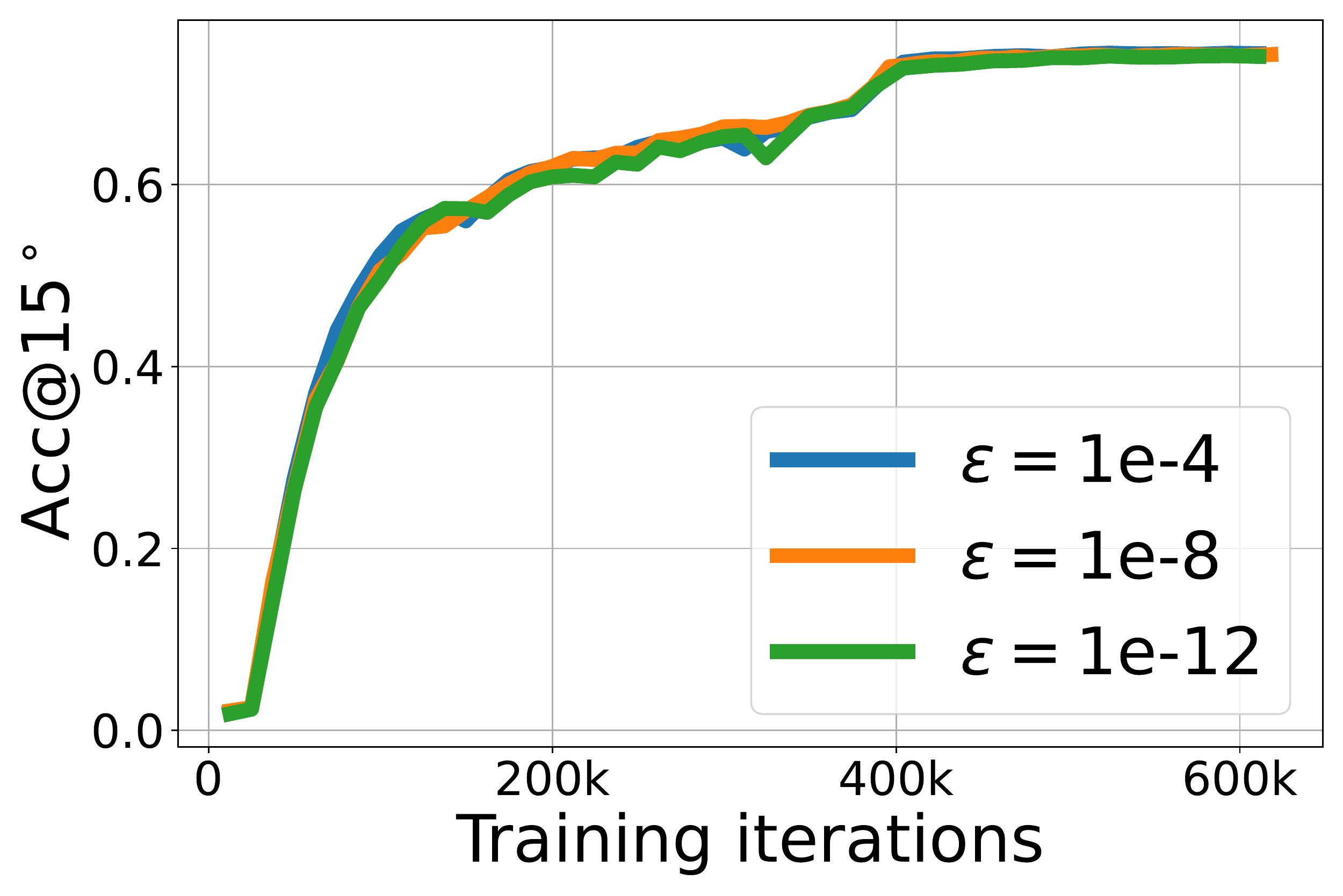}
    &\includegraphics[width=0.3\linewidth]{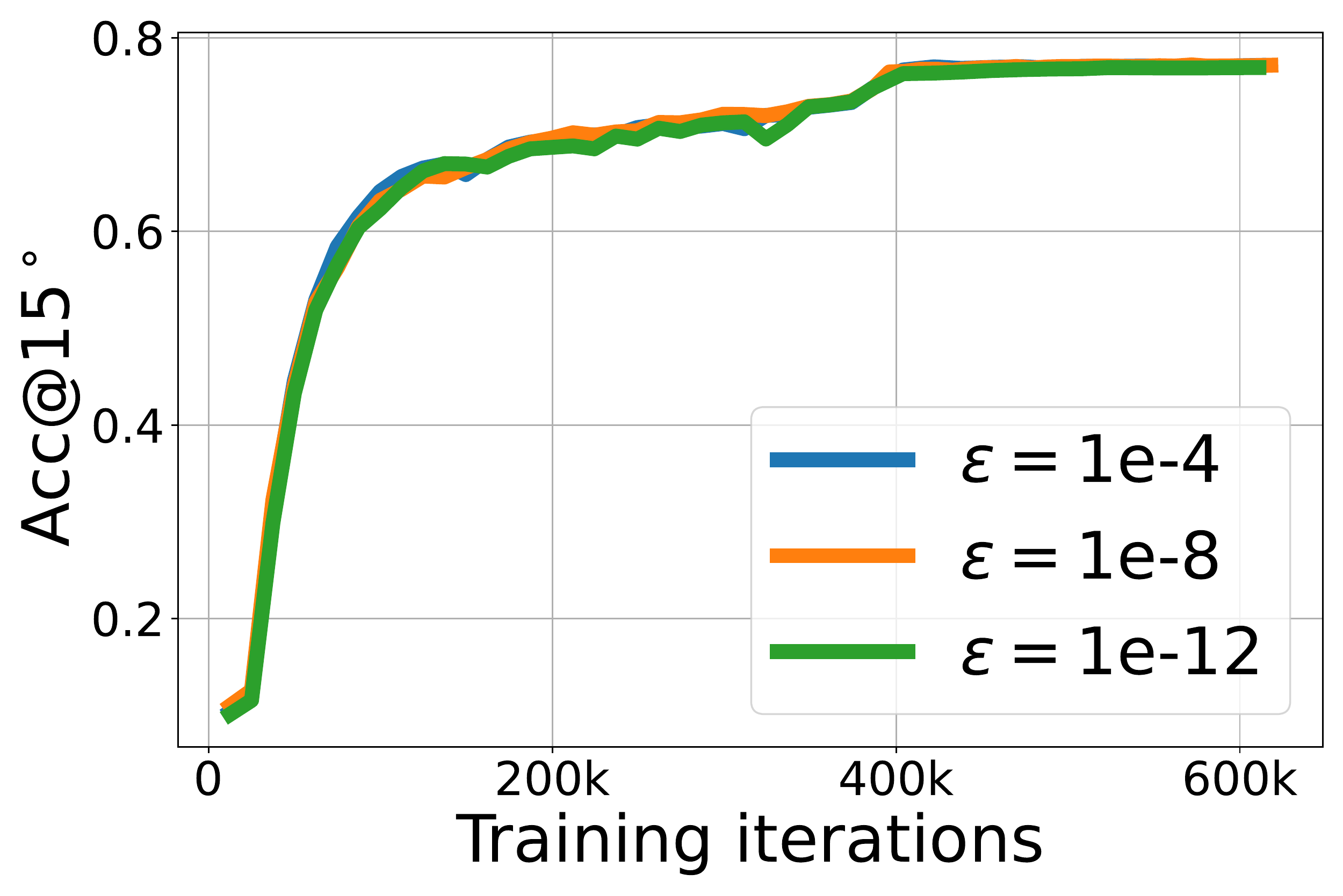}
    &\includegraphics[width=0.3\linewidth]{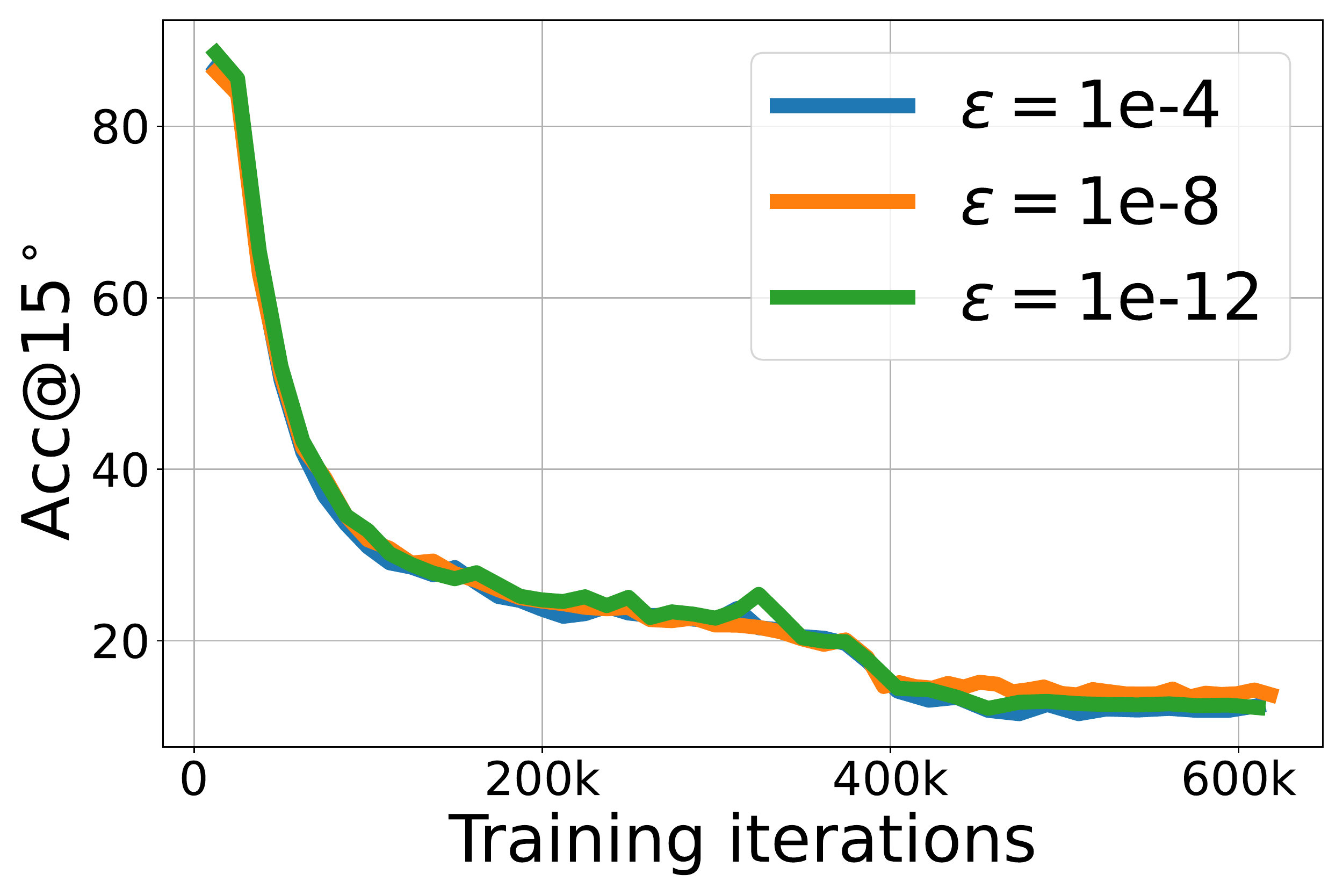}
    \\
    \small{Acc@15$^\circ$} &\small{Acc@30$^\circ$} &\small{Med. ($^\circ$)} 
    \end{tabular}
    \vspace{-2mm}
    \caption{\small \textbf{The visualization of the evaluation metrics along with the training process with different clipping parameter $\epsilon$.} The clipping parameter does not affect the process. The experiments are on ModelNet10-SO3 dataset averaged on all categories.}
    \vspace{-2mm}
	\label{fig:ablation_clip}
\end{figure}

}
\revision{
\subsection{Experiments on Wahba's Problem}
We also benchmark our method on Wahba's problem to showcase the optimization process.

\begin{problem} \textbf{Wahba's Problem}

Given two sets of n points $\{ p_1, p_2, ..., p_n\}$ and $\{ p_1^*, p_2^*, ..., p_n^*\}$, where $n\ge 2$, find the rotation matrix $\mathbf{R}$ which brings the first set into the best least squares coincidence with the second. That is, find $\mathbf{R}$ which minimizes
\begin{equation}
    \sum_{i=1}^{n} \| p_i^* - \mathbf{R}p_i\|^2
\end{equation}
\end{problem}

In accordance with the methodology outlined in \cite{zhou2019continuity, peretroukhin2020smooth}, our neural network is trained on a dataset comprising 2290 airplane point clouds sourced from the ShapeNet dataset \cite{shapenet2015}. Subsequently, testing is conducted on 400 held-out point clouds. Throughout both training and evaluation phases, each point cloud undergoes a random rotation. The training process involves 10,000 iterations utilizing the Adam optimizer, with a batch size of 100 and a learning rate set to 1e-5.

The progress of training and test errors across iterations is visually depicted in Figure \ref{fig:wahba}. Notably, the rotation Laplace distribution achieves superior or comparable performances with alternative rotation
representations or distributions.
}
\retwo{
We provide more experiments on Wahba's problem with instance-level setting in the supplementary.
}

\jiangran{



\section{Analysis of the robustness with Outliers}
\label{sec:outlier}
\subsection{Analysis on Gradient w.r.t. Outliers}

\begin{figure*}[t]
    \centering
    \includegraphics[width=0.4\linewidth]{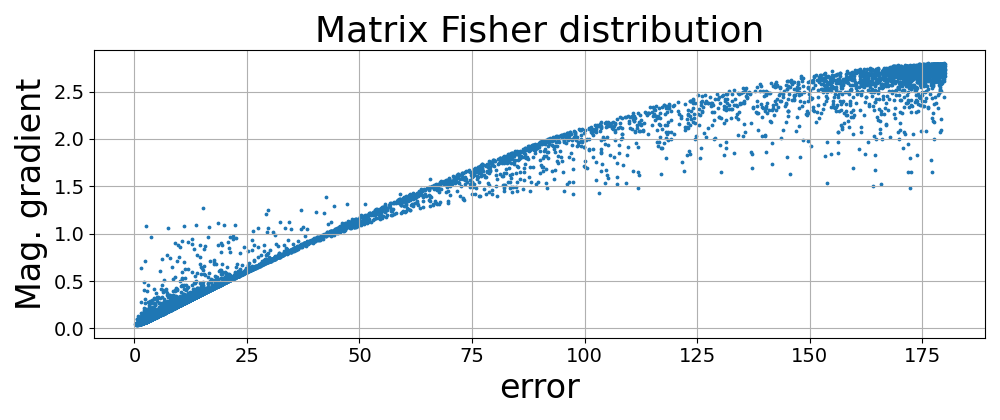}
    \hspace{6mm}
    \includegraphics[width=0.4\linewidth]{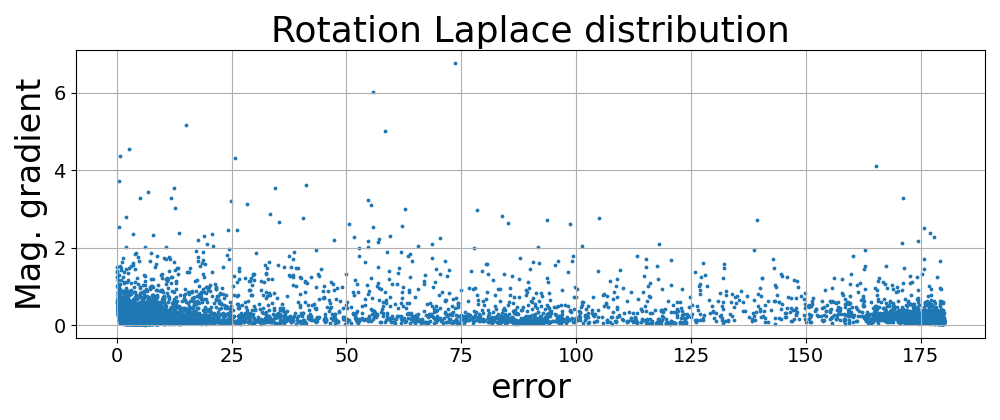}
    \vspace{-4mm}
    \caption{\ree{Visualization of the gradient magnituide $\|\partial\mathcal{L}/\partial\text{(distribution param.)}\|$ w.r.t. the prediction errors  on ModelNet10-SO3 dataset after convergence.}}
	\label{fig:grad_scatter}
\end{figure*}

In the task of rotation regression, predictions with really large errors (e.g., 180$^\circ$ error) are fairly observed due to rotational ambiguity or lack of discriminative visual features. Properly handling these outliers during training is one of the keys to success in probabilistic modeling of rotations. 

In Figure \ref{fig:grad_scatter}, for matrix Fisher distribution and rotation Laplace distribution, we visualize the gradient magnitudes $\|\partial\mathcal{L}/\partial\text{(distribution param.)}\|$ w.r.t. the prediction errors on ModelNet10-SO3 dataset after convergence, where each point is a data point in the test set.
As shown in the figure, for matrix Fisher distribution, predictions with larger errors clearly yield larger gradient magnitudes, and those with near 180$^\circ$ errors (the outliers) have the biggest impact. 
Given that outliers may be inevitable and hard to be fixed, they may severely disturb the training process and the sensitivity to outliers can result in a poor fit \cite{murphy2012machine,nair2022maximum}. In contrast, for our rotation Laplace distribution, the gradient magnitudes are not affected by the prediction errors much, leading to a stable learning process.

Consistent results can also be seen in Figure \ref{fig:teaser} of the main paper, where the red dots illustrate the \textit{sum} of the gradient magnitude over the population within an interval of prediction errors. We argue that, at convergence, the gradient should focus more on the large population with low errors rather than fixing the unavoidable large errors.

\subsection{Experiments on ModelNet10-SO3 Dataset with Outlier  Injections}
To further demonstrate the robustness of our distribution, we manually inject outliers to the perfectly labeled synthetic dataset and compare rotation Laplace distribution with matrix Fisher distribution. 
Specifically, we randomly choose 1\%, 5\%, 10\% and 30\% images from the training set of ModelNet10-SO3 dataset respectively, and apply a random rotation in $\SO$ to the given ground truth. Thus, the chosen images become outliers in the dataset due to the perturbed annotations. We fix the processed dataset for different methods.

The results on the perturbed dataset are shown in Table \ref{tab:outlier} and Figure \ref{fig:outlier}, where our method consistently outperforms matrix Fisher distribution under different levels of perturbations. More importantly, as shown in Figure \ref{fig:outlier}, our method clearly better tolerates the outliers, resulting in less performance degradation and remains a reasonable performance even under intense perturbations. For example, Acc@30$^\circ$ of matrix Fisher distribution greatly drops from 0.751 to 0.467 with 30\% outliers, while that of our method merely goes down from 0.770 to 0.700, which shows the superior robustness of our method.

\begin{table*}[t]
  \centering
  \fontsize{7.8}{9.5}\selectfont
  \caption{\ree{Comparisons on the perturbed ModelNet10-SO3 dataset where random outliers are injected.}}
    \begin{tabular}{clcccccc}
    \toprule
        Outlier Inject  &  Method     & Acc@3$^\circ$$\uparrow$ & Acc@5$^\circ$$\uparrow$ & Acc@10$^\circ$$\uparrow$ & Acc@15$^\circ$$\uparrow$ & Acc@30$^\circ$$\uparrow$ & Med.($^\circ$)$\downarrow$ \\
    \midrule
    
    \multirow{2}[0]{*}{0\%} 
    & Mohlin \textit{et al.}\cite{mohlin2020probabilistic}    &  0.164& 0.389 &  0.615 &  0.684&  0.751&   17.9 \\
    & rotation Laplace                  &  \textbf{0.446}  & \textbf{0.613}  &  \textbf{0.714}  &  \textbf{0.741}  &  \textbf{0.770} & \textbf{12.2}   \\
    \midrule
    \multirow{2}[0]{*}{1\%} 
    & Mohlin \textit{et al.}\cite{mohlin2020probabilistic}    &  0.141    &	0.336&	0.589&	0.664&	0.740&	20.5\\
    & rotation Laplace                  &  \textbf{0.429}&	\textbf{0.601}&	\textbf{0.711}&	\textbf{0.739}&   \textbf{0.769}	 &\textbf{12.4} 
    \\
    \midrule
    \multirow{2}[0]{*}{5\%} 
    & Mohlin \textit{et al.}\cite{mohlin2020probabilistic}    &  0.0818&	0.229&	0.501&	0.605&	0.711&	24.8 \\
    & rotation Laplace                  &   \textbf{0.368}&	\textbf{0.561}&	\textbf{0.693}&	\textbf{0.727}&	\textbf{0.762}&	\textbf{12.6} \\
    \midrule
    \multirow{2}[0]{*}{10\%} 
    & Mohlin \textit{et al.}\cite{mohlin2020probabilistic}    &   0.0493&	0.151&	0.403&	0.536&	0.677&	26.8\\
    & rotation Laplace                  &   \textbf{0.329}&	\textbf{0.523}&	\textbf{0.668}&	\textbf{0.706}&	\textbf{0.747}&	\textbf{16.1}\\
    \midrule
    \multirow{2}[0]{*}{30\%} 
    & Mohlin \textit{et al.}\cite{mohlin2020probabilistic}    &   0.0063&	0.0255&	0.126&	0.243&	0.467&	45.3\\
    & rotation Laplace                  &   \textbf{0.151}&	\textbf{0.345}&	\textbf{0.565}&	\textbf{0.634}&	\textbf{0.700}&	\textbf{22.7}\\
    \bottomrule
    \end{tabular}%
  \label{tab:outlier}%
\end{table*}%

\begin{figure}[t]
    \centering
    \includegraphics[width=0.32\linewidth]{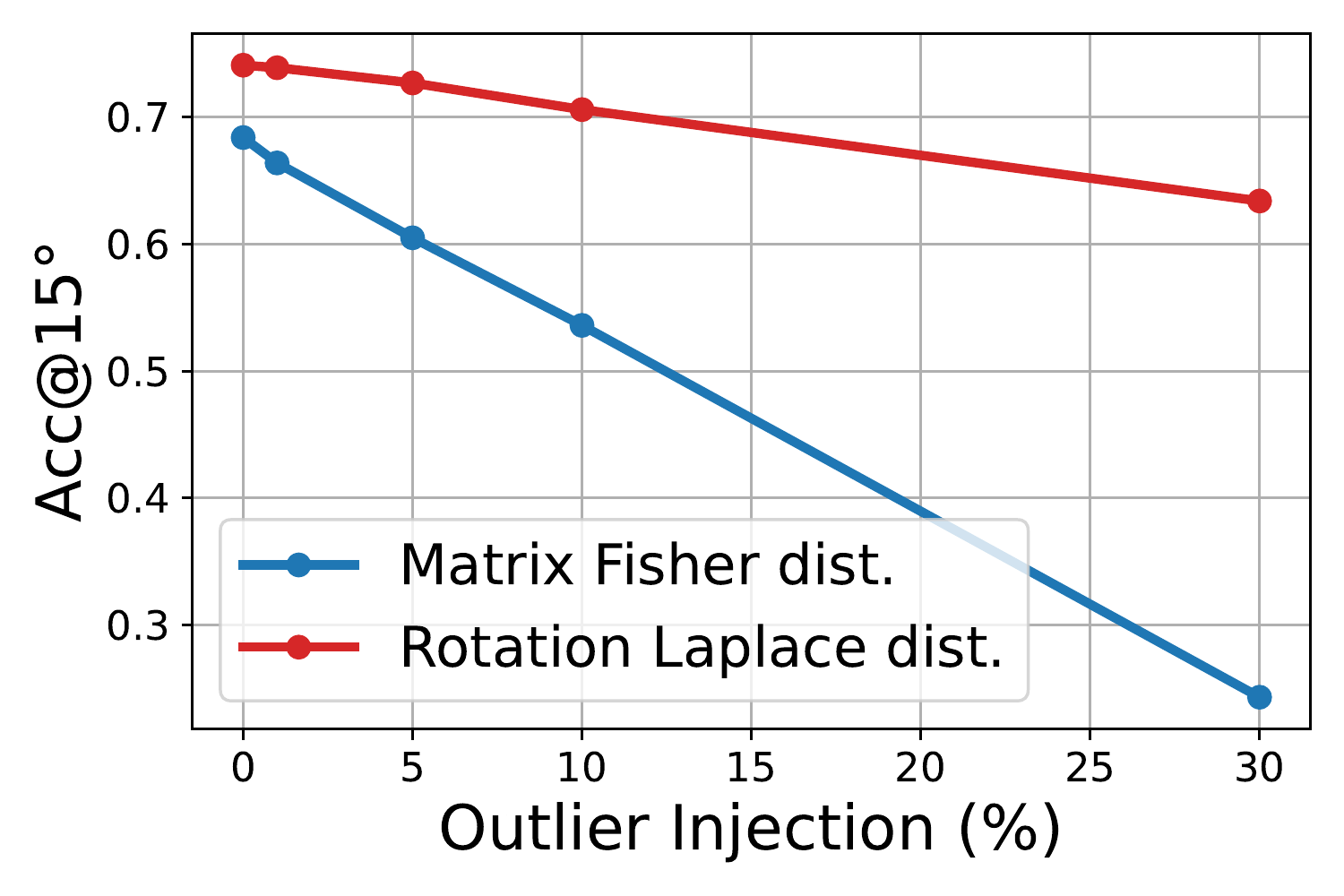}
    \includegraphics[width=0.32\linewidth]{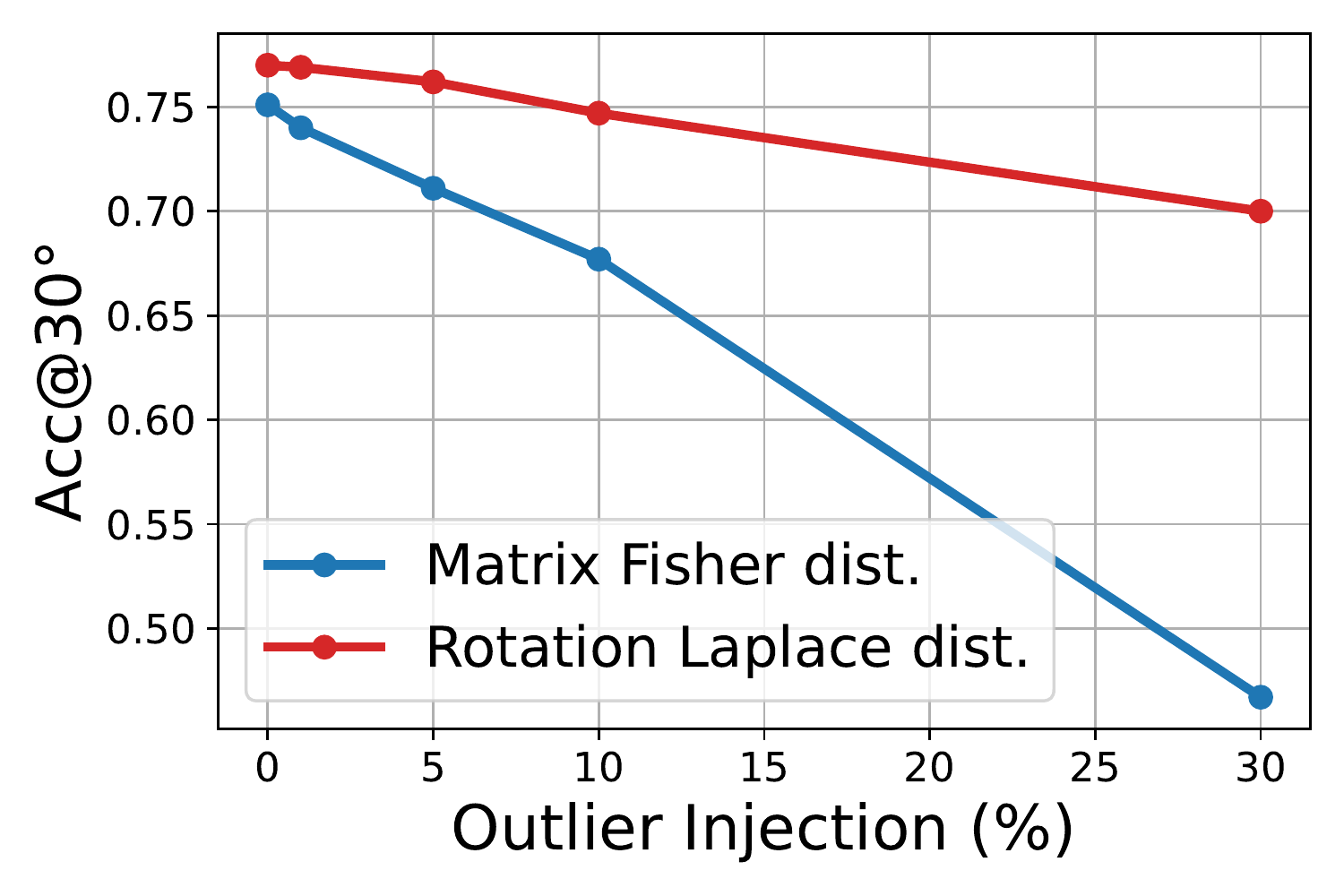}
    \includegraphics[width=0.32\linewidth]{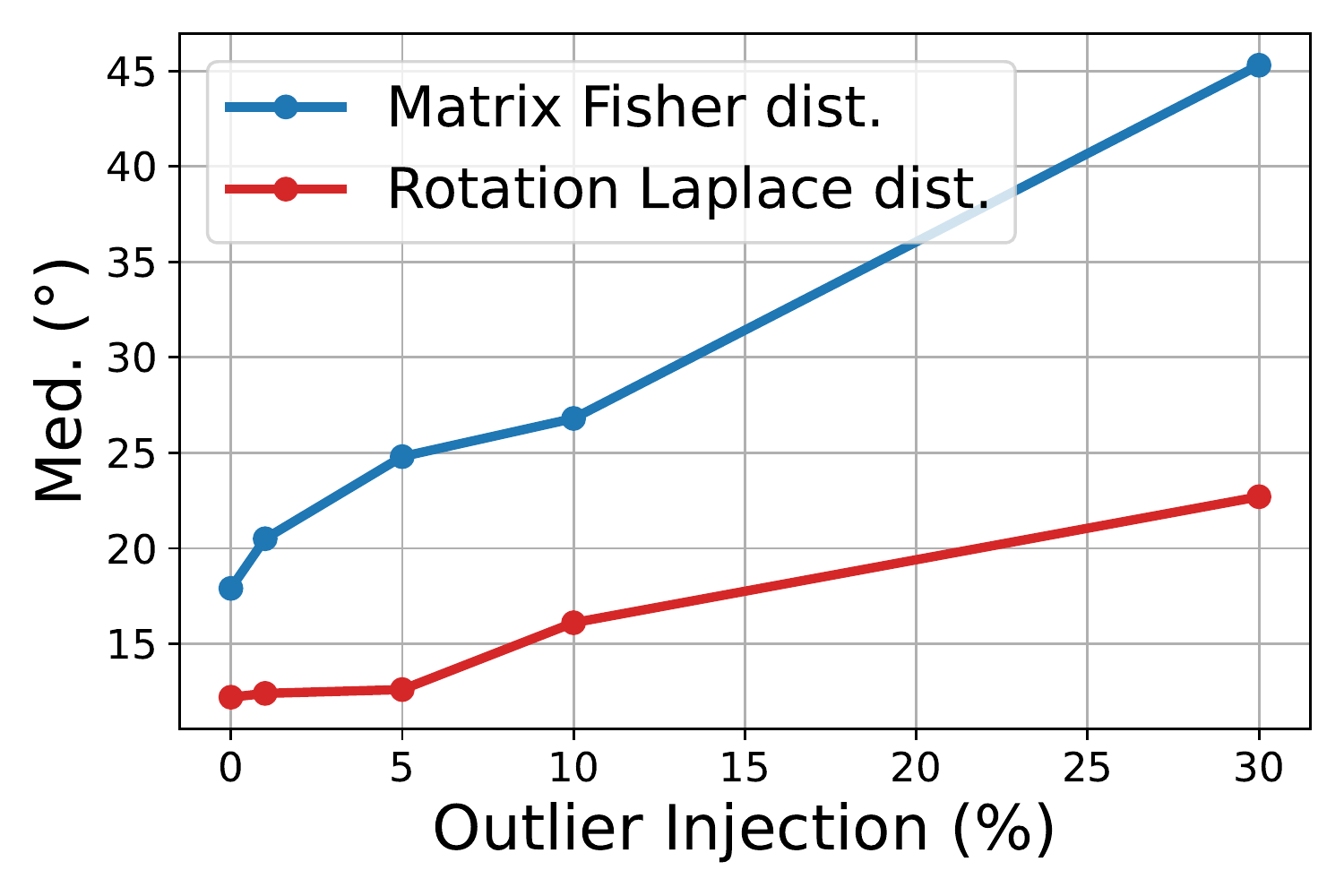}
    \caption{\ree{Comparisons on the perturbed ModelNet10-SO3 dataset where random outliers are injected.
    The horizontal axis is the percentage of perturbed images, and the vertical axis represents the corresponding metric.}}
    \vspace{-2mm}
	\label{fig:outlier}
\end{figure}

}
\jiangran{\section{Analysis of the robustness with Noise}
\label{sec:noise}
Mathematically, our rotation Laplace distribution benefits from the heavy-tail nature of the distribution, which introduces a small gradient when the difference between prediction and label is large, resulting in its robustness to outliers. 
However, one possible trade-off is that when the prediction is close to the label, the gradient is large, which could result in the wrong gradient dominating the training process in scenarios with slight noise. To investigate this issue, we experiment with perturbed data that includes small noise injections in \ref{subsec:noise_injection}. We further evaluate and apply the robustness to the semi-supervised rotation regression task where pseudo labels are noisy in \ref{subsec:ssl}.

\begin{figure}[t]
    \centering
    \includegraphics[width=0.32\linewidth]{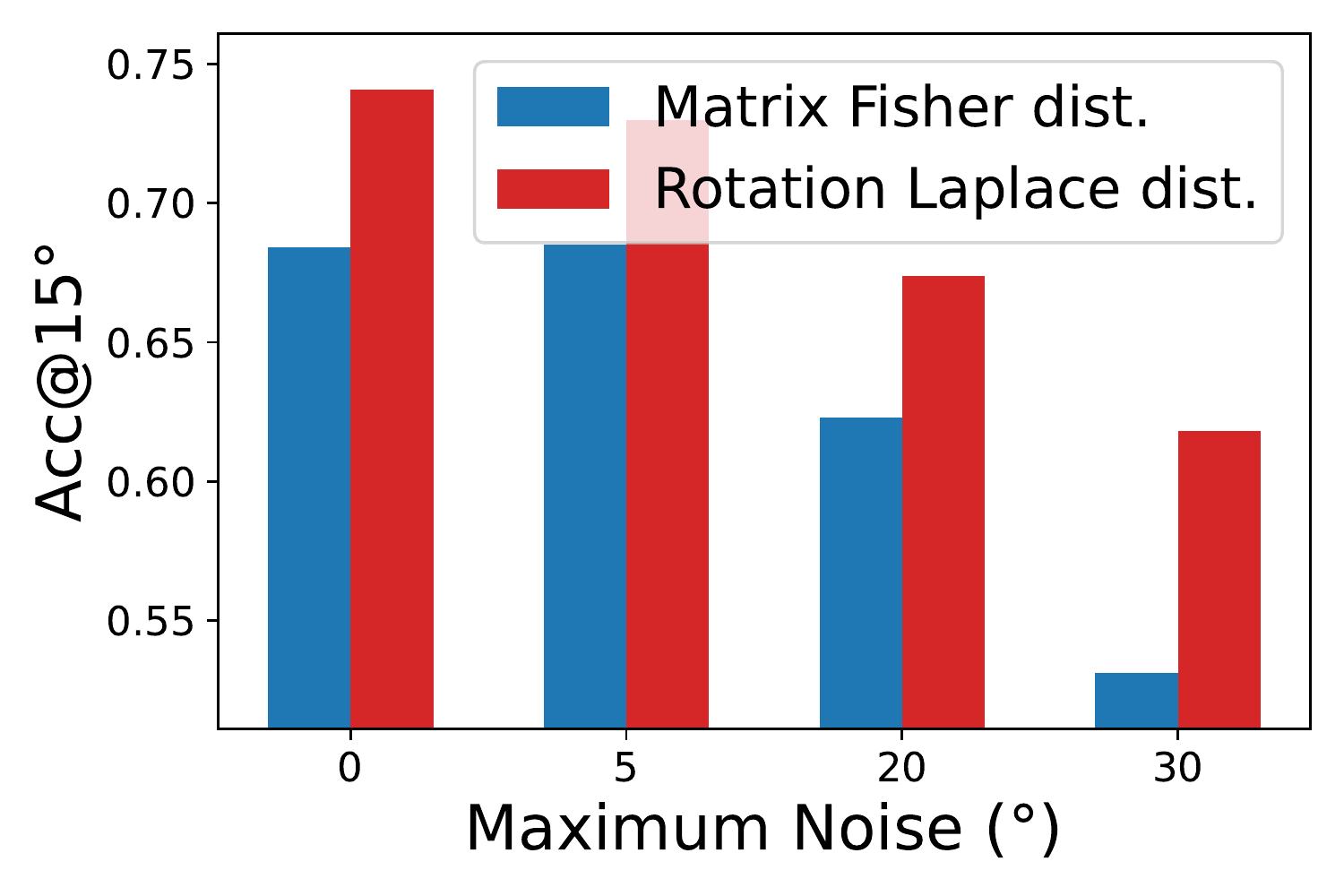}
    \includegraphics[width=0.32\linewidth]{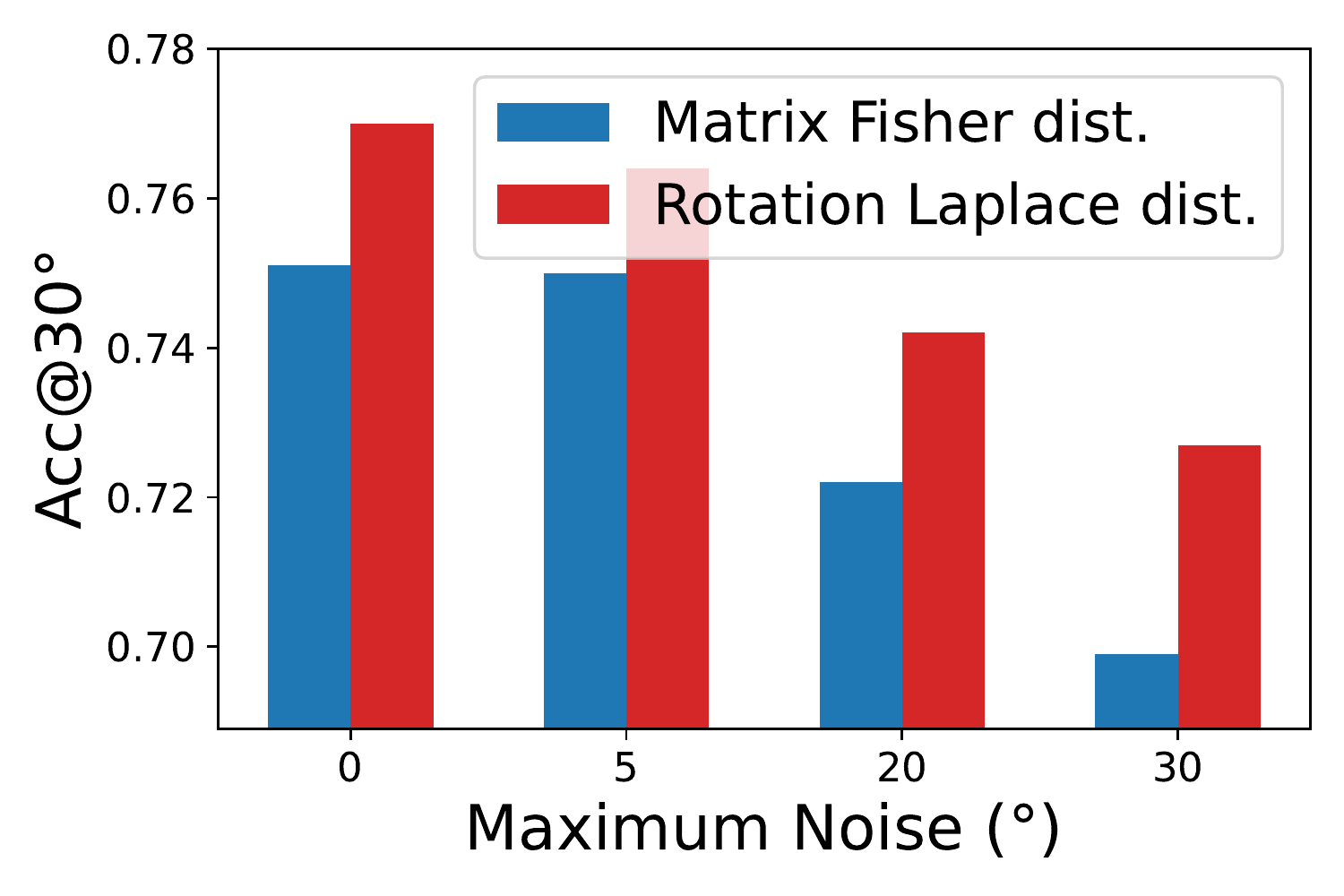}
    \includegraphics[width=0.32\linewidth]{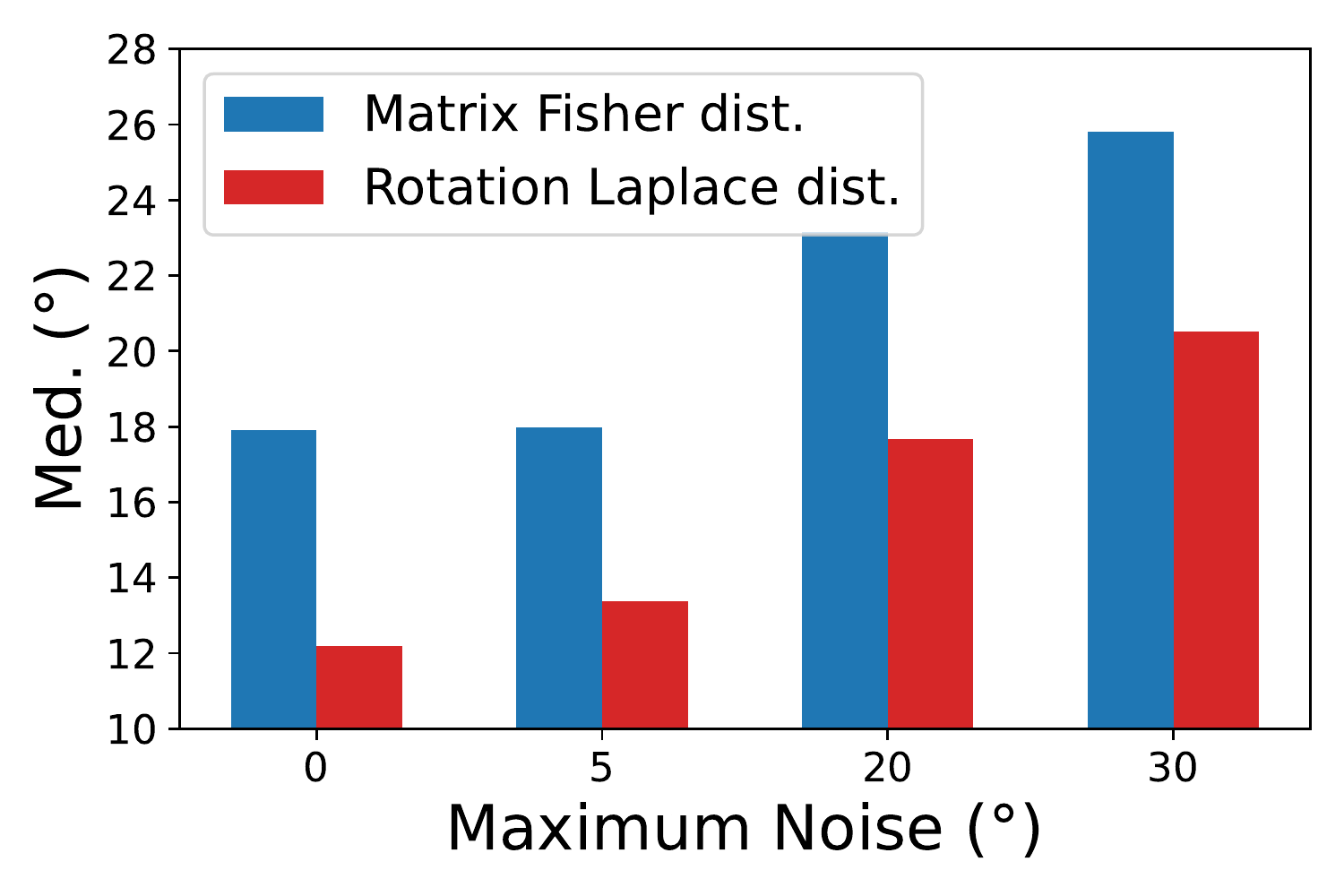}
    \caption{\ree{Comparisons on the perturbed ModelNet10-SO3 dataset where random noise are injected and fixed.
    The horizontal axis is the maximum noise, and the vertical axis represents the corresponding metric.}}
    \vspace{-2mm}
	\label{fig:const_noise}
\end{figure}

\subsection{Experiments on ModelNet10-SO3 Dataset with Noise Injections}
\label{subsec:noise_injection}

Noises on dataset labels, such as human labeling errors, can cause overfitting on imperfect training sets and lead to poor performance during testing.
To assess the robustness of our rotation Laplace distribution to such noise, we perturb ModelNet10-SO3 with an injection of random noise and compare the performance of our rotation Laplace with baseline under this noisy condition. Specifically, we generate random rotation in the form of the axis-angle representation where the axis direction and the angle's magnitude are sampled from uniform distributions on the unit sphere $\mathrm{S}^2$ and $ \left[ 0, u \right]$, respectively. The upper bound of the magnitude $u$ was set to 5$^\circ$, 20$^\circ$, and 30$^\circ$ in the following experiments. We add this random rotation to all training data labels as noise and fixed the processed dataset for different methods.
\begin{table}[htbp]
  \centering
  \caption{Comparisons on the perturbed ModelNet10-SO3 dataset where random noise are injected and fixed.}
    \begin{tabular}{c|c|cccc}
    \toprule
    \multirow{2}[4]{*}{Metric} & \multirow{2}[4]{*}{Method} & \multicolumn{4}{c}{Maximum Noise ($^\circ$)} \\
\cmidrule{3-6}          &       & 0 & 5  & 20 & 30  \\
    \midrule
    \multirow{2}[2]{*}{Acc@15$^\circ$$\uparrow$} & Mohlin \textit{et al.}\cite{mohlin2020probabilistic} & 0.684 & 0.683 & 0.623 & 0.531 \\
          & rotation Laplace & \textbf{0.741} & \textbf{0.730}  & \textbf{0.674} & \textbf{0.618} \\
    \midrule
    \multirow{2}[2]{*}{Acc@30$^\circ$$\uparrow$} & Mohlin \textit{et al.}\cite{mohlin2020probabilistic} & 0.751 & 0.750  & 0.722 & 0.699 \\
          & rotation Laplace & \textbf{0.770}  & \textbf{0.764} & \textbf{0.742} & \textbf{0.727} \\
    \midrule
    \multirow{2}[2]{*}{Med.($^\circ$)$\downarrow$} & Mohlin \textit{et al.}\cite{mohlin2020probabilistic} & 17.9  & 18.0 & 23.2 & 25.8 \\
          & rotation Laplace & \textbf{12.2}  & \textbf{13.4} & \textbf{17.7} & \textbf{20.5} \\
    \bottomrule
    \end{tabular}%
  \label{tab:const_noise}%
\end{table}%

\begin{table}[htbp]
  \centering
  \caption{Comparisons on the perturbed ModelNet10-SO3 dataset where dynamically changing noise are injected.}
    \begin{tabular}{c|c|cccc}
    \toprule
    \multirow{2}[4]{*}{Metric} & \multirow{2}[4]{*}{Method} & \multicolumn{4}{c}{Maximum Noise ($^\circ$)} \\
\cmidrule{3-6}          &       & \multicolumn{1}{c}{0}  & \multicolumn{1}{c}{5}  & \multicolumn{1}{c}{20} & \multicolumn{1}{c}{30} \\
    \midrule
    \multirow{2}[2]{*}{{Acc@15$^\circ$$\uparrow$}} & Mohlin \textit{et al.}\cite{mohlin2020probabilistic} & 0.684 & 0.682 & 0.678 & 0.667 \\
          & rotation Laplace & \textbf{0.741} & \textbf{0.737} & \textbf{0.717} & \textbf{0.708} \\
    \midrule
    \multirow{2}[2]{*}{{Acc@30$^\circ$$\uparrow$}} & Mohlin \textit{et al.}\cite{mohlin2020probabilistic} & 0.751 & 0.753 & 0.745 & 0.745 \\
          & rotation Laplace & \textbf{0.770} & \textbf{0.770} & \textbf{0.761} & \textbf{0.764} \\
    \midrule
    \multirow{2}[2]{*}{Med.($^\circ$)$\downarrow$ } & Mohlin \textit{et al.}\cite{mohlin2020probabilistic} & 17.9  & 16.8 & 17.7 & 17.1 \\
          & rotation Laplace & \textbf{12.2} & \textbf{12.0} & \textbf{13.6} & \textbf{13.1} \\
\bottomrule
\end{tabular}%
  \label{tab:dynamic_noise}%
\end{table}%

The results on the perturbed dataset are presented in Table \ref{tab:const_noise} and Figure \ref{fig:const_noise}, where our method consistently outperforms matrix Fisher distribution under different levels of perturbations. For instance, at a maximum human labeling noise of 5$^\circ$ and 30$^\circ$, Acc@15$^\circ$ of our methods outperforms that of the matrix Fisher distribution by 4.7\% and 8.7\%. Our method maintains reasonable performance even under relatively severe perturbations, indicating its superior practical value in real-world scenarios involving human labeling errors.

\begin{figure}[t]
    \centering
    \includegraphics[width=0.32\linewidth]{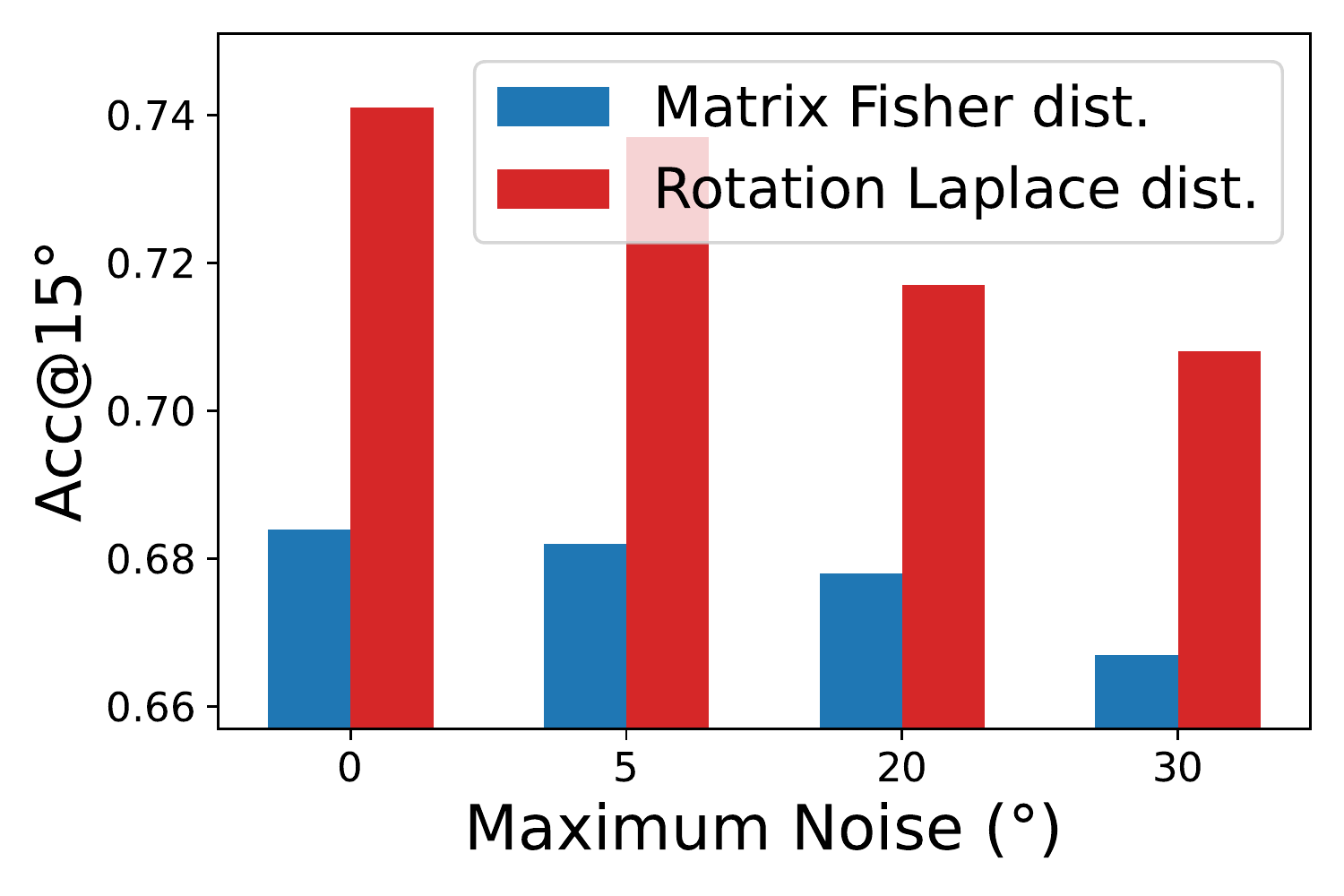}
    \includegraphics[width=0.32\linewidth]{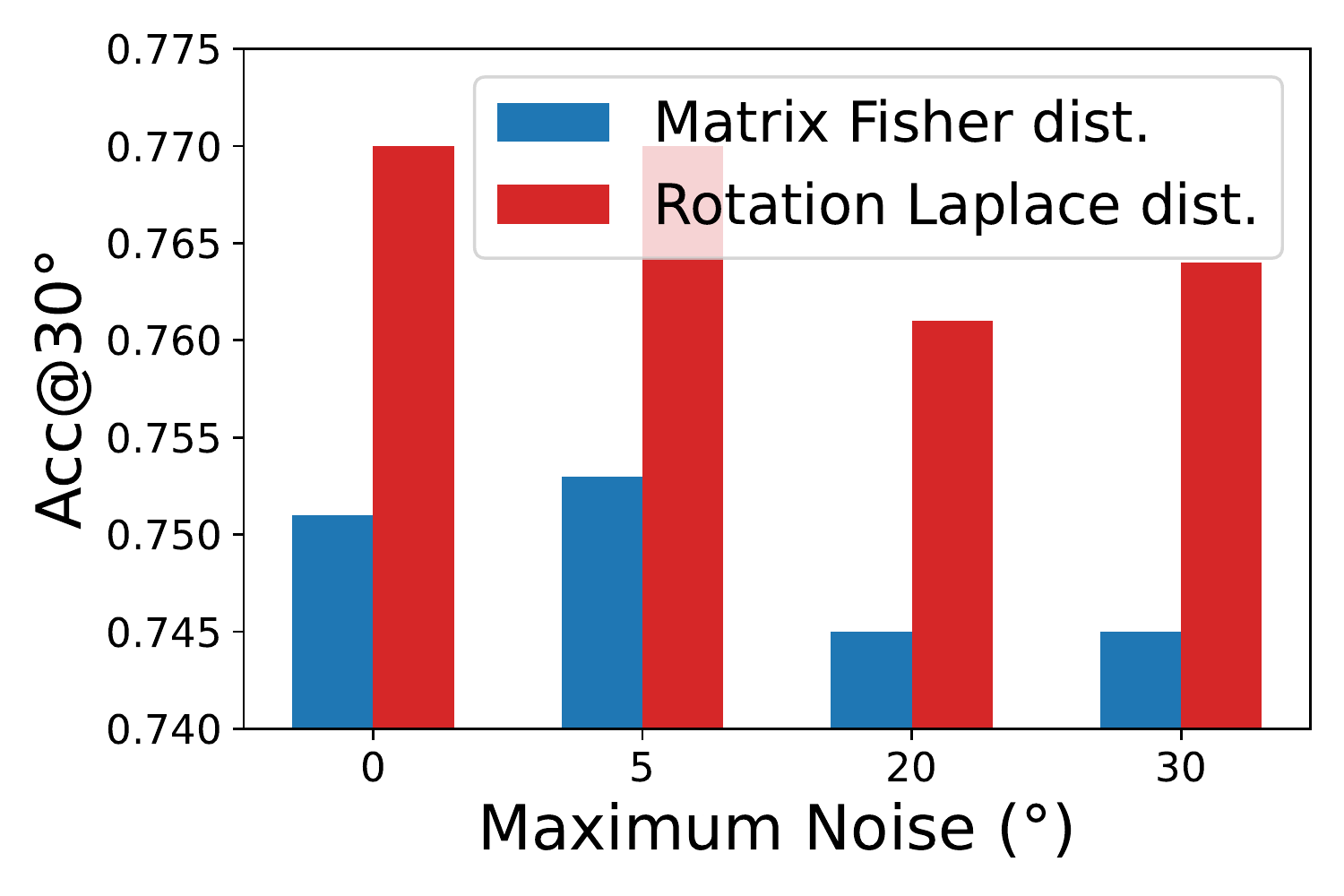}
    \includegraphics[width=0.32\linewidth]{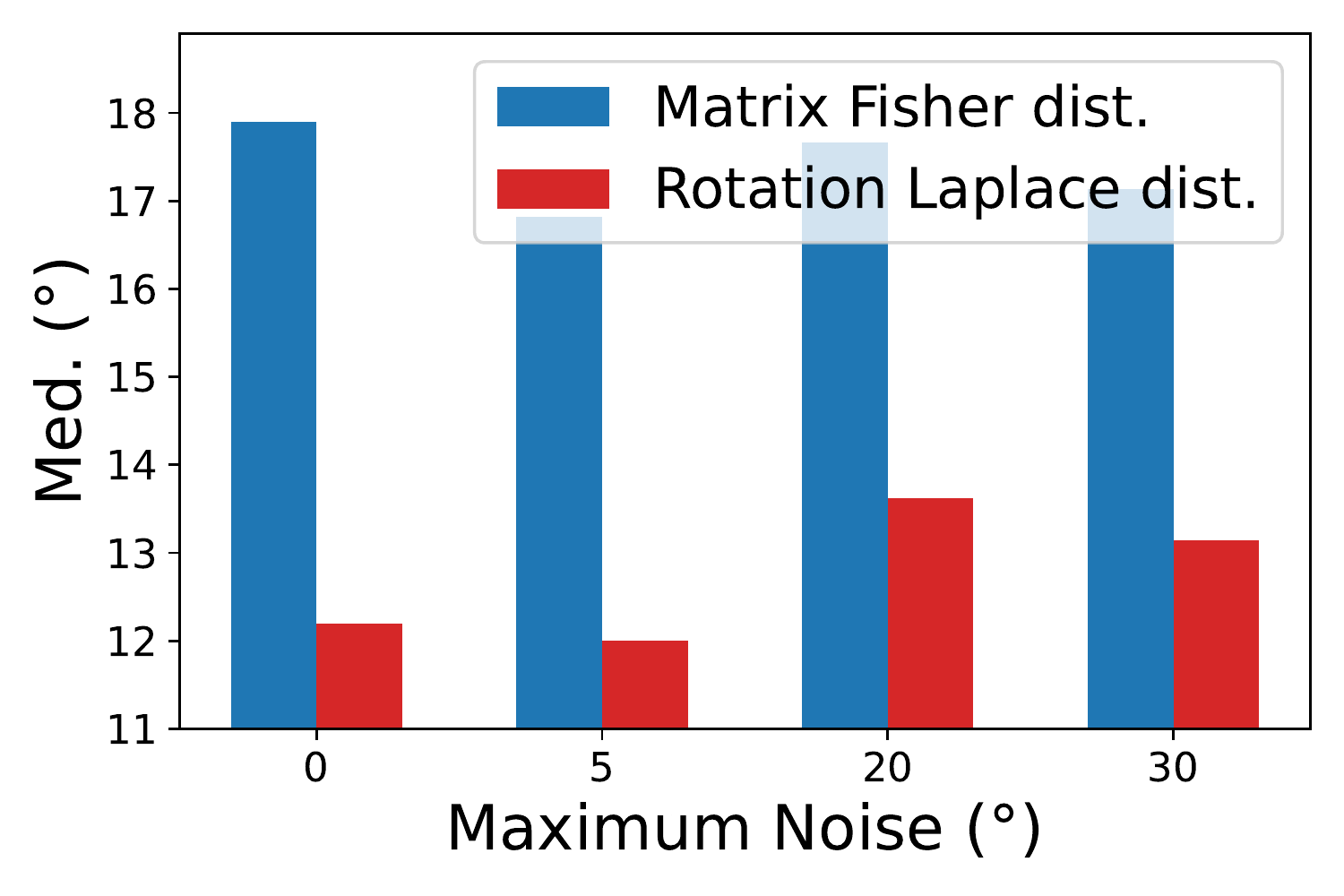}
    \caption{\ree{Comparisons on the perturbed ModelNet10-SO3 dataset where dynamically changing noise are injected.
    The horizontal axis is the maximum noise, and the vertical axis represents the corresponding metric.}}
    \vspace{-2mm}
	\label{fig:dynamic_noise}
\end{figure}

We also perform experiments by introducing dynamically changing noise into the ground truth labels, simulating the use of pseudo-labeling in semi-supervised learning as discussed in Section \ref{subsec:ssl}. In this experiment, we regenerate random rotation noise each epoch using the same method as before, with an upper bound of 5$^\circ$, 20$^\circ$, and 30$^\circ$. This approach allows us to evaluate the robustness of the rotation Laplace and matrix Fisher distributions under noisy while evolving labels.

Table \ref{tab:dynamic_noise} and Figure \ref{fig:dynamic_noise} present the quantitative comparisons of our rotation Laplace distribution and the matrix Fisher distribution on the perturbed ModelNet10-SO3 dataset, where dynamically changing noise is injected. Notably, our method consistently outperforms the matrix Fisher distribution in scenarios with both small and relatively large levels of dynamically changing noise while maintaining a reasonable margin. And both methods demonstrate better performance than training on a fixed perturbed dataset, for this setting alleviates overfitting on noisy training sets. 

Although our method achieves good performance under various levels of dynamic noise, we observed that in some cases, training results with large noise may slightly outperform those with small noise with some measure. For example, the median error is lower when the maximum noise is 5$^\circ$ compared to when there is no noise, and the Acc@15$^\circ$ is higher when the maximum noise is 30$^\circ$ compared to 20$^\circ$. This can be attributed to the fact that dynamic noise makes the labels smoother to some extent, which achieves data augmentation through jittering. However, this situation is limited to cases where the noise gap is small. When the noise gap is large, training performance is still better with low noise levels, as evidenced by the comparison between no noise and 30$^\circ$ noise.

\subsection{Application on Semi-supervised Rotation Regression}
\label{subsec:ssl}
One of the major obstacles to improving
rotation regression is expensive rotation annotations.
Though many large-scale image datasets have been curated with sufficient semantic annotations, obtaining a large-scale
real dataset with rotation annotations can be extremely laborious,
expensive and error-prone\cite{xiang2014beyond}. To reduce the amount of supervision, Yin \textit{et al.} proposed FisherMatch\cite{yin2022fishermatch} to learn regressor of matrix Fisher distribution from minor labeled data and a large amount of unlabeled data, namely semi-supervised learning. The core method is based on pseudo-labeling and further leverages the \textit{entropy} of matrix Fisher distribution as uncertainty to filter low-quality pseudo-labels. According to our prior analysis, our method still demonstrates satisfactory performance in scenarios with dynamically changing noise. Therefore, it is more appropriate to utilize it for training scenes that involve the use of pseudo-labeling with minimal errors. In light of this, we incorporate rotation Laplace distribution into the framework of FisherMatch to evaluate its applicability in semi-supervised rotation regression tasks.   

\subsubsection{Uncertainty Quantification Measured by Distribution  Entropy}
\label{sec:entropy}

\begin{figure}[t]
    \centering
    \includegraphics[width=0.98\linewidth]{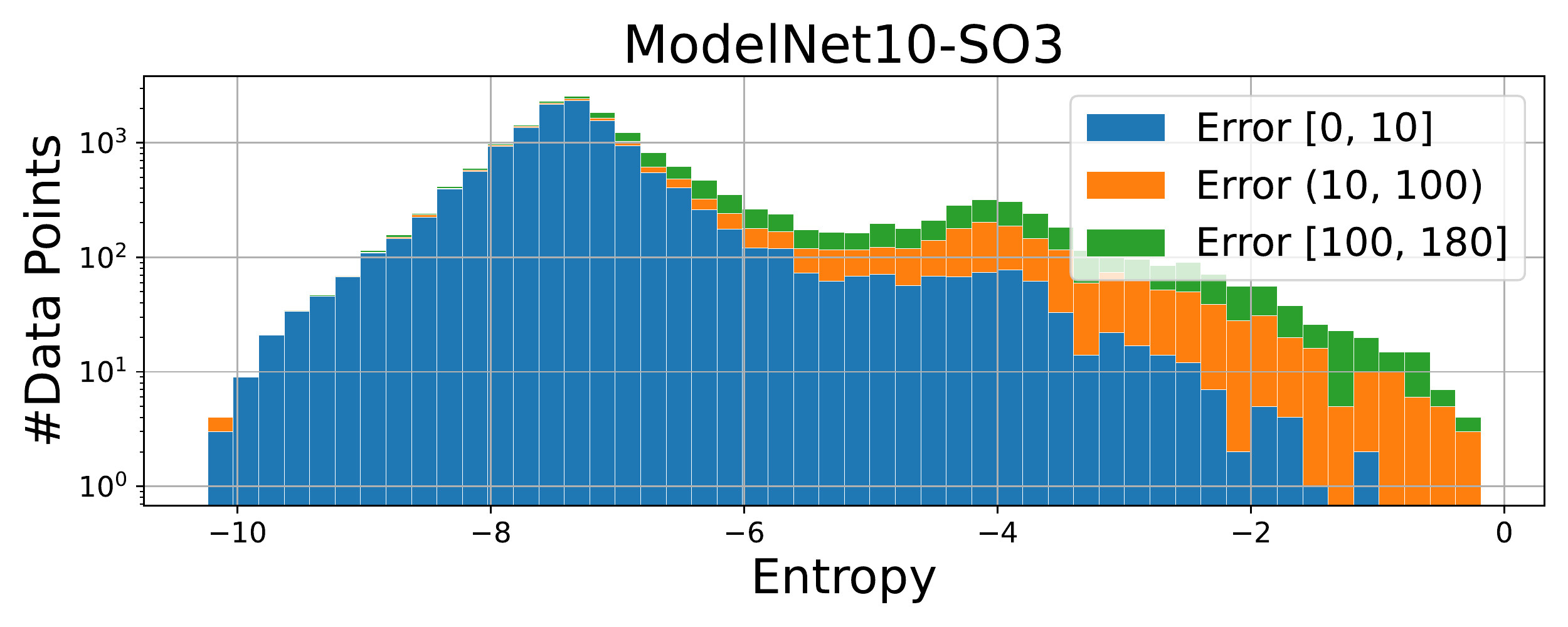}
    
    \vspace{-3mm}
    \caption{Visualization of the indication ability of the distribution entropy w.r.t. the performance. The horizontal axis is the distribution entropy and the vertical axis is the number of data points (in log scale), color coded by the errors (in degrees). The experiments are done on the test set of ModelNet10-SO3 dataset.}
    \vspace{-2mm}
	\label{fig:uncertain}
\end{figure}

Probabilistic modeling of rotation naturally models the uncertainty information of rotation regression. Yin \textit{et al.}\cite{yin2022fishermatch} proposes to use the \textit{entropy} of the distribution as an uncertainty measure. We adopt it as the uncertainty indicator of rotation Laplace distribution and plot the relationship between the error of the prediction and the corresponding distribution entropy on the testset of ModelNet10-SO3 in Figure \ref{fig:uncertain}.
As shown in the figure, 
predictions with lower entropies (i.e., lower uncertainty) clearly achieve higher accuracy than predictions with large entropies, demonstrating the ability of uncertainty estimation of our rotation Laplace distribution.
We compute the entropy via discretization, 
where $\SO$ space is quantized into a finite set of equivolumetric girds $\mathcal{G}=\{\mathbf{R}|\mathbf{R}\in \SO\}$, and

\begin{equation*}
\scriptsize
    H\left(p\right)=-\int_{\SO} p \log p \mathrm{d} \mathbf{R}
    \approx 
    -\sum_{\mathbf{R}_i \in \mathcal{G}} p_i \log p_i \Delta \mathbf{R}_i
\end{equation*}
We use about 0.3M grids to discretize $\SO$ space.

\setlength{\tabcolsep}{4pt}
\begin{table}[t]
  \footnotesize
  \caption{ 
  Comparing our method with the baselines under different ratios of labeled data for semi-supervised rotation regression.
  }
  \vspace{-3mm}

    \begin{tabular}{clcccccc}
    \toprule
    \multirow{2}[3]{*}{Category} & \multicolumn{1}{c}{\multirow{2}[3]{*}{Method}} & \multicolumn{2}{c}{5\%} & \multicolumn{2}{c}{10\%}  \\
\cmidrule{3-6}          &       & \multicolumn{1}{c}{Mean($^{\circ}$)} & \multicolumn{1}{c}{Med.($^{\circ}$)} & \multicolumn{1}{c}{Mean($^{\circ}$)} & \multicolumn{1}{c}{Med.($^{\circ}$)} \\
    \midrule
    \multirow{3}{*}{Sofa} 
& SSL-9D-Consist.        &   36.86   &   8.65    &    25.94  &    6.81     \\
& SSL-FisherMatch       &   31.09   &   7.55    &    21.16  &    5.21    \\
& SSL-rot.Laplace        &   \textbf{26.12}   &   \textbf{6.07}    &    \textbf{19.32}  &    \textbf{4.25}    \\
    \midrule
    \multirow{3}{*}{Chair} 
& SSL-9D-Consist.        &    31.20  &    11.29  &    23.59  &    8.10   \\
& SSL-FisherMatch       &    26.45  &    9.65   &    20.18  &    7.63   \\
& SSL-rot.Laplace        &    \textbf{23.66}  &    \textbf{8.51}   &    \textbf{18.35}  &    \textbf{6.75}   \\
    \bottomrule
    \end{tabular}
  \label{tab:ssl}
\end{table}

\subsubsection{Revisit FisherMatch}
FisherMatch leverages a teacher-student mutual learning framework composed of a learnable student model and an exponential-moving-average (EMA) teacher model. A training batch for this framework contains a mixture of $\left\{\boldsymbol{x}_{i}^{l}\right\}_{i=1}^{B_{l}}$ labeled samples and $\left\{\boldsymbol{x}_{i}^{u}\right\}_{i=1}^{B_{u}}$ unlabeled samples. 

On labeled data, the student network is trained by the ground-truth labels with the supervised loss; while on unlabeled data, the student model takes the pseudo labels from the EMA teacher. An entropy-based filtering technique is leveraged to filter out noisy teacher predictions. The overall loss term is as follows:
\begin{equation}
L_{ssl}=L_{l}\left(\boldsymbol{x}^{l},\boldsymbol{y}^{l}\right)+
\lambda_{u} L_{u}\left(\boldsymbol{x}^{u}\right)
\end{equation}
where $\lambda_u$ is the unsupervised loss weight.

For \textit{pseudo label filtering}, a fixed entropy threshold $\tau$ is set, and the prediction will be reserved as a pseudo label only if its entropy is lower than the threshold. We denote rotation regressor as $\Phi$ that takes a single RGB image $\boldsymbol{x}$ as input and $\mathcal{MF}$ as matrix Fisher distribution. Specifically, for unlabeled data $\boldsymbol{x}^u$, 
assume $p_t=\mathcal{MF}(\mathbf{A}^u_t)$ is the teacher output with $\mathbf{A}^u_t=\mathbf{\Phi}_t(\boldsymbol{x}^u)$ and
$p_s=\mathcal{MF}(\mathbf{A}^u_s)$ is the student output with $\mathbf{A}^u_s=\mathbf{\Phi}_s(\boldsymbol{x}^u)$ , the loss on unlabeled data is therefore:
\begin{equation}
\label{eq:unsuper}
L_u\left(\boldsymbol{x}^{u}\right) =  \mathbbm{1}  \left(H(p_t) \leq \tau\right) L_u\left(p_t, p_s\right)
\end{equation}
We recommend the readers to \cite{yin2022fishermatch} for more details.

\subsubsection{Incorporating Rotation Laplace into FisherMatch}
Note that FisherMatch is a framework independent of probabilistic distribution of rotation, we can simply replace the matrix Fisher distribution with our rotation Laplace distribution. According to Equ \ref{eq:unsuper}, we can derive unsupervised loss as follows:
\begin{equation}
\label{eq:unsuper_laplace}
L_u\left(\boldsymbol{x}^{u}\right) =  \mathbbm{1}  \left(H(\mathcal{RL}(\mathbf{A}^u_t)) \leq \tau\right) L\left(\mathcal{RL}(\mathbf{A}^u_t), \mathcal{RL}(\mathbf{A}^u_t)\right)
\end{equation}
Specifically, we use NLL loss for supervised loss similar to Sec.\ref{sec:exp} and CE (cross entropy) loss for unsupervised loss in Eq. \ref{eq:unsuper_laplace}.

We conduct experiment on ModelNet10-SO3 dataset, following the same semi-supervised learning setting in FisherMatch, which uses MobileNet-V2\cite{howard2017mobilenets} architecture and Adam optimizer. Both the baselines come from Yin \textit{et al.}\cite{yin2022fishermatch} due to they are the unique work to tackle semi-supervised rotation {regression} on SO(3).
\textit{SSL-L1-Consistency} refers to adopting the teacher-student mutual learning framework which applies L1 loss as the consistency supervision between the student and teacher predictions without filtering.

The experimental results in Table \ref{tab:ssl} indicate that our method consistently outperforms both baselines and achieves a significant improvement over FisherMatch in both 5\% and 10\% labeled data scenarios. Specifically, for the sofa class at a 5\% label ratio, our method reduces the mean error by 4.97$^\circ$ and the median error by 1.48$^\circ$. These results demonstrate the superior noise-robustness of our rotation Laplace distribution for semi-supervised learning.}
\revision{\section{Application on Object Pose estimation}
\label{sec:6dpose}
Monocular object pose estimation emerges as a significant downstream task for our proposed rotation representation. This task involves estimating the 3D translation and 3D rotation of an object from a single RGB image, with broad applications in real-world tasks such as robotic manipulation, augmented reality, and autonomous driving. Leveraging the inherent nature of probabilistic modeling, rotation Laplace distribution can seamlessly integrate into existing regression-based methods. We conduct experiments using GDR-Net \cite{wang2021gdr} as the baseline to showcase the advantages of incorporating our method. 

\begin{table*}[t]
\centering
\footnotesize
\caption{Application of rotation Laplace distribution for monocular object pose estimation on YCB-video and LINEMOD datasets. We select GDR-Net\cite{wang2021gdr} as baseline and report the results evaluated w.r.t. ADD(-S), AUC of ADD-S, ADD(-S), rotation accuracy and mean error. }
\begin{tabular}{ccccccccc}
\toprule
Dataset & Method & ADD(-S) & \begin{tabular}[c]{@{}c@{}}AUC of\\    ADD-S\end{tabular} & \begin{tabular}[c]{@{}c@{}}AUC of \\ ADD(-S)\end{tabular} & Acc@2$^\circ$ & Acc@5$^\circ$ & Acc@10$^\circ$ & Mean ($^\circ$) \\ \midrule
\multirow{2}{*}{YCB-video} & GDR-Net\cite{wang2021gdr} & 49.1 & 89.1 & 80.2 & {9.1} & {36.3} & {67.9} & 25.2 \\
 & GDR-Net-rot.Laplace & \textbf{57.0} & \textbf{90.5} & \textbf{82.4} & \textbf{9.7} & \textbf{44.7} & \textbf{78.1} & \textbf{20.4} \\ \midrule
\multirow{2}{*}{LINEMOD} & GDR-Net\cite{wang2021gdr} & 93.7 & \textbf{-} & \textbf{-} & 63.2 & 97.0 & \textbf{99.7} & 1.97 \\
 & GDR-Net-rot.Laplace & \textbf{94.2} & \textbf{-} & \textbf{-} & \textbf{66.2} & \textbf{97.5} & \textbf{99.7} & \textbf{1.87} \\ 
 \bottomrule
\end{tabular}

\label{tab:6dpose}
\end{table*}

\textbf{Datasets.} LINEMOD dataset \cite{hinterstoisser2013model} consists of 13 sequences, each containing approximately 1.2k images with ground-truth poses for a single object, including clutter and mild occlusion. Following the approaches in \cite{brachmann2016uncertainty, wang2021gdr}, we use approximately 15\% of the RGB images for training and 85\% for testing. During training, we also incorporate 1k rendered RGB images for each object, following the methodology outlined in \cite{li2019cdpn}.
YCB-video \cite{xiang2017posecnn} is a highly challenging dataset that exhibits strong occlusion, clutter, and features several symmetric objects. It comprises over 110k real images captured with 21 objects. In line with the approach presented in \cite{wang2021gdr}, we also utilize publicly available synthetic data generated using physically-based rendering (pbr) \cite{hodavn2020bop} for training.

\textbf{Metrics.} 
In evaluating 6DoF object pose, we employ widely used metrics, i.e. ADD(-S). Originating from \cite{hinterstoisser2013model}, the ADD metric gauges if the average deviation of transformed model points falls below 10\% of the object’s diameter (0.1d). For objects with symmetry, we turn to the ADD-S metric, measuring error as the average distance to the nearest model point.
When evaluating on YCB-video, our evaluation extends to computing the AUC (area under the curve) of ADD(-S), adjusting the distance threshold with a capped maximum of 10 cm \cite{xiang2017posecnn}. Furthermore, to better evaluate the performance of rotation estimation, we include rotation accuracy and the mean rotation error in our analysis. Rotation accuracy is measured as the percentage of predictions whose rotation error is below the specified threshold. The computation of rotation error concerns the smallest error among all conceivable ground-truth poses for symmetric objects, aligning with the methodology in GDR-Net \cite{wang2021gdr}.

\textbf{Implementation Details.} GDR-Net predicts 2D-3D correspondence as an intermediate representation and subsequently regresses rotation and translation. The original rotation representation employed in GDR-Net is 6D \cite{zhou2019continuity}. To ensure a fair comparison, we straightforwardly replace the 6D rotation representation with our rotation Laplace distribution and then utilize the mode of the distribution as the rotation prediction.
We maintain consistency with the baseline training strategy, employing the Ranger optimizer with a batch size of 24 and a base learning rate of 1e-4. The learning rate is annealed at 72\% of the training phase using a cosine scheduler.
A single model is trained for all objects within each dataset.


\textbf{Results.} As shown in Table 3, our method enhances performance on both datasets. Particularly noteworthy is the significant improvement observed on YCB-video, where the rotation Laplace distribution leads to an increase of approximately 8 points for ADD(-S) and 5 degrees for the mean rotation error.
}
\jiangran{\section{Multimodal prediction with rotation Laplace mixture model}

For the purpose of capturing multimodal rotation space, especially with symmetric objects, we extend rotation Laplace distribution into rotation Laplace mixture model.

\subsection{Rotation Laplace Mixture Model}
Rotation Laplace mixture model can be built upon a set of single-modal rotation Laplace distributions by assigning a weight factor for each component and combining them linearly.
In this way, each unimodal component may capture one plausible solution to the corresponding task and the mixture model takes multi-modality into account and covers a broader solution space.

\begin{definition} \emph{Rotation Laplace mixture model.}
The random variable $\mathbf{R}\in \SO$ follows rotation Laplace mixture model with parameter $\mathcal{A}=\{\mathbf{A}_i\}_{i=1}^{M}$ and weights $\mathcal{W}=\{w_i\}_{i=1}^{M}$, if its probability density function is defined as
\begin{equation}
\label{eq:rl_mixture}
\small
\begin{aligned}
    p(\mathbf{R}; \mathcal{A}, \mathcal{W}) 
    &= \sum_{i=1}^M w_i \cdot p(\mathbf{R};\mathbf{A}_i)\\
    &=\sum_{i=1}^M  
    \frac{w_i}{F(\mathbf{A}_i)}
    \frac{\exp\left(-\sqrt{\operatorname{tr}\left(\mathbf{S}_i - \mathbf{A}_i^T \mathbf{R}\right)}\right)}
    {\sqrt{\operatorname{tr}\left(\mathbf{S}_i -\mathbf{A}_i^T \mathbf{R}_i\right)}}
\end{aligned}
\end{equation}
where $M$ is the number of components, $w_i\in [0, 1]$ is the scalar weight of each component, and
\begin{equation*}
    \sum_{i=1}^M w_i = 1
\end{equation*}
$\mathbf{A}_i\in \mathbb{R}^{3\times 3}$ is an unconstrained matrix, and $\mathbf{S}_i$ is the diagonal matrix composed of the proper singular values of matrix $\mathbf{A}_i$, i.e., $\mathbf{A}_i=\mathbf{U}_i\mathbf{S}_i\mathbf{V}_i^T$. 
We also denote rotation Laplace mixture model as $\mathbf{R} \sim \mathcal{RL}_{mix}(\mathcal{A}, \mathcal{W})$.
\end{definition}

\subsection{Loss Function}
\textbf{Mixture rotation Laplace loss.}
Similar to Eq. \ref{eq:nll_loss}, we define mixture rotation Laplace loss as the negative log-likelihood (NLL) of the distribution of the mixture model, as follows:

\begin{equation}
\label{eq:mixture_nll_loss}
    \mathcal{L}_\text{nll}(\boldsymbol{x}, \mathbf{R}_{\boldsymbol{x}}) = -\log p\left( \mathbf{R}_{\boldsymbol{x}};\mathcal{A}_{\boldsymbol{x}},  \mathcal{W}_{\boldsymbol{x}}\right)
\end{equation}
where $\mathcal{A}_{\boldsymbol{x}}$ and $\mathcal{W}_{\boldsymbol{x}}$ are predicted by the network.
Mixture rotation Laplace loss can be viewed as the linear combination of the NLL loss of each component.

In theory, mixture rotation Laplace loss is sufficient for capturing the multi-modality of the solution space. However, as pointed out by previous works \cite{bishop1994mixture,makansi2019overcoming}, directly optimizing with the NLL loss of the mixture model can lead to numerical instabilities and mode collapse. Thus a proper technique used for encouraging diverging predictions is crucial.

\textbf{Winner-Take-ALL loss.}
Inspired by \cite{deng2022deep}, we incorporate a ``Winner-Take-All'' (WTA) strategy in our training process, which has been shown to be effective with multimodal scenarios \cite{rupprecht2017learning,manhardt2019explaining}.
In concrete, in WTA strategy, we only update the branch with highest probability density function of ground truth and leave the other branches unchanged. This provably leads to the Voronoi tesselation of the output space \cite{deng2022deep}.

Denote $\mathcal{RL}_i$ as the i-th component of the mixture model. We first select the ``winner'' branch by checking the pdf of ground truth $\mathbf{R}_{\boldsymbol{x}}$:
\begin{equation}
    i^*=\underset{i}{\arg \max } \mathcal{RL}_i\left(\mathbf{R}_{\boldsymbol{x}} ; \mathbf{A}_i\right)
\end{equation}
And then optimizing this branch with NLL loss (Eq. \ref{eq:nll_loss}):
\begin{equation}
    \mathcal{L}_{WTA} = \mathcal{L}_{i^*} (\boldsymbol{x}, \mathbf{R}_{\boldsymbol{x}} )
\end{equation}
In addition, Rupprecht et al. \cite{rupprecht2017learning} find that a \textit{relaxed} version of WTA which allows a small portion of gradient for unselected branches will help with training by avoiding ``dead'' branches which never get updated due to the bad initialization. Thus we use the relaxed WTA strategy:
\begin{equation}
\begin{aligned}
    \mathcal{L}_{RWTA} = \sum_{i=1}^{M}\pi_i\mathcal{L}_{i} (\boldsymbol{x}, \mathbf{R}_{\boldsymbol{x}} )\\
    \pi_i= \begin{cases}1-\epsilon & i=i^* \\ \frac{\epsilon}{M-1} & \text { others }\end{cases}
\end{aligned}
\end{equation}
We use $\epsilon=0.05$ in experiments. RWTA loss guides our mixture model to better cover different modes of solution space.

The overall loss of rotation Laplace mixture model is defined as follows
\begin{equation}
    \mathcal{L}_\text{mix} = \mathcal{L}_\text{nll} + \lambda \mathcal{L}_\text{RWTA}
\end{equation}
We set $\lambda=1$ in experiments.

\subsection{Experiments with Rotation Laplace Mixture Model}
We validate rotation Laplace mixture model on the task of object rotation estimation, similar to Sec. \ref{sec:exp}. To better evaluate the multi-modality property of the distributions, we follow IPDF \cite{murphy2021implicit} to use \textit{top-k} metrics where the best results from k candidates are reported. 

\subsubsection{Top-k Metrics}
Since only a single ground truth is available, for symmetric objects, the precision metrics can be misleading because they may penalize correct predictions that do not have a corresponding annotation. We follow IPDF \cite{murphy2021implicit} to use top-k metrics: $N$ pose candidates are predicted by the mixture model, and the best results of the $k$ ($k\le N$) predictions are reported. We set $N=4$ and $k=2, 4$ as IPDF. Note that we only train one model with $N$ branches without the burden to re-train it for every specified $k$.

\subsubsection{Quantitative Results}
\setlength{\tabcolsep}{7pt}
\begin{table}[t]
  \centering
  \caption{Numerical comparisons on multimodal probabilistic distributions with top-k metrics on ModelNet10-SO3 dataset averaged on all categories.}
    \begin{tabular}{clccc}
    \toprule
          &       & Acc@15$^\circ$$\uparrow$ & Acc@30$^\circ$$\uparrow$ & Med$\downarrow$ \\
    \midrule
    \multirow{4}[2]{*}{\makecell{Single \\modal}} & 
        Deng \textit{et al.}\cite{deng2022deep}             &   0.562 &  0.694 &   32.6 \\
        &  Mohlin \textit{et al.}\cite{mohlin2020probabilistic}  &   0.693 &  0.757 &   17.1  \\
        &  Murphy \textit{et al.}\cite{murphy2021implicit}       &  0.719 &  0.735 &   21.5 \\
        & rotation Laplace & \textbf{0.742}   & \textbf{0.772} & \textbf{12.7} \\
    \midrule
    \multirow{4}[2]{*}{Top-2} & 
          Deng \textit{et al.}\cite{deng2022deep} & 0.863 & 0.897 & 3.8 \\
          & Mohlin \textit{et al.}\cite{mohlin2020probabilistic} & 0.864 & 0.903 & 3.8 \\
          & Murphy \textit{et al.}\cite{murphy2021implicit}  & 0.868 & 0.888 & 4.9 \\
          & rotation Laplace & \textbf{0.900}   & \textbf{0.918} & \textbf{2.3} \\
    \midrule
    \multirow{4}[2]{*}{Top-4} & 
          Deng \textit{et al.}\cite{deng2022deep} & 0.875 & 0.915 & 3.7 \\
          & Mohlin \textit{et al.}\cite{mohlin2020probabilistic} & 0.882 & 0.926 & 3.7 \\
          & Murphy \textit{et al.}\cite{murphy2021implicit}  & 0.904 & 0.926 & 4.8 \\
          & rotation Laplace  & \textbf{0.919} & \textbf{0.940}  & \textbf{2.2} \\
    \bottomrule
    \end{tabular}%
  \label{tab:mixture}%
\end{table}%

We compare our rotation Laplace distribution with other probabilistic baselines on all categories of ModelNet10-SO3 dataset, and the results are reported in Tab \ref{tab:mixture}.
Shown in the table, the performance of all methods increases dramatically when we allow for $k$ (even when $k=2$) candidates, which demonstrate the advantage of taking multi-modality into account. 

Besides superior performance with single modal output, our rotation Laplace distribution also performs the best among all the baselines under all metrics in both top-2 and top-4 settings, which validates the feasibility and effectiveness of rotation Laplace mixture model.

\subsubsection{Qualitative Results}

\begin{figure}[t]
    \centering
    \begin{tabular}{ccccc}
    \hspace{-4mm}
    \includegraphics[height=1.4cm]{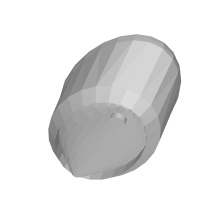} 
    &\hspace{-4mm}
    \includegraphics[height=0.77cm]{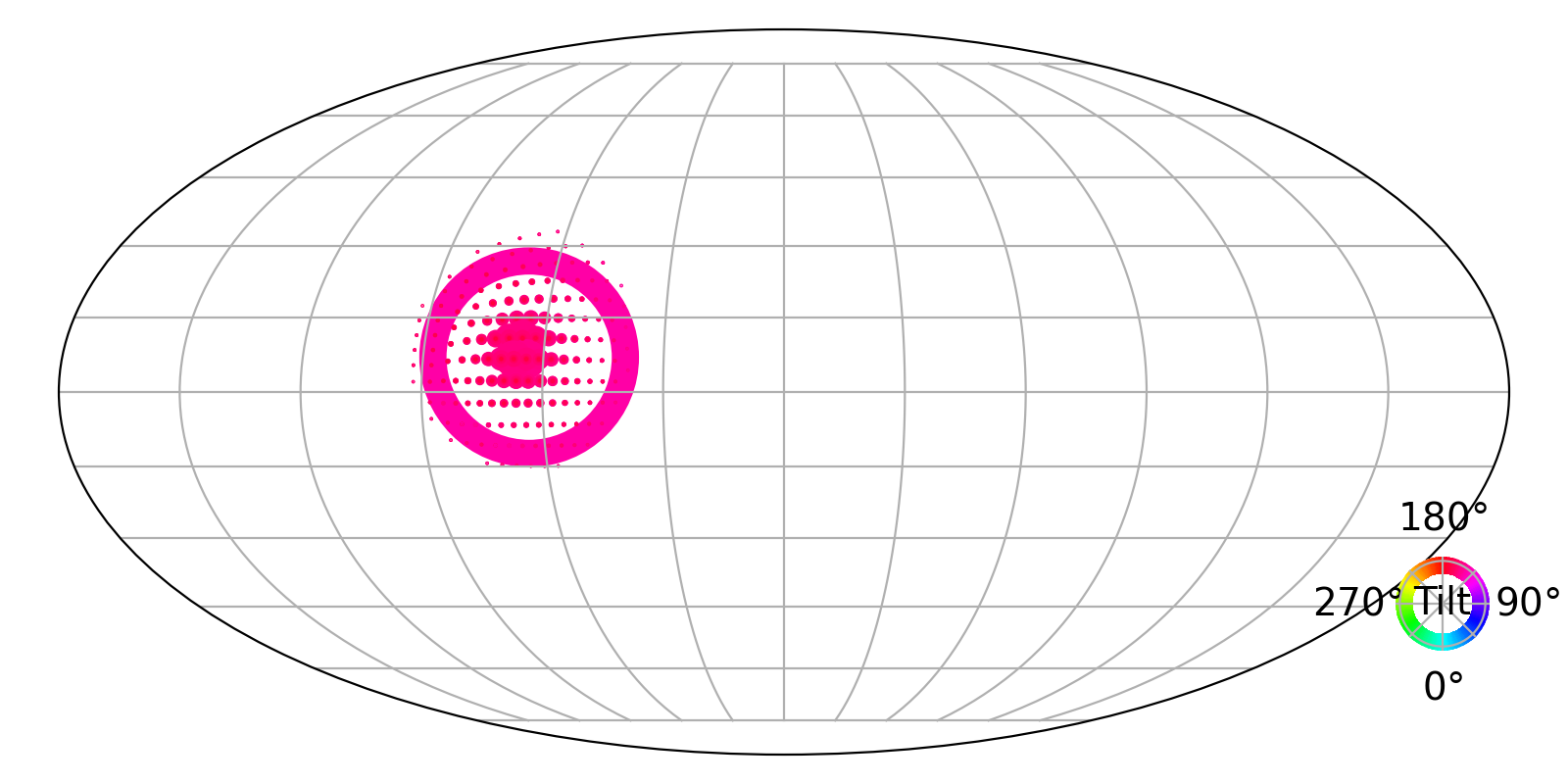} 
    &\hspace{-4mm}
    \includegraphics[height=0.77cm]{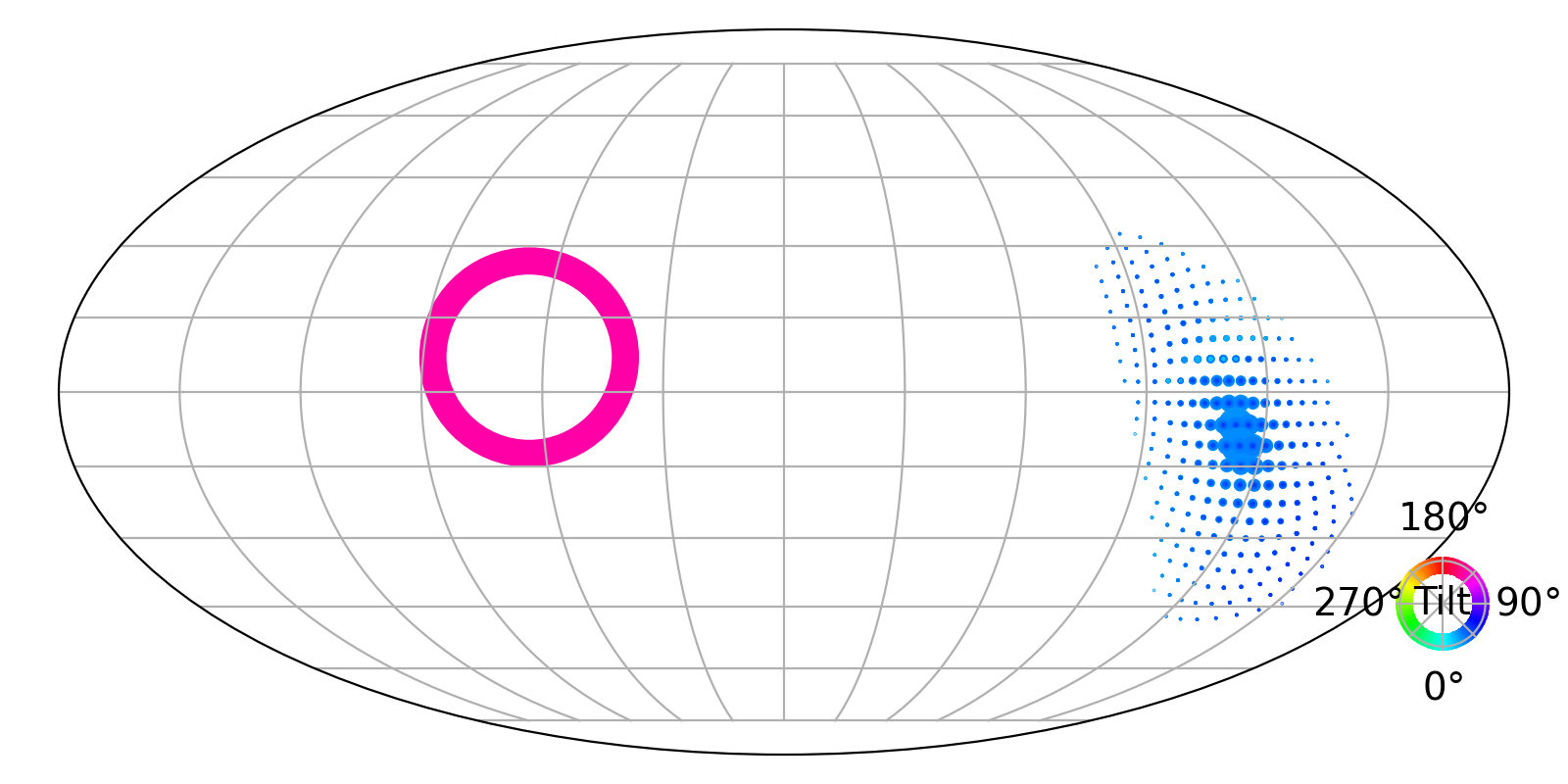}
    &\hspace{-4mm}
    \includegraphics[height=0.77cm]{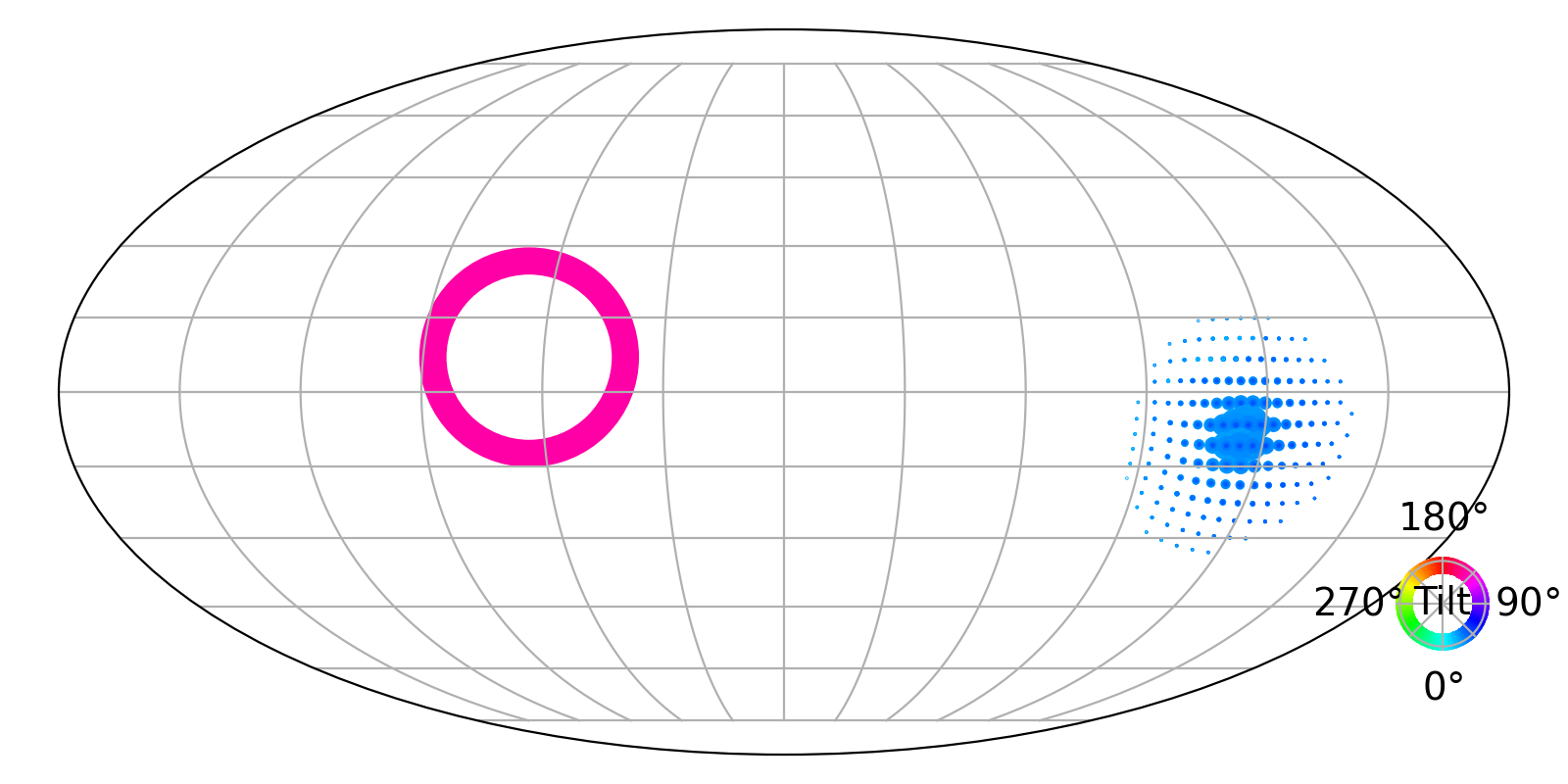} 
    &\hspace{-4mm}
    \includegraphics[height=0.77cm]{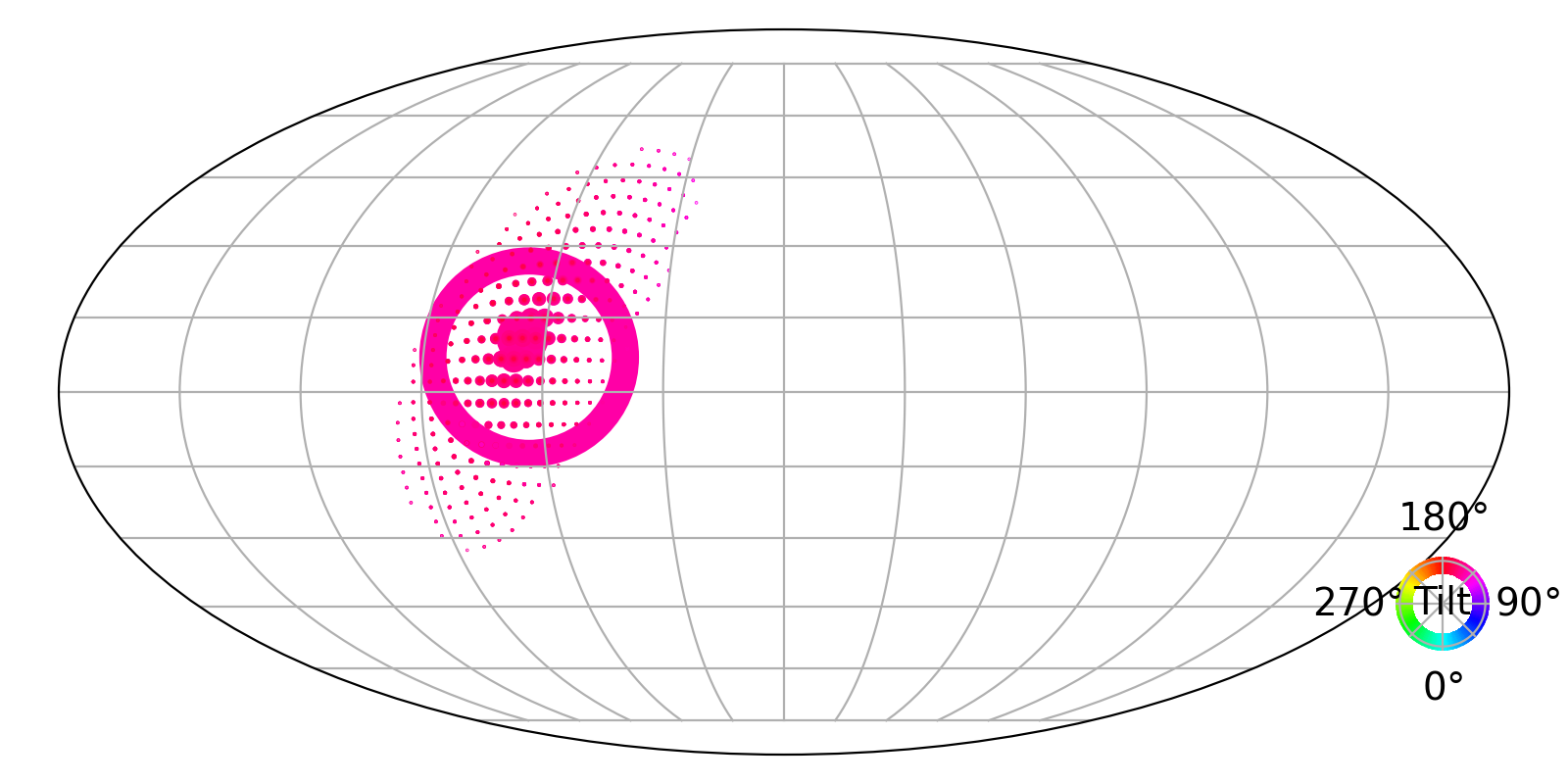}
    \hspace{-0mm}
    \\

    \hspace{-4mm}
    \includegraphics[height=1.4cm]{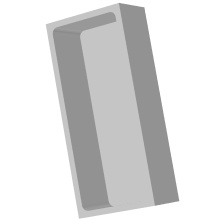} 
    &\hspace{-4mm}
    \includegraphics[height=0.77cm]{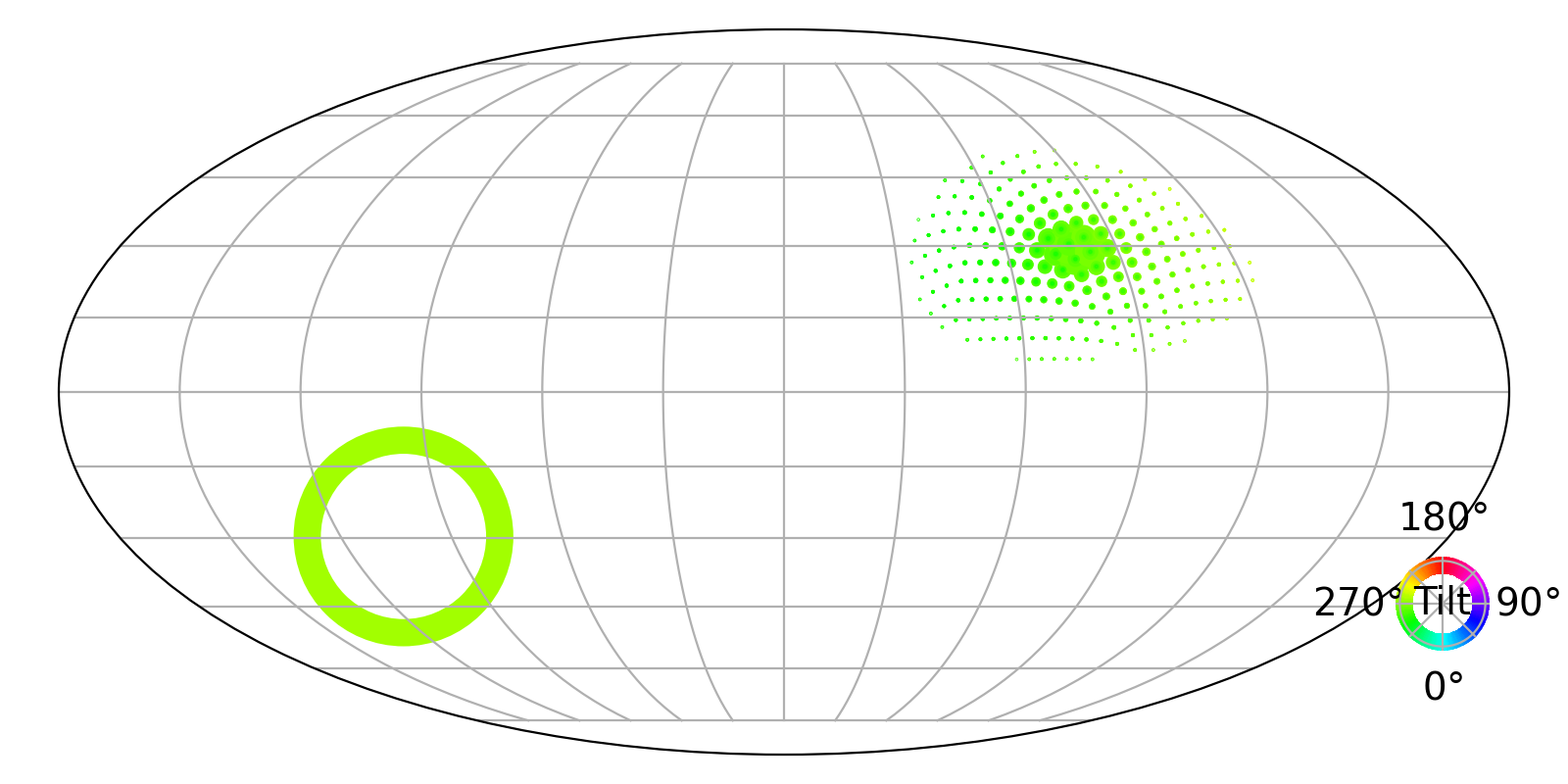}
    &\hspace{-4mm}
    \includegraphics[height=0.77cm]{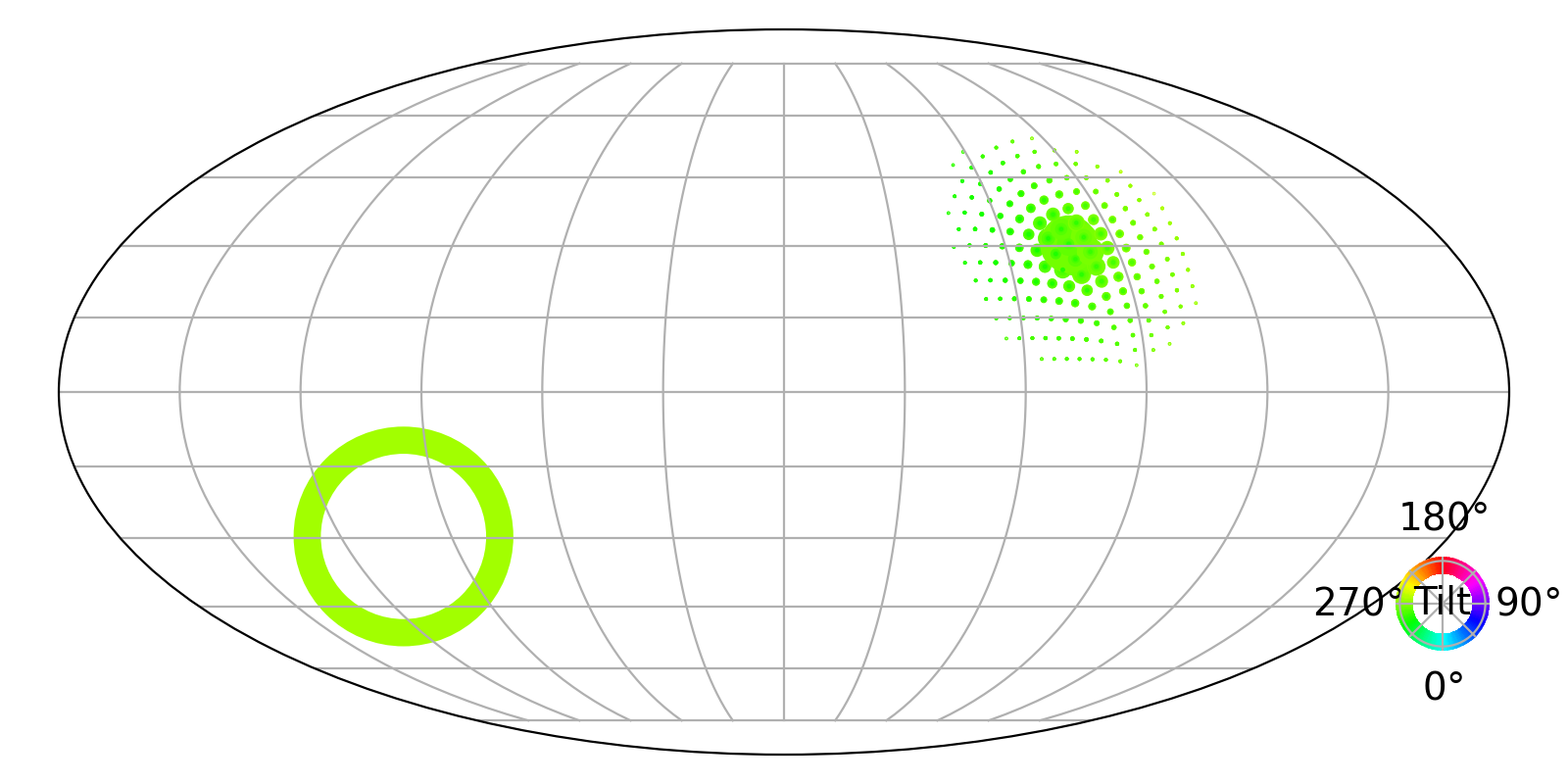} 
    &\hspace{-4mm}
    \includegraphics[height=0.77cm]{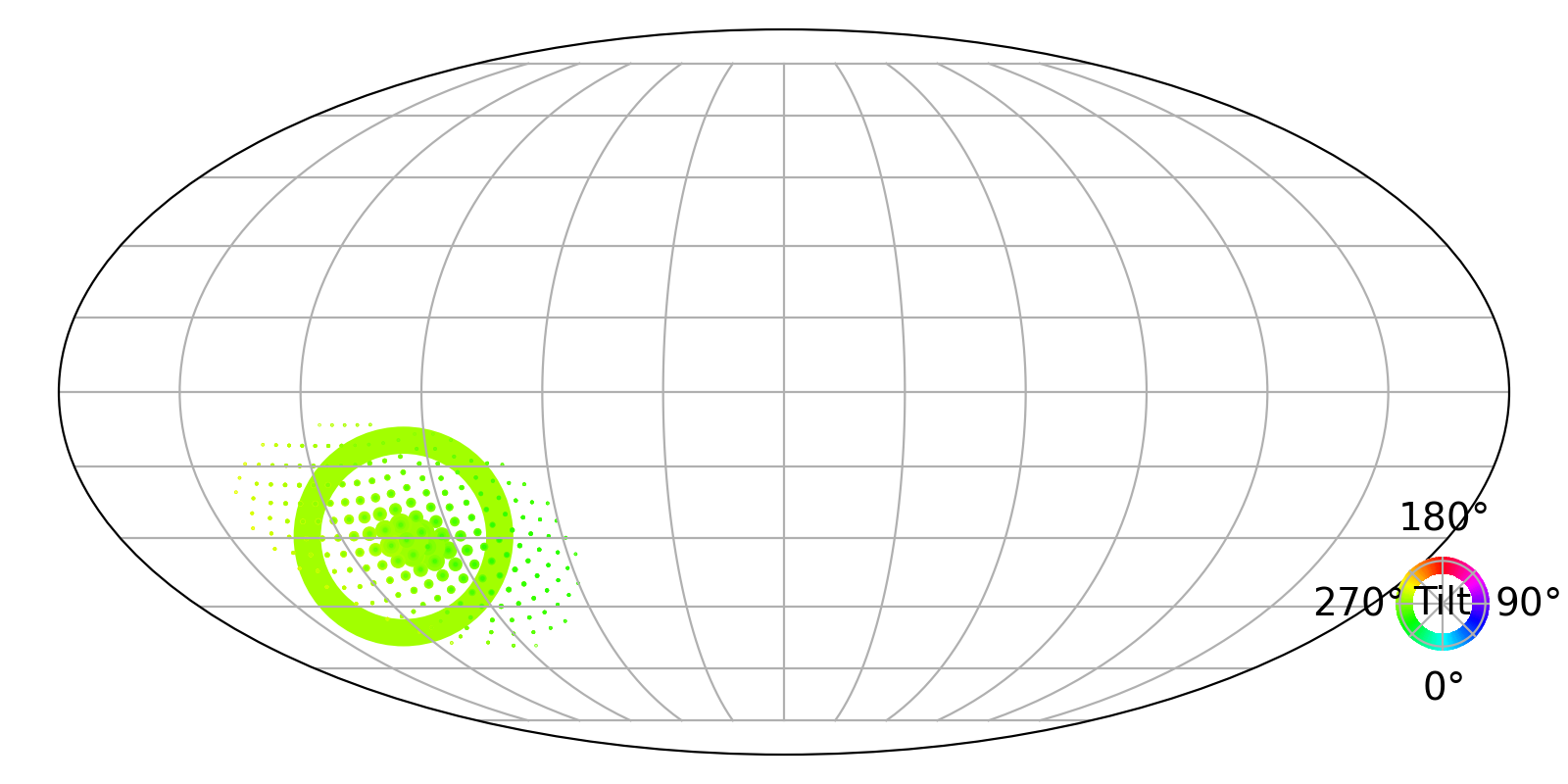}
    &\hspace{-4mm}
    \includegraphics[height=0.77cm]{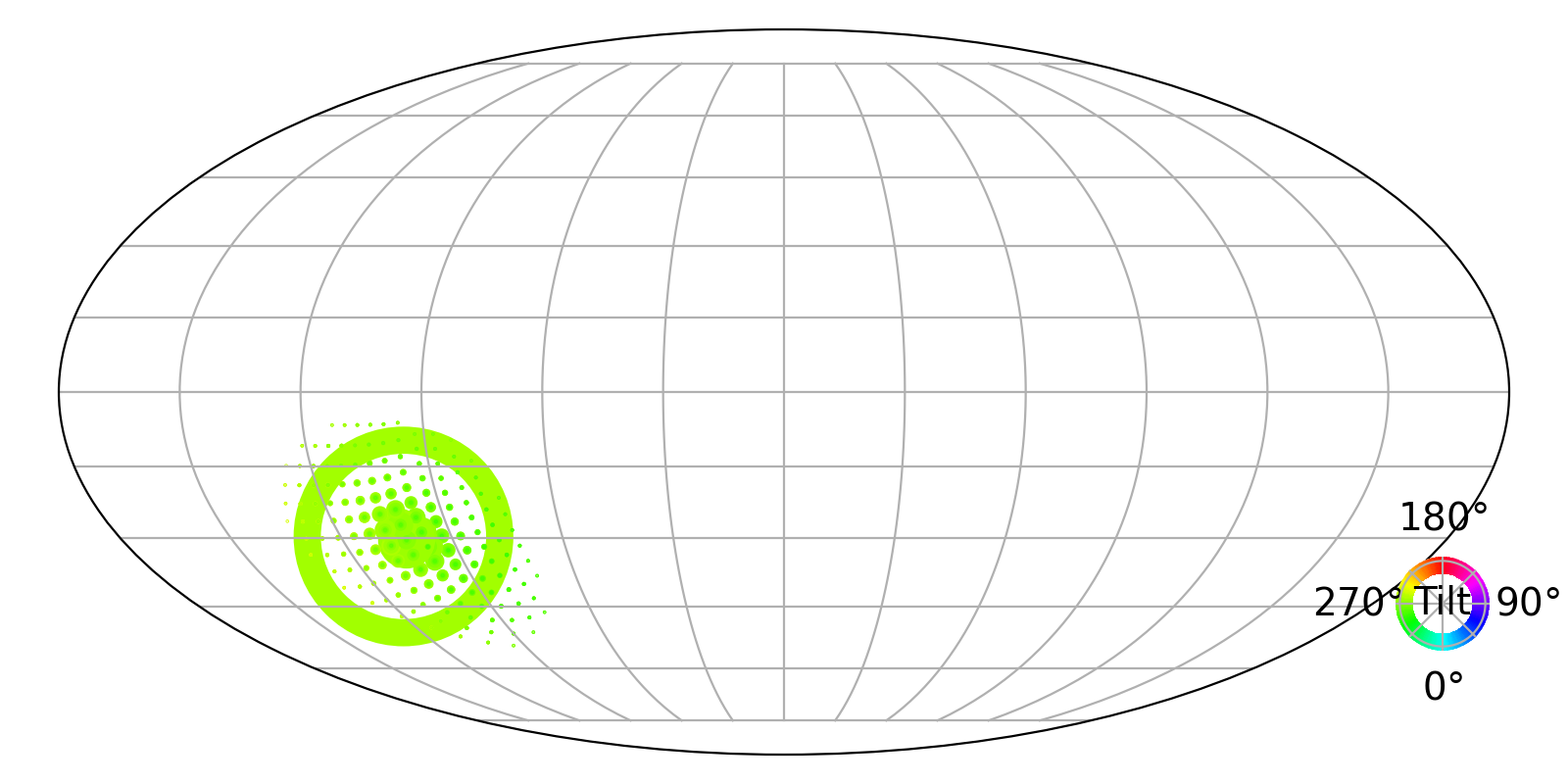}
    \hspace{-0mm}
    \\

    \hspace{-4mm}
    \includegraphics[height=1.4cm]{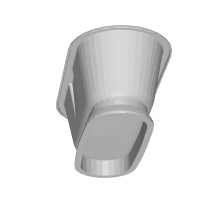}  
    &\hspace{-4mm}
    \includegraphics[height=0.77cm]{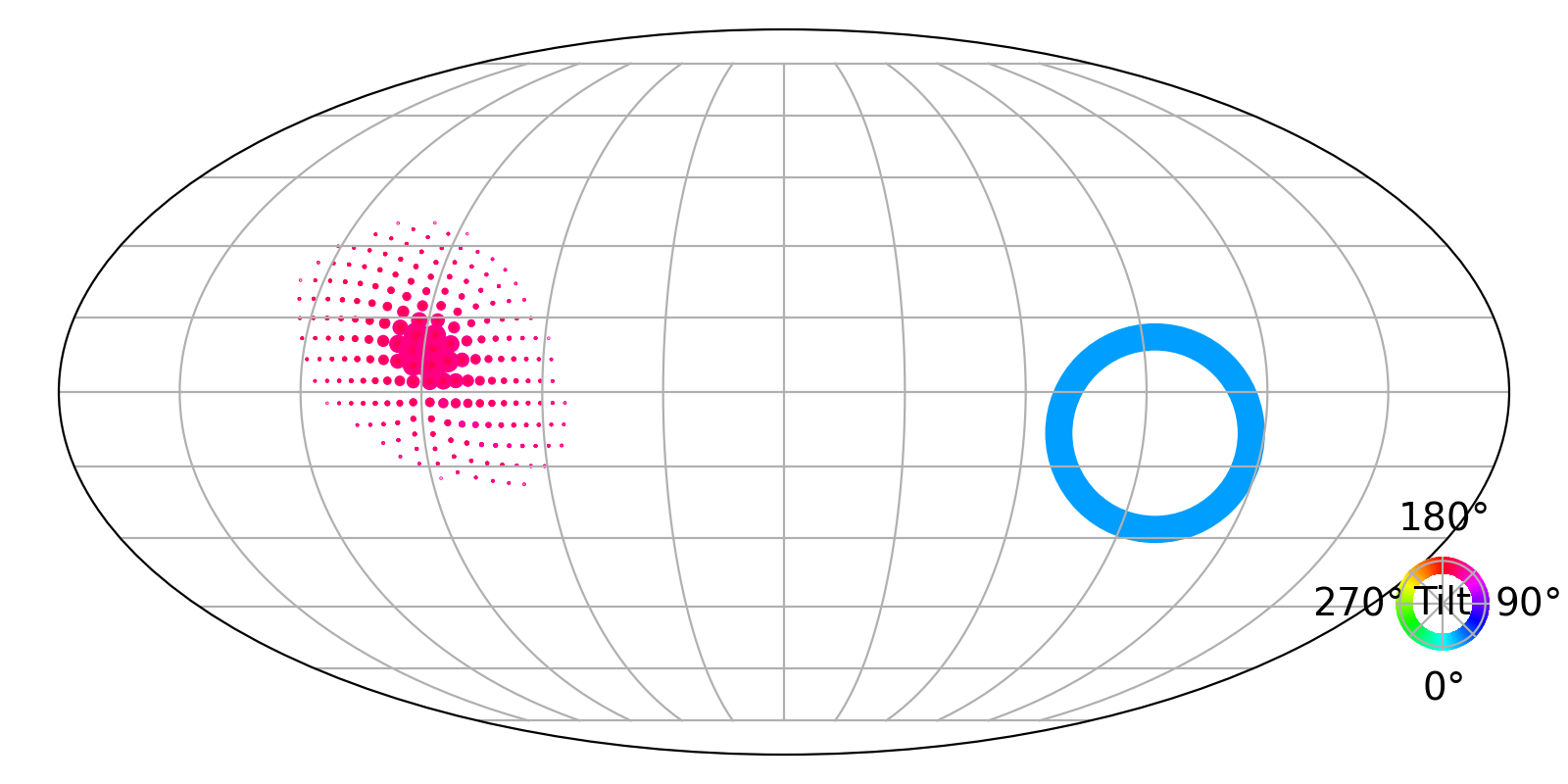} 
    &\hspace{-4mm}
    \includegraphics[height=0.77cm]{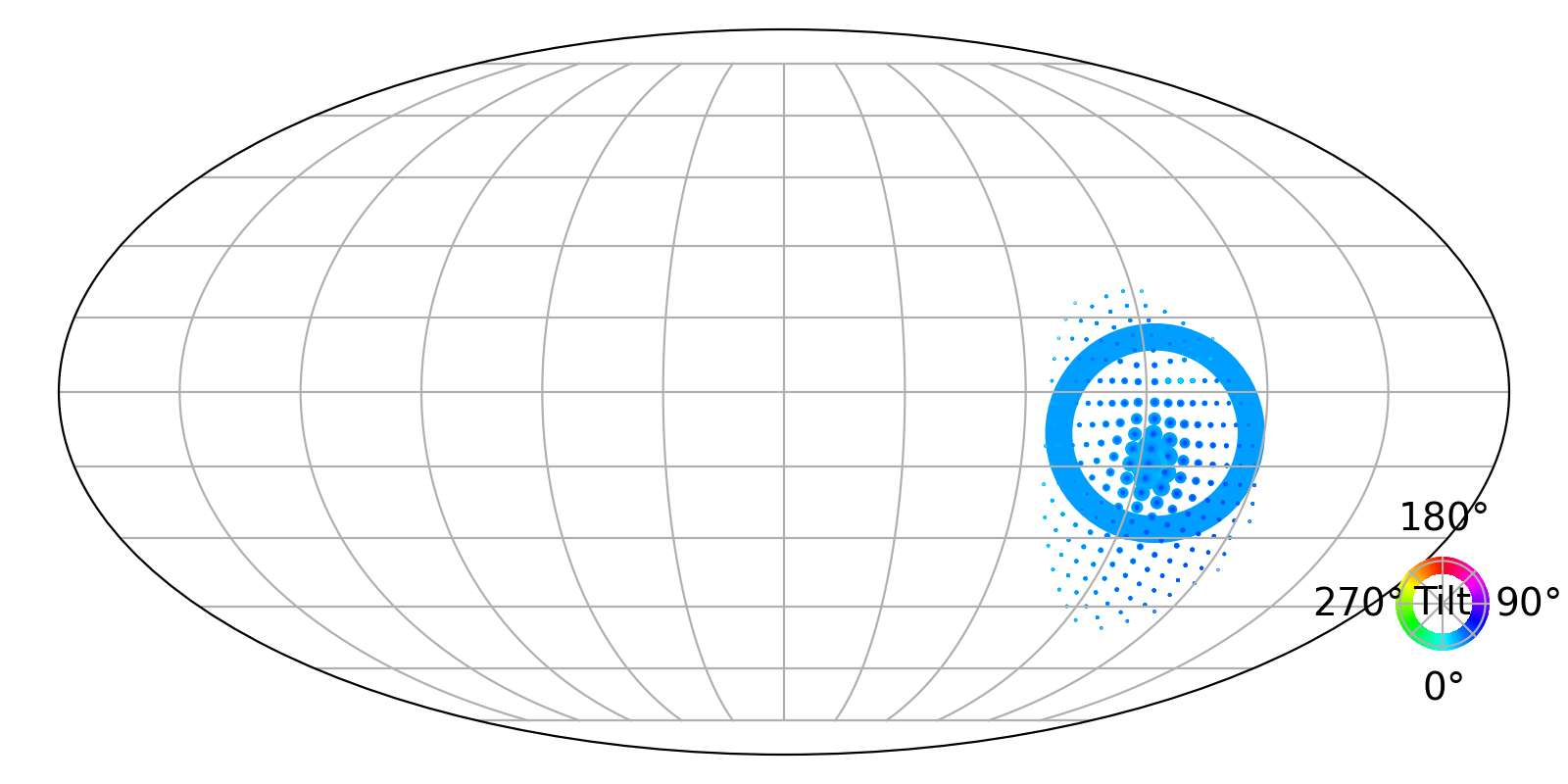} 
    &\hspace{-4mm}
    \includegraphics[height=0.77cm]{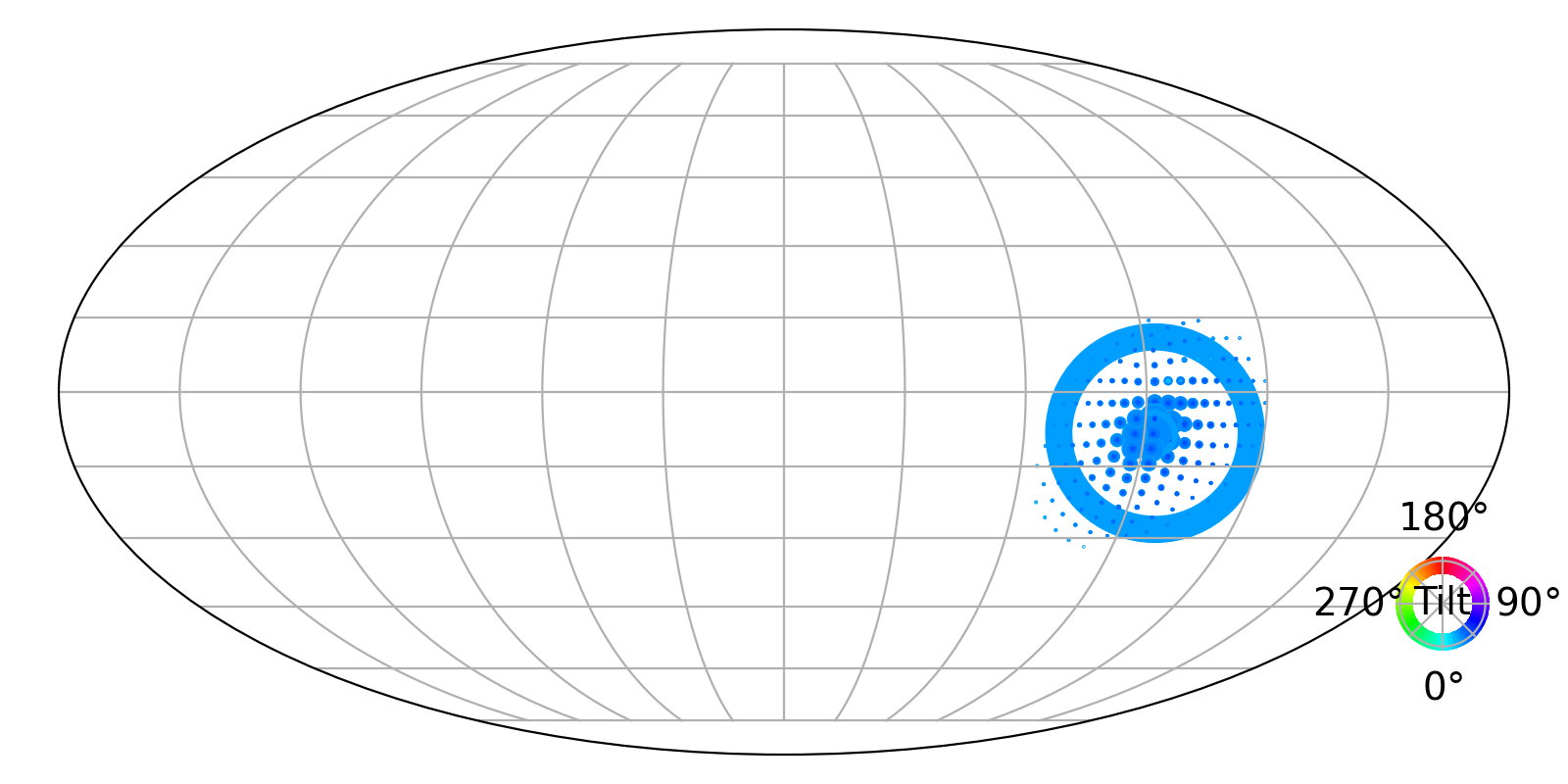} 
    &\hspace{-4mm}
    \includegraphics[height=0.77cm]{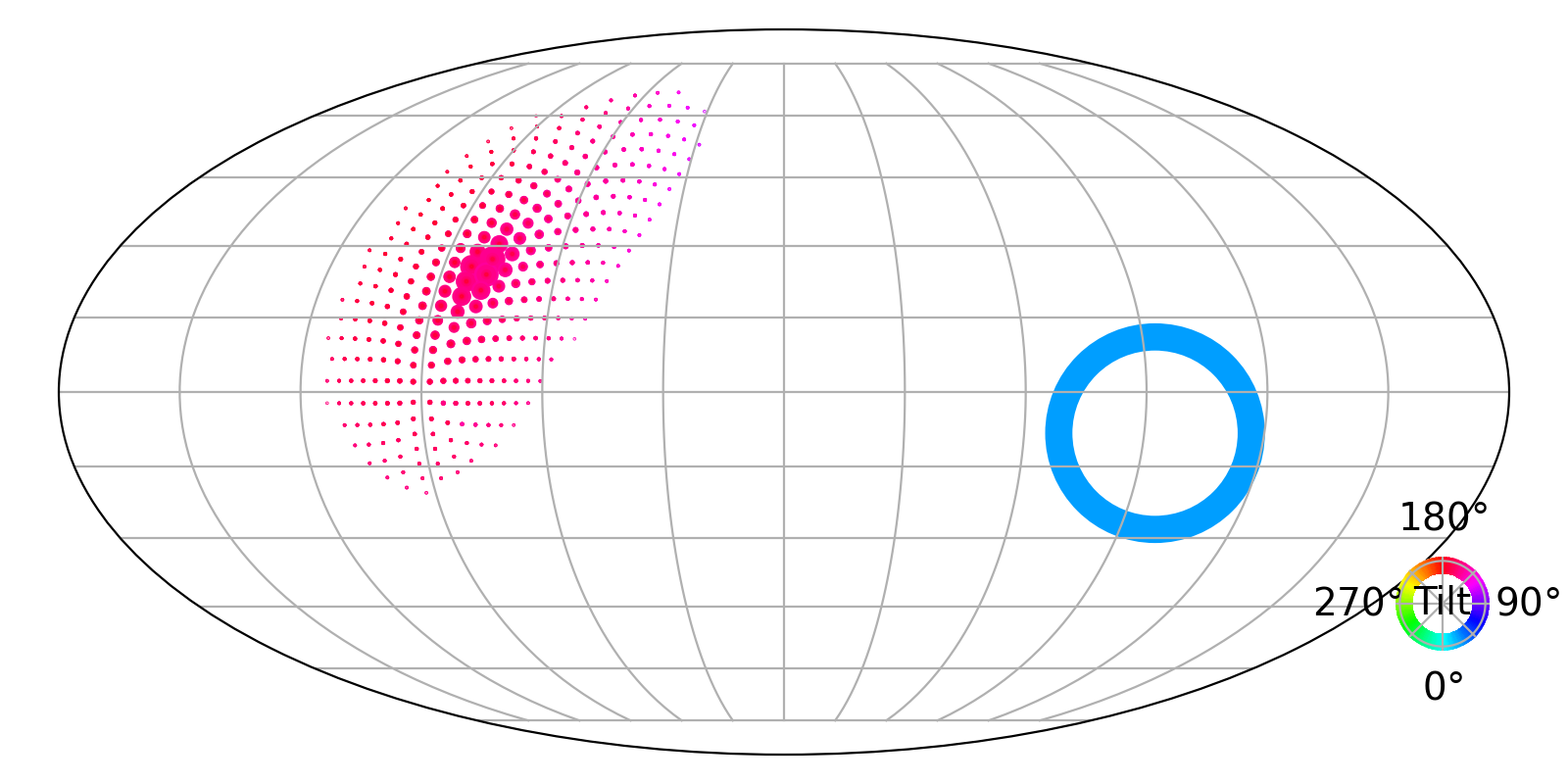}
    \hspace{-0mm}
    \\

    \hspace{-4mm}
    \includegraphics[height=1.4cm]{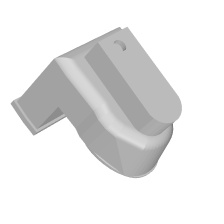} 
    &\hspace{-4mm}
    \includegraphics[height=0.77cm]{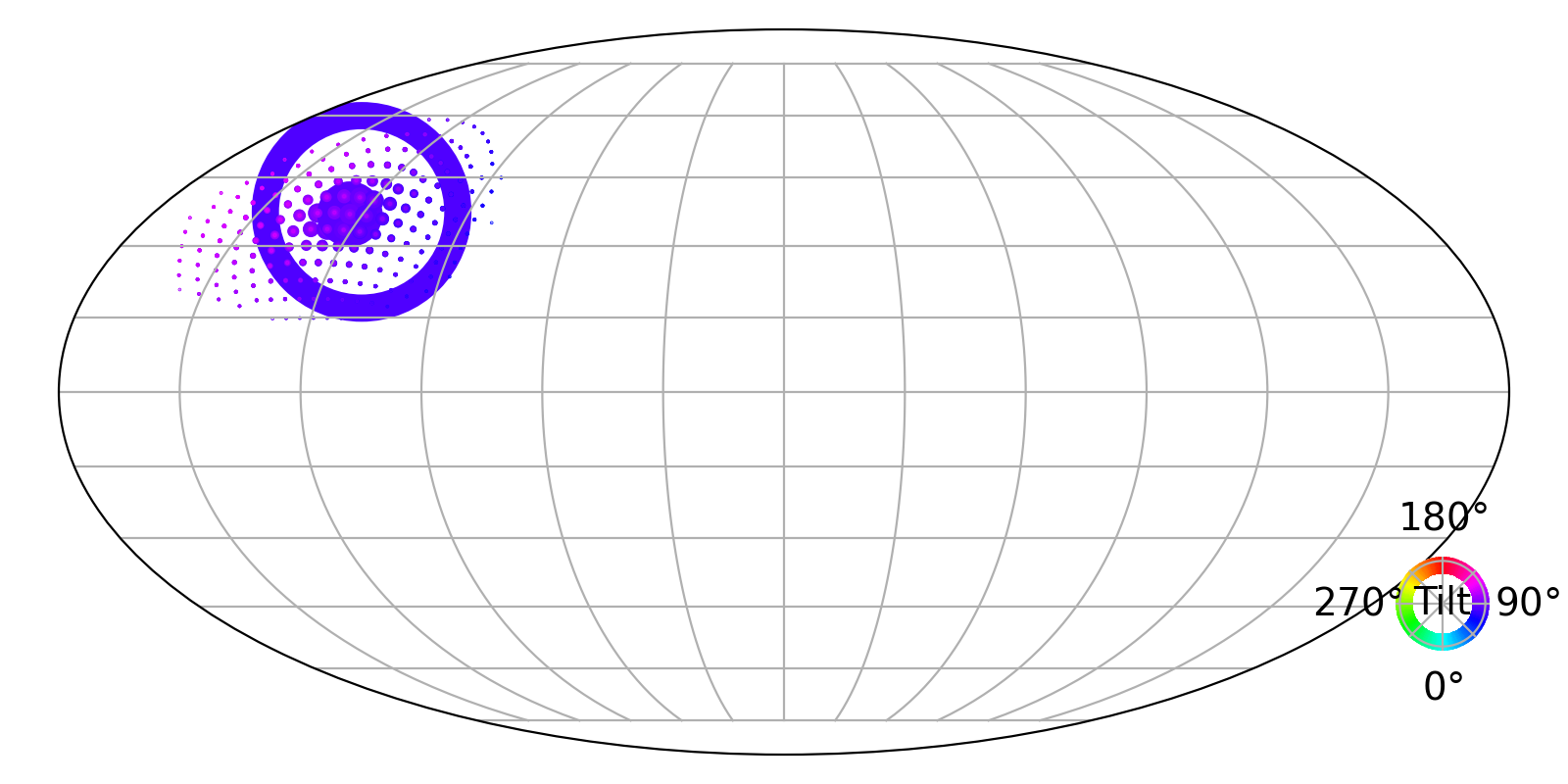} 
    &\hspace{-4mm}
    \includegraphics[height=0.77cm]{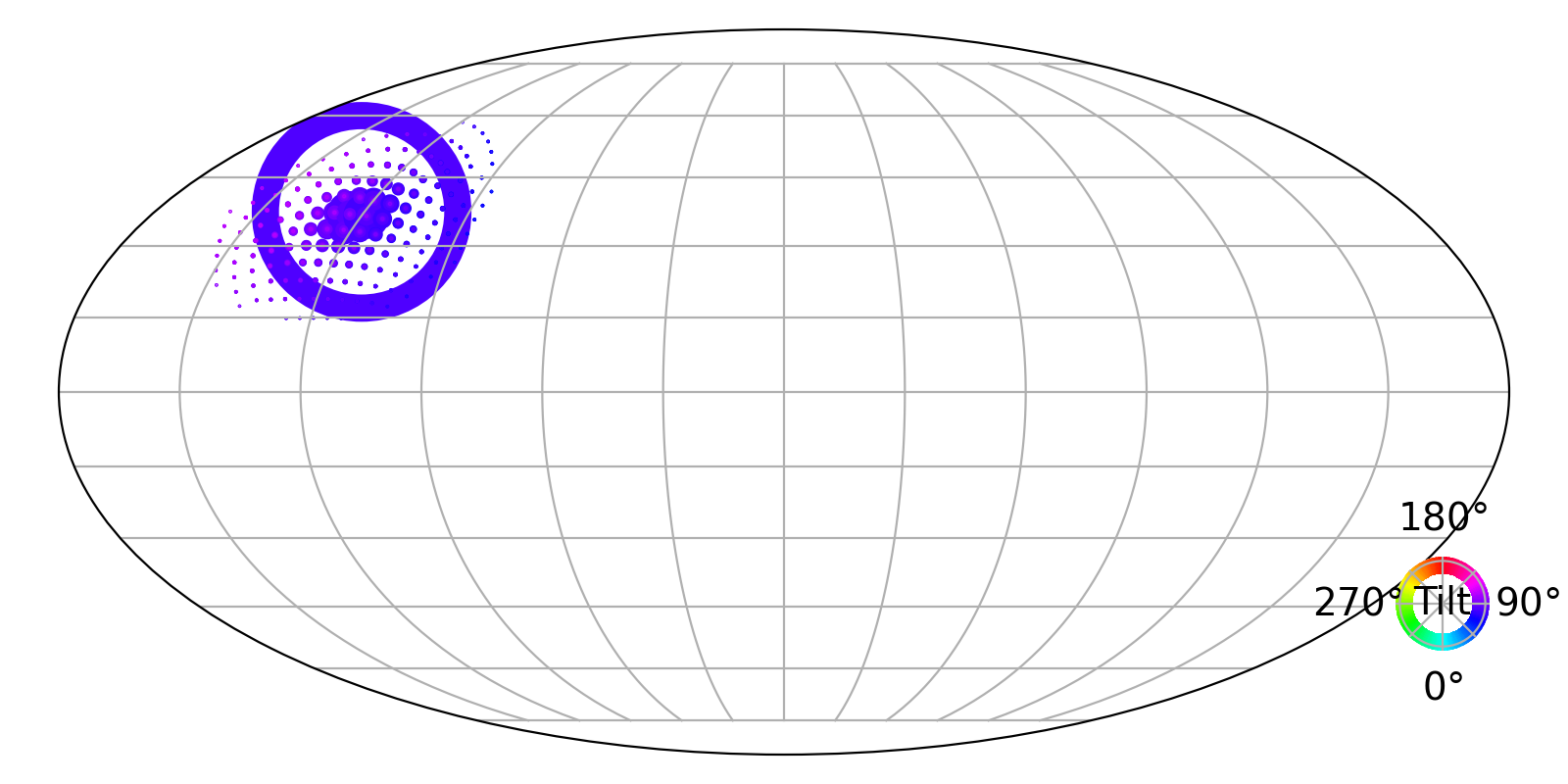} 
    &\hspace{-4mm}
    \includegraphics[height=0.77cm]{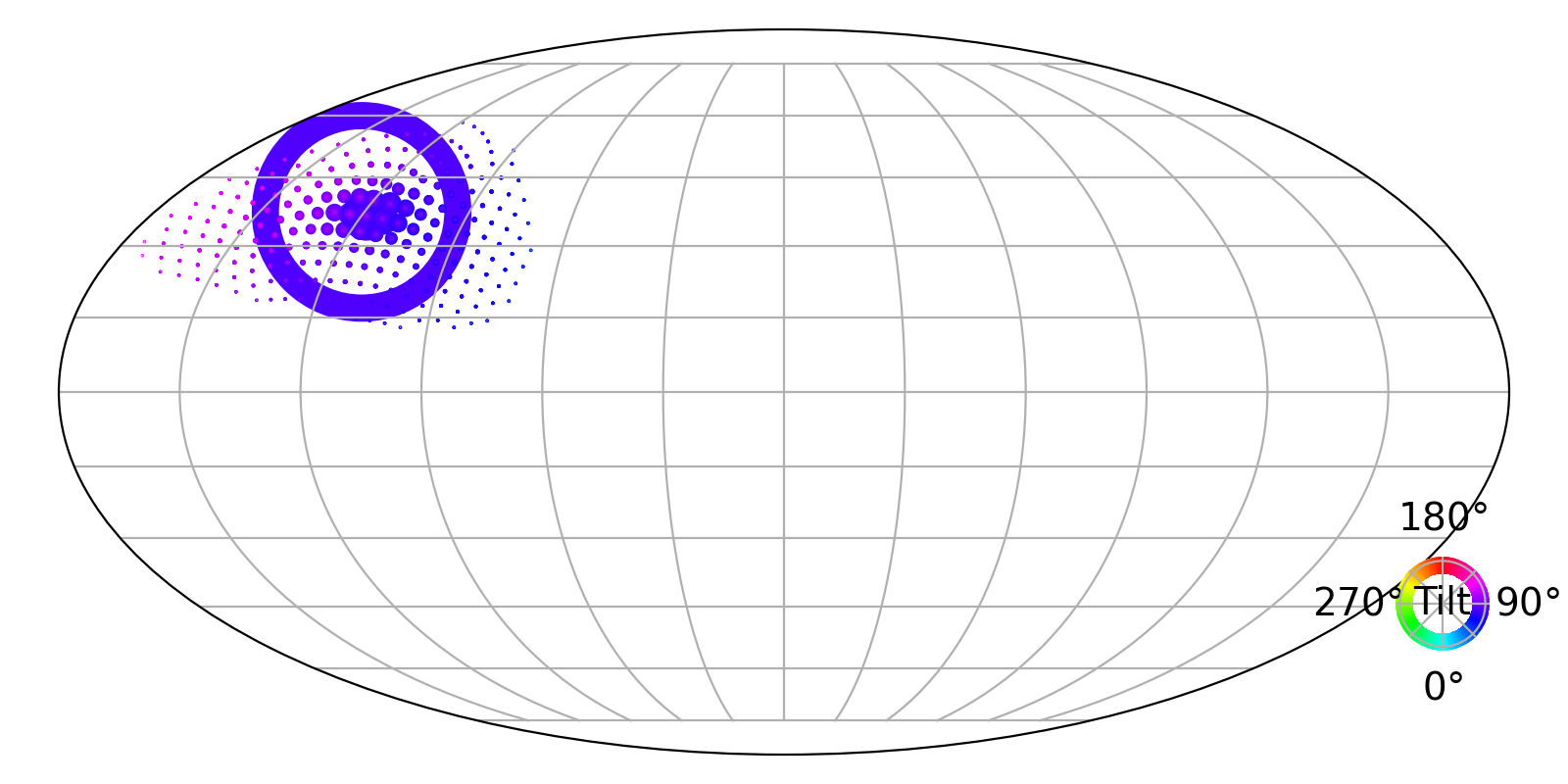} 
    &\hspace{-4mm}
    \includegraphics[height=0.77cm]{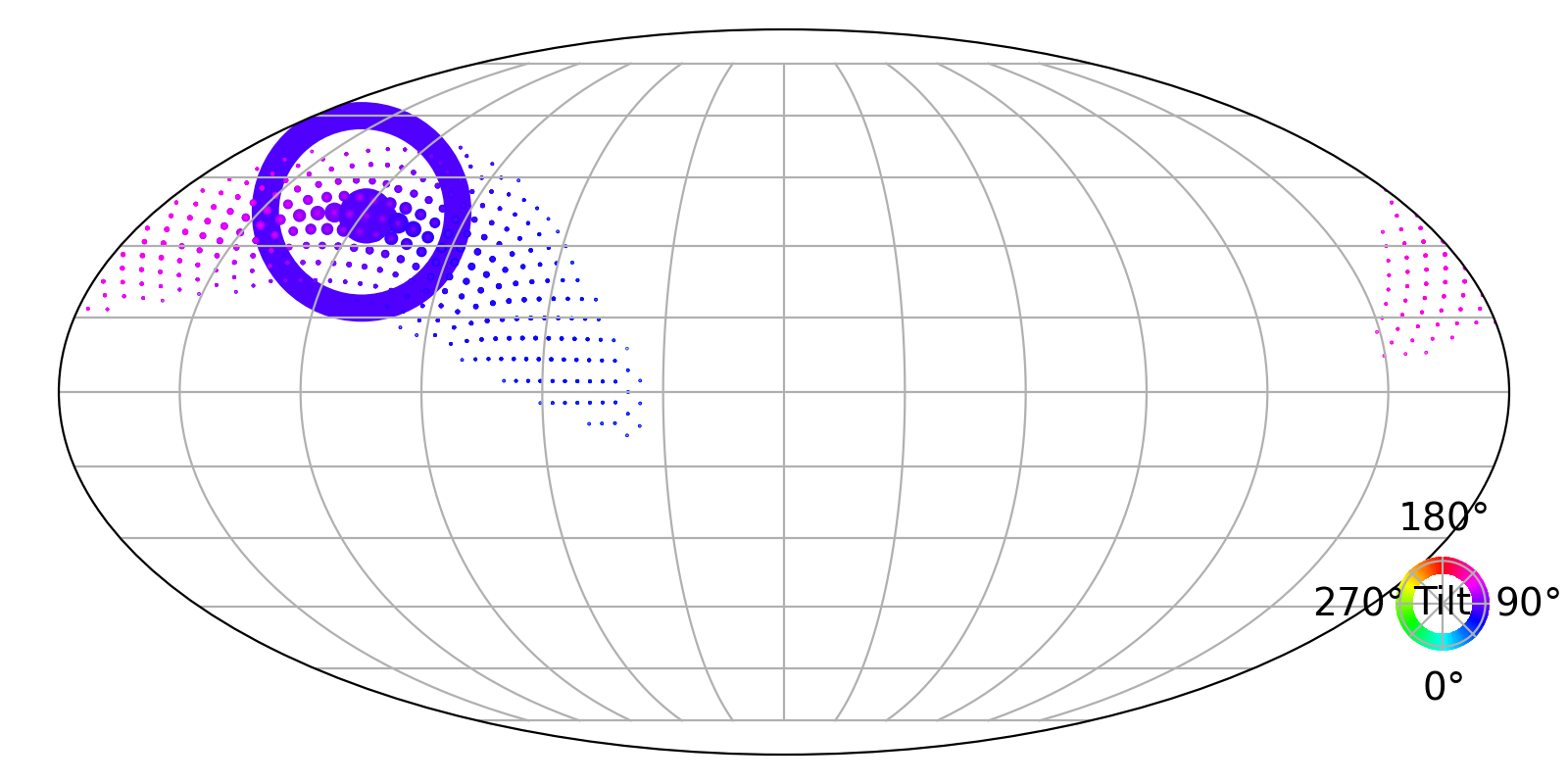}
    \hspace{-0mm}
    \\

    \hspace{-4mm}
    \includegraphics[height=1.6cm]{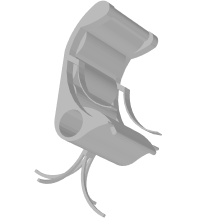} 
    &\hspace{-4mm}
    \includegraphics[height=0.77cm]{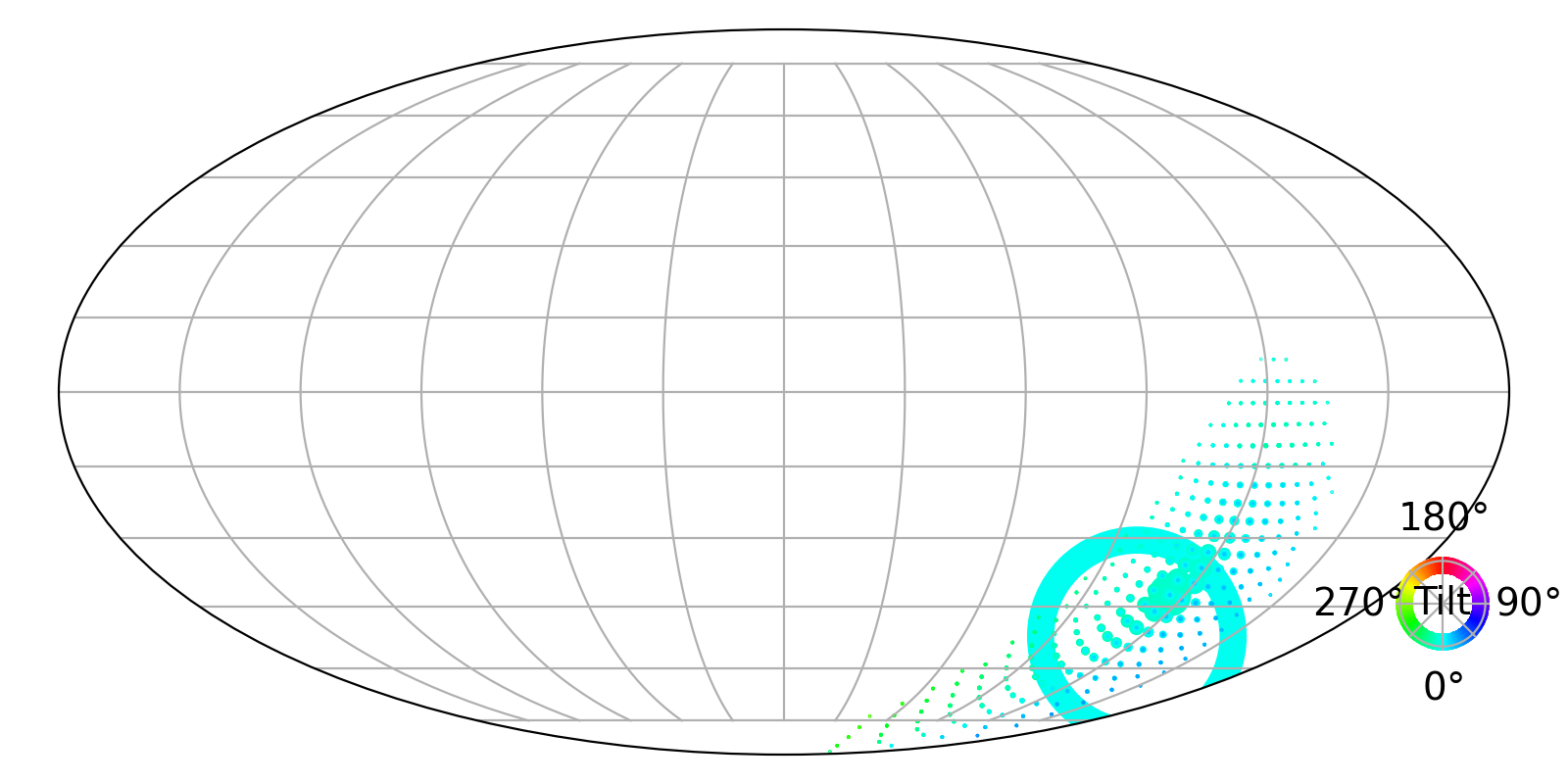} 
    &\hspace{-4mm}
    \includegraphics[height=0.77cm]{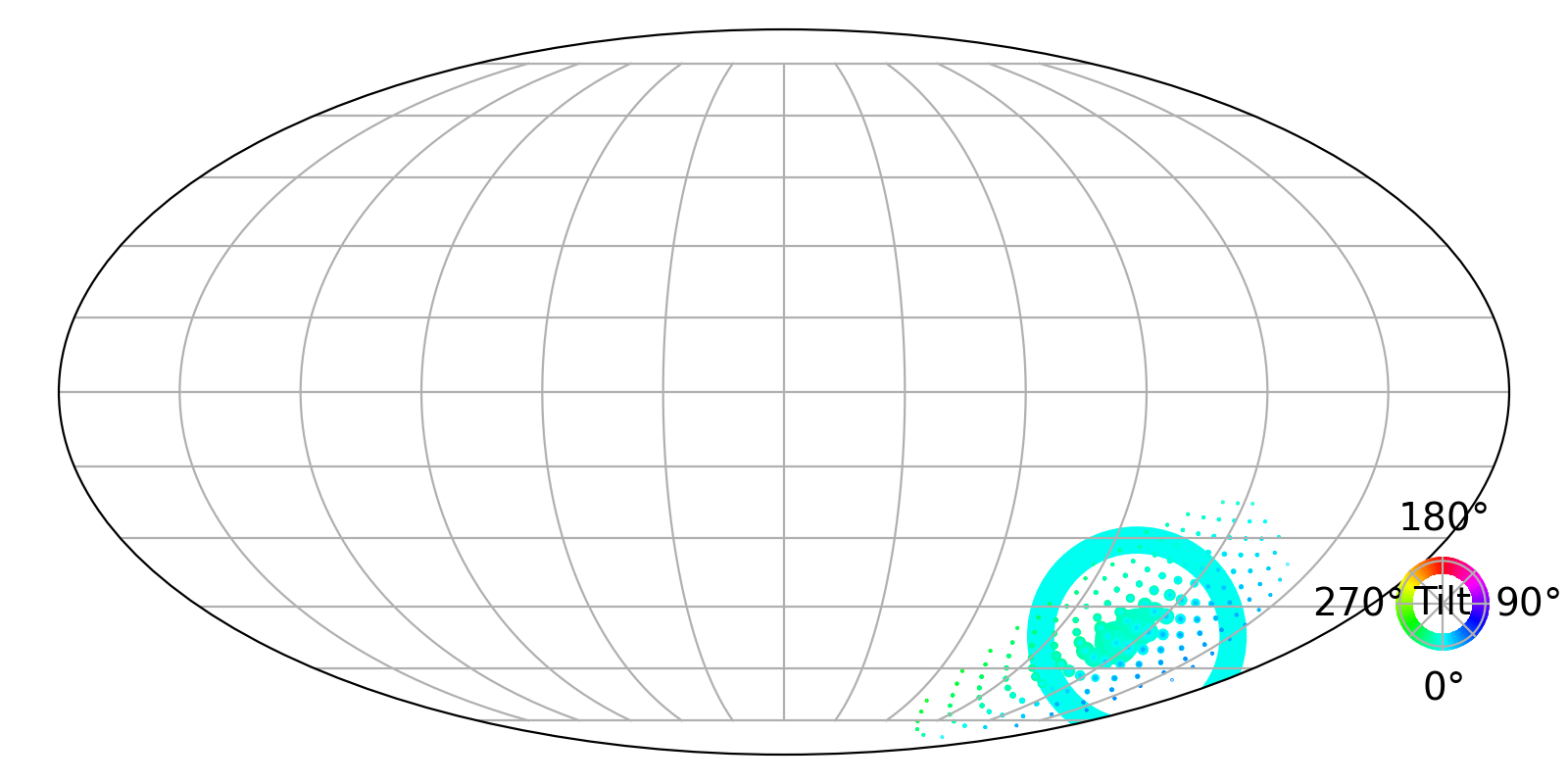} 
    &\hspace{-4mm}
    \includegraphics[height=0.77cm]{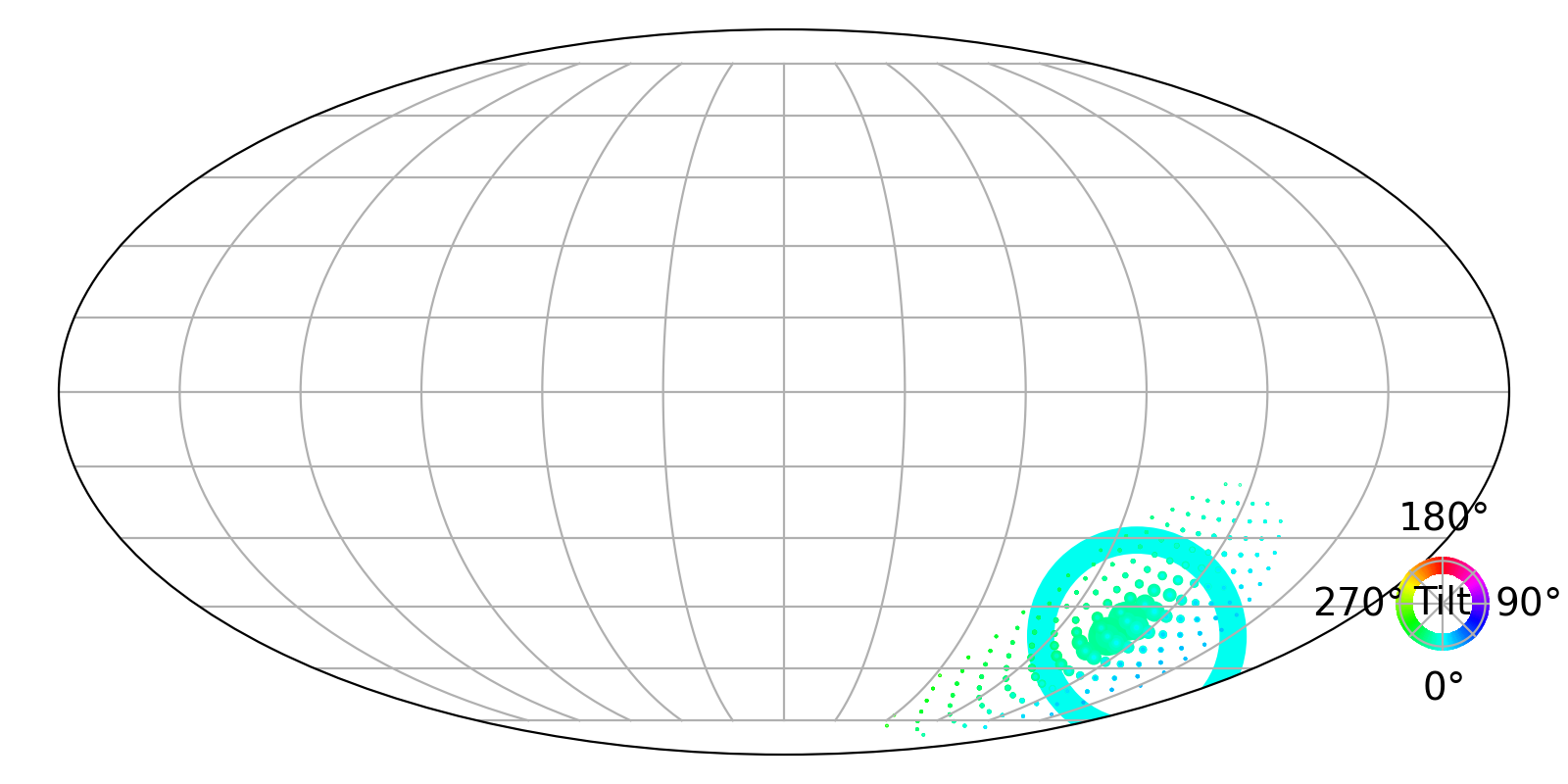} 
    &\hspace{-4mm}
    \includegraphics[height=0.77cm]{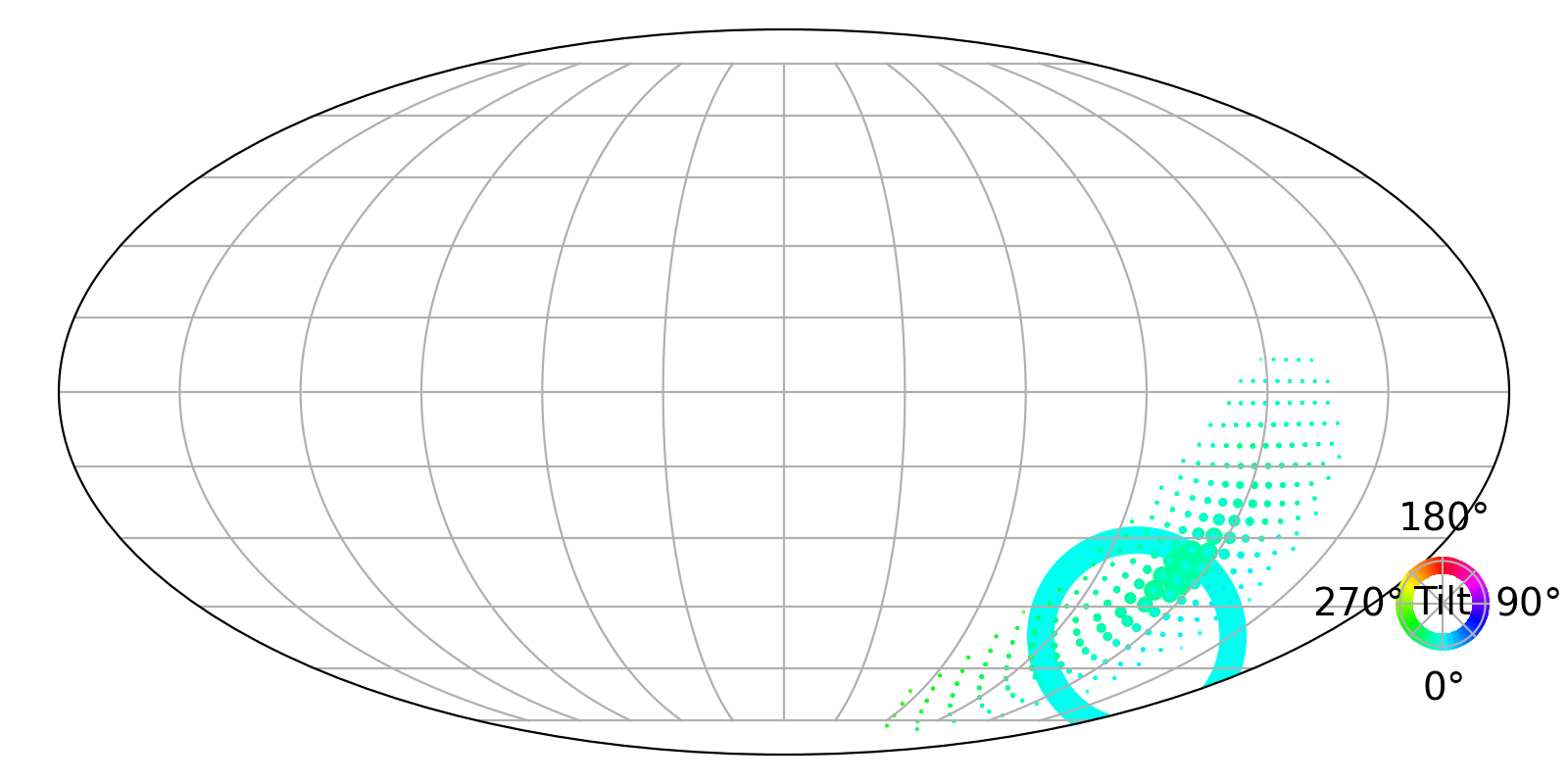}
    \hspace{-0mm}
    \\
    
    \small{Input image} 
    & \multicolumn{4}{c}{\small{Prediction of each branche}} 
    \\
    \end{tabular}
    \caption{\small Visualizations of the predicted distributions of rotation Laplace mixture model. The first column displays the input images and the remaining columns show predicted distributions of each branch.}
    \vspace{-2mm}
	\label{fig:mixture}
\end{figure}

In Figure \ref{fig:mixture}, we visualize the output distributions of each branch of rotation Laplace mixture model, and demonstrate how our model is able to capture different modes with ambiguous inputs. 
\revision{
We have included a popular visualization method utilized in \cite{murphy2021implicit}. This approach involves discretizing over $\SO$, projecting a great circle of points onto $\SO$ for each point on the 2-sphere, and using the color wheel to indicate the location on the great circle. The probability density is represented by the size of the points on the plot. We recommend consulting the corresponding papers for further details. 

Given our configuration of the mixture model with four modes, the visualizations are distributed across the four columns. The solid circles denote the ground truths.
}
Shown in the first three rows, for objects with rotational symmetries, e.g., bathtub, the multi-hypothesis predictions well present the plausible solution space of the input. While in cases where an object does not carry ambiguity, illustrated in the last two rows, different branches tend to agree with each other and correctly collapse to a single mode. 

\revision{
\subsubsection{Comparison with Min-of-N loss}
In this section, we experiment with min-of-N loss which is adopted in previous literature \cite{wang2021gdr} to handle symmetry issues.
The min-of-N loss considers the set of all possible ground truth rotations under symmetry and supervises the network prediction by selecting the ground truth with the minimum error, defined as:
\begin{equation}
    \mathcal{L}_{\text{min-of-N}} = \min_{\mathbf{R}\in\mathcal{R}} \mathcal{L}(\hat{\mathbf{R}}, \mathbf{R})
\end{equation}
where $\hat{\mathbf{R}}$ and $\mathbf{R}$ denote the prediction and the ground truth, respectively.

In our case, we manually compute the set of possible ground truths for symmetric objects. Specifically, we assume 2-symmetry for \texttt{bathtub}, \texttt{desk}, \texttt{dresser}, \texttt{night-stand} and \texttt{table} categories of ModelNet10-SO3 dataset.
{Due to the absence of object CAD models in our category-level rotation regression tasks, instead of min-of-N Point-Matching loss in \cite{wang2021gdr}, we experiment with min-of-N NLL loss.}
We compare the min-of-N loss and our multi-modal distribution in Table \ref{tab:min-of-N}.
For min-of-N loss, we report the better performance between the prediction and the two possible ground truths for symmetric objects. Similarly, for multi-modal distribution, we report the better result of the top-2 predictions.

\begin{table}[h]
  \centering
  \footnotesize
  \caption{\small Numerical comparisons with different methods to handle symmetry. The experiments are on ModelNet10-SO3 dataset averaged on all categories.}
    \begin{tabular}{lccc}
    \toprule
          & Acc@15$^\circ$$\uparrow$ & Acc@30$^\circ$$\uparrow$ & Med.($^\circ$)$\downarrow$  \\
    \midrule
    Min-of-N loss              &  0.844 & 0.874 & 3.1       \\
    Multi-modal distribution              & \textbf{ 0.900}  &  \textbf{0.918} &  \textbf{2.3}         \\
    \bottomrule
    \end{tabular}%
  \label{tab:min-of-N}%
\end{table}%

As depicted in the results, our multi-modal distribution consistently outperforms the min-of-N loss when handling symmetric objects. Moreover, in contrast to the min-of-N loss, which necessitates knowledge of the symmetry pattern to obtain all possible ground truths before training, the multi-modal distribution automatically generates different modes for ambiguous inputs from the data.

}}
\revision{\section{Limitation}

While rotation Laplace distribution offers advantages in robustness against the disturbance of outlier data, it is subject to several limitations.

Primarily, the probability density function of our distribution is relatively complex, which poses a significant challenge in deriving an analytical solution for the normalization factor and may limit its applicability across various downstream tasks. 
\remove{Moreover, due to the existence of the denominator in the probability density function, rotation Laplace distribution may exhibit numerical issues in training when the predicted mode is near the ground truth or the predicted $\mathbf{S}$ is near zero. Additional operations, e.g., clipping the denominator, are required to address the issues.}
\retwo{Moreover, the distribution suffers from singularity around the mode in two aspects.
When $\mathbf{S}$ is close to zero, the distribution becomes ill-defined, leading to difficulties in fitting distributions with very large uncertainties, such as the uniform distribution. When the predicted mode is close to the ground truth, the probability density tends to become infinite, which may result in unstable training.
}
\yingda{Refining the computation of the normalization constant or introducing a more elegant form of distribution can be important future directions.}}
\section{Conclusion}
In this paper, we draw inspiration from multivariant Laplace distribution and derive a novel distribution for probabilistic rotation regression, namely, rotation Laplace distribution.
We demonstrate that our distribution is robust to the disturbance of both outliers and small noises, thus achieving significantly superior performance on supervised and semi-supervised rotation regression tasks over all the baselines. We also extend rotation Laplace distribution to rotation Laplace mixture model to better capture the multi-modal rotation space.
Extensive comparisons with both probabilistic and non-probabilistic baselines demonstrate the effectiveness and advantages of our proposed distribution.


\ifCLASSOPTIONcompsoc
  \section*{Acknowledgments}
\else
  \section*{Acknowledgment}
\fi

This work is supported in part by National Key R\&D Program of China 2022ZD0160801.

\ifCLASSOPTIONcaptionsoff
  \newpage
\fi



\bibliographystyle{IEEEtran}
\bibliography{main}
%



%

\begin{IEEEbiography}[{\includegraphics[width=1in,height=1.25in,clip,keepaspectratio]{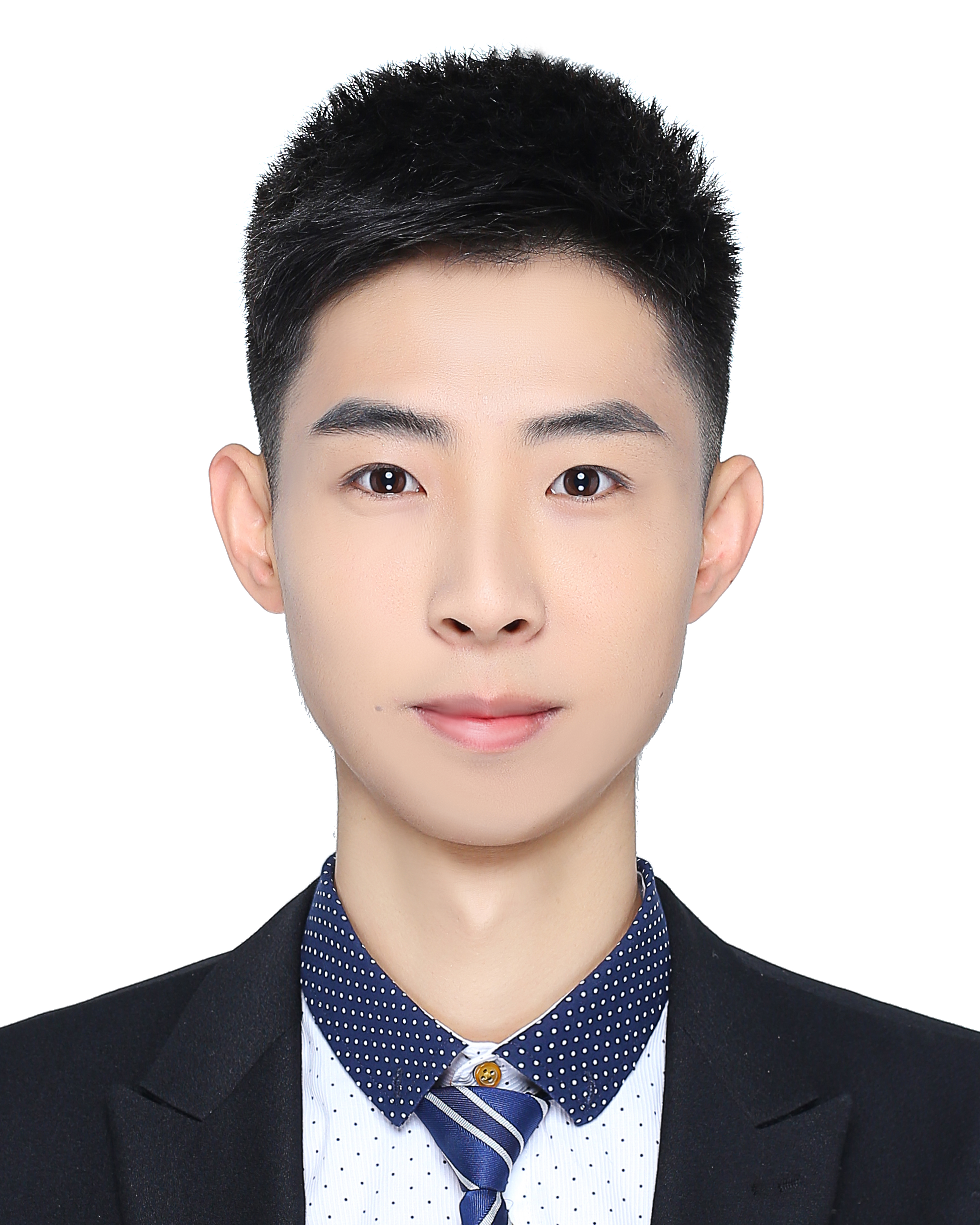}}]{Yingda Yin}
is a Ph.D. student at School of Computer Science, Peking University. His research interests lie in 3D computer vision, including object pose estimation, camera localization and 3D segmentation.
\end{IEEEbiography}

\begin{IEEEbiography}[{\includegraphics[width=1in,height=1.25in,clip,keepaspectratio]{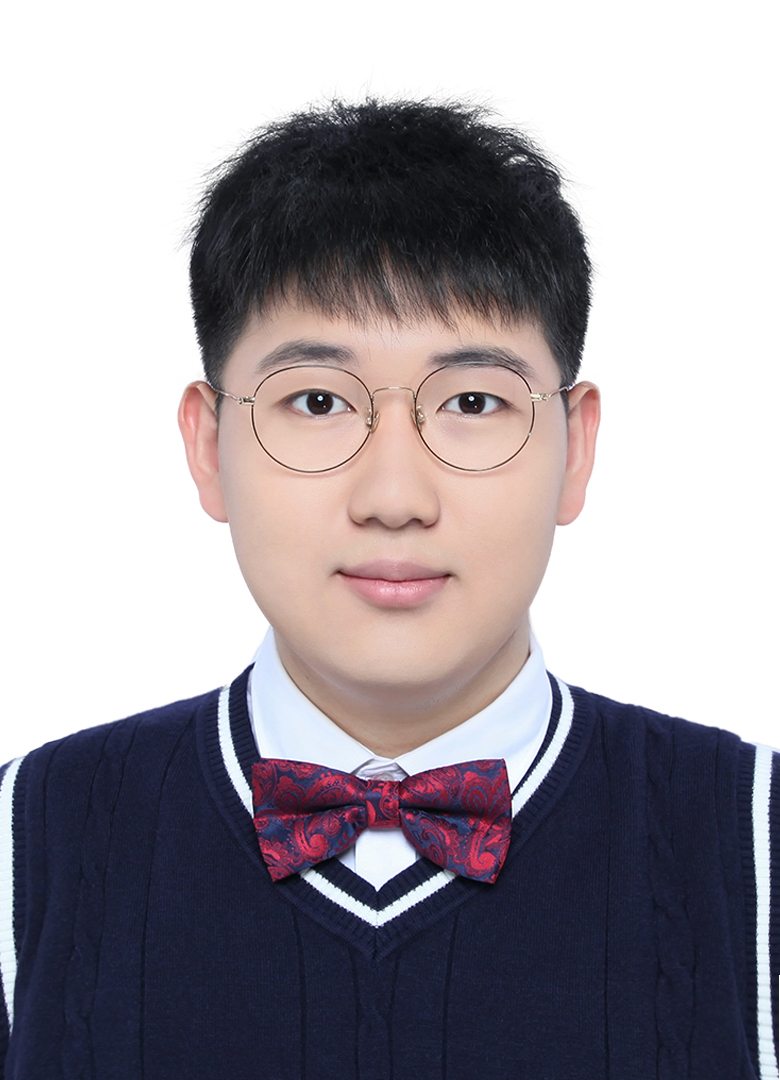}}]{Jiangran Lyu} is currently working
toward the PhD degree in computer science at
Peking University. His
current research span 3D computer vision and robotics, with a special interest in embodied AI.
\end{IEEEbiography}

\begin{IEEEbiography}[{\includegraphics[width=1in,height=1.25in,clip,keepaspectratio]{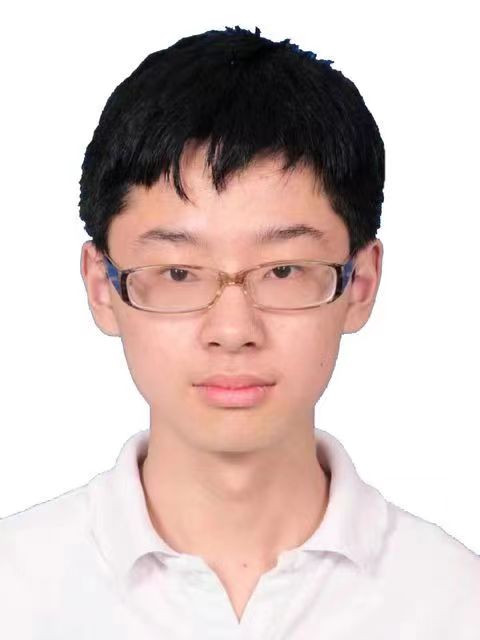}}]{Yang Wang} is an undergraduate student at School of EECS, Peking University. His research interests are focused on the field of 3D computer vision, including rotation regression, object pose estimation and 3D reconstruction.
\end{IEEEbiography}

\begin{IEEEbiography}[{\includegraphics[width=1in,height=1.25in,clip,keepaspectratio]{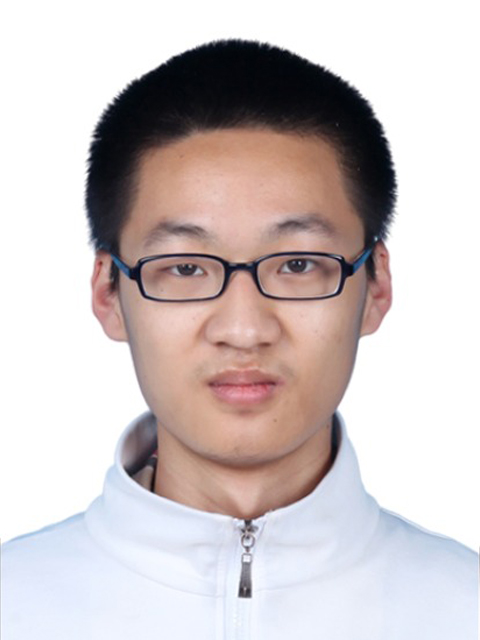}}]{Haoran Liu} is an undergraduate student at School of EECS, Peking University. His research interests are focused on the field of 3D computer vision and probabilistic modeling.
\end{IEEEbiography}

\begin{IEEEbiography}
[{\includegraphics[width=1in,height=1.25in,clip,keepaspectratio]{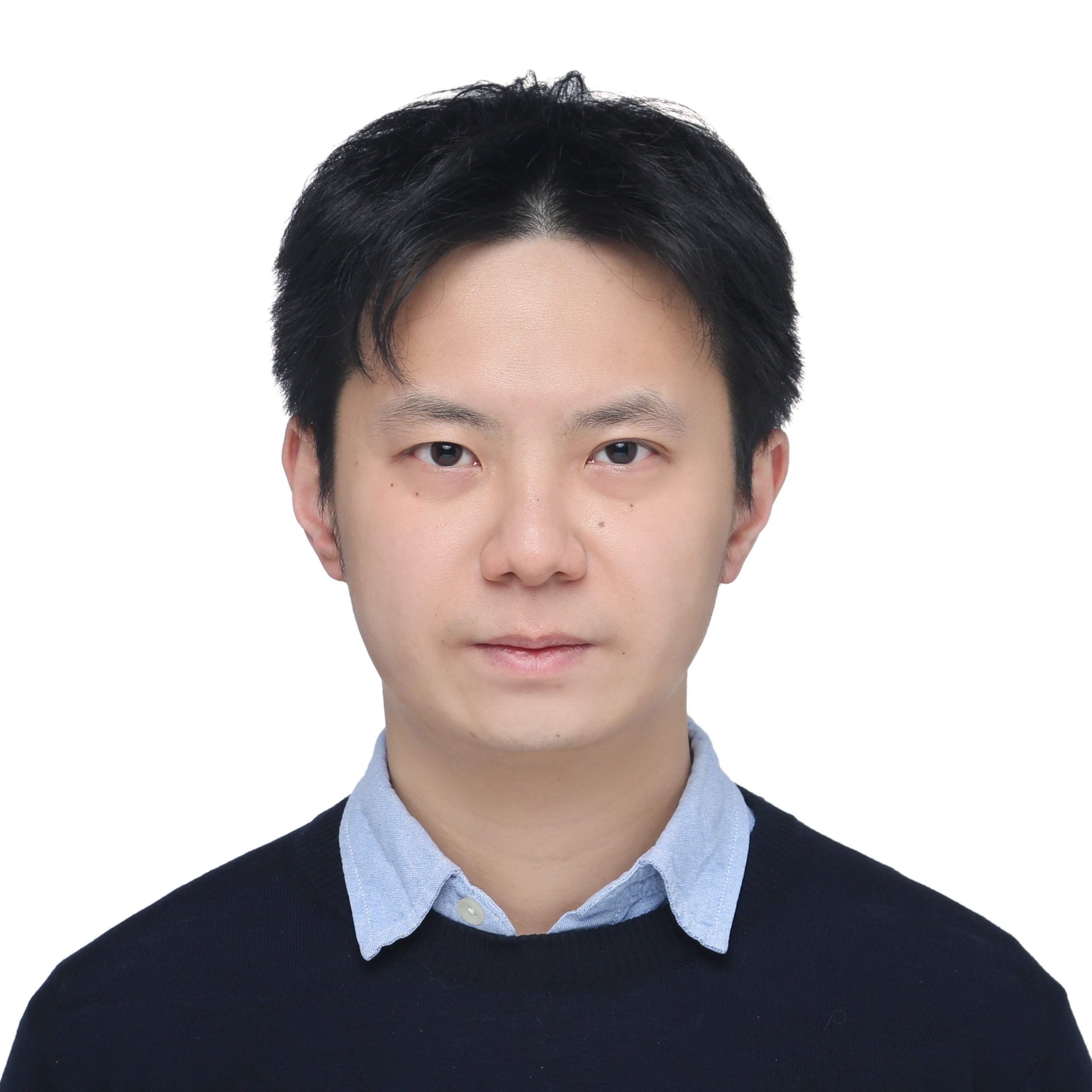}}]{He Wang} is a tenure-track assistant professor in the Center on Frontiers of Computing Studies (CFCS) at Peking University, where he founds and leads Embodied Perception and InteraCtion (EPIC) Lab. His research interests span 3D vision, robotics, and machine learning, with a special focus on embodied AI. His research objective is to endow robots working in complex real-world scenes with generalizable 3D vision and interaction policies in a scalable way. He has published more than 40 papers on top conferences and journals of computer vision, robotics and learning, including CVPR/ICCV/ECCV/TRO/ICRA/IROS/NeurIPS/ICLR/AAAI. His pioneering work on category-level 6D pose estimation, NOCS, receives 2022 World Artificial Intelligence Conference Youth Outstanding Paper (WAICYOP) Award and his work also receives ICRA 2023 outstanding manipulation finalist and Eurographics 2019 best paper honorable mention. He serves as an associate editor of Image and Vision Computing and serves area chairs in CVPR 2022 and WACV 2022. Prior to joining Peking University, he received his Ph.D. degree from Stanford University in 2021 under the advisory of Prof. Leonidas J. Guibas and his Bachelor's degree in 2014 from Tsinghua University.
\end{IEEEbiography}

\begin{IEEEbiography}[{\includegraphics[width=1in,height=1.25in,clip,keepaspectratio]{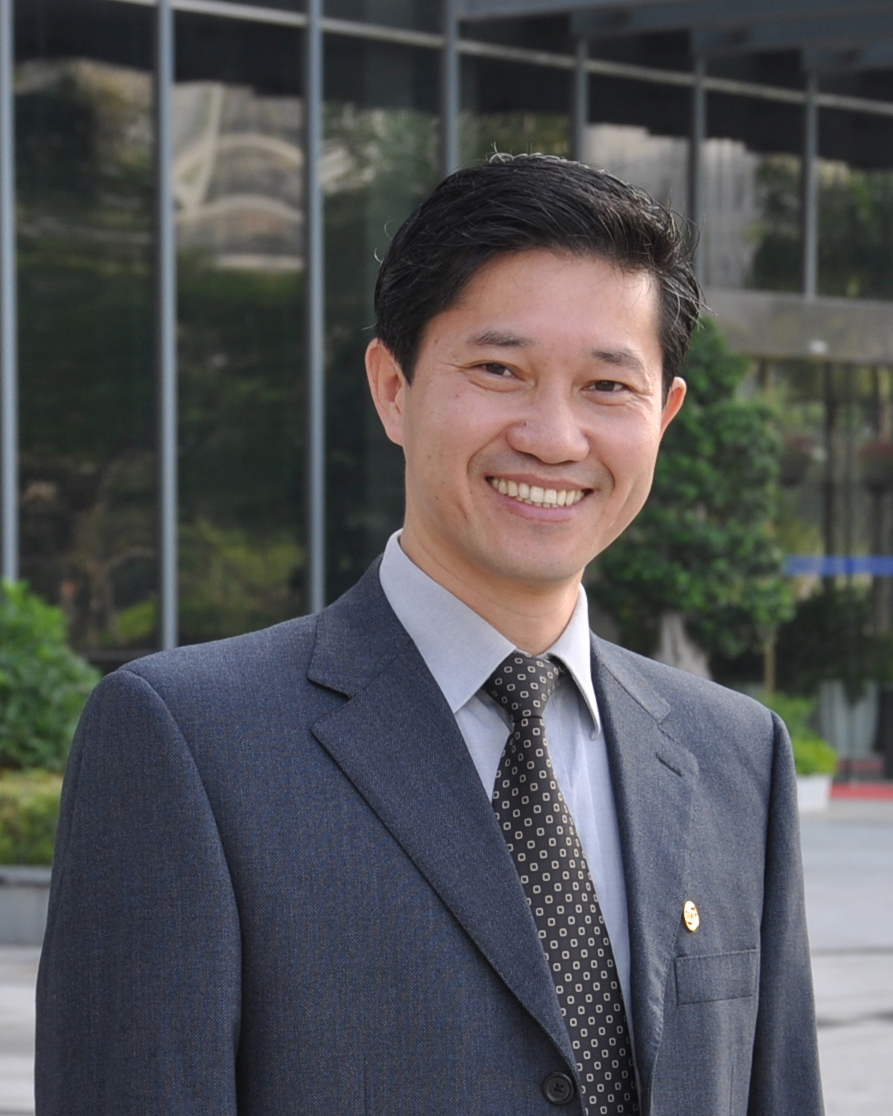}}]{Baoquan Chen} is a Professor of Peking University, where he is the Associate Dean of the School of Artificial Intelligence. For his contribution to spatial data (modeling) and visualization, he was elected IEEE Fellow 2020. His research interests generally lie in computer graphics, visualization, and human-computer interaction, focusing specifically on large-scale city modeling, simulation and visualization. He has published more than 200 papers in international journals and conferences, including 40+ papers in ACM Transactions on Graphics (TOG)/SIGGRAPH/SIGGRAPH Asia. Chen serves/served as associate editor of ACM TOG/IEEE Transactions on Visualization and Graphics (TVCG), and has served as conference steering committee member (ACM SIGGRAPH Asia, IEEE VIZ), conference chair (SIGGRAPH Asia 2014, IEEE Visualization 2005), program chair (IEEE Visualization 2004), as well as program committee member of almost all conferences in the visualization and computer graphics fields for numerous times. Prior to the current post, he was, Dean, School of Computer Science and Technology, SDU. Chen received an MS in Electronic Engineering from Tsinghua University, Beijing (1994), and a second MS (1997) and then PhD (1999) in Computer Science from the State University of New York at Stony Brook.

\end{IEEEbiography}







\end{document}